\documentstyle[fullpage,url,epic]{elsart}
\newcommand{\commentout}[1]{}
\newcommand{\denselist}{\itemsep 3pt}
\newcommand{\<}{\langle}
\renewcommand{\>}{\rangle}
\newcommand{\beqn}{\begin{equation}}
\newcommand{\eeqn}{\end{equation}}
\newcommand{\eg}{e.g.,~}
\newcommand{\ie}{i.e.,~}
\newcommand{\resp}{resp.\ }
\newcommand{\respc}{resp.,\ }
\newcommand{\Thm}{\begin{thm}}
\newcommand{\Lem}{\begin{lem}}
\newcommand{\Pro}{\begin{prop}}
\newcommand{\Cor}{\begin{cor}}
\newcommand{\eThm}{\end{thm}}
\newcommand{\eLem}{\end{lem}}
\newcommand{\ePro}{\end{prop}}
\newcommand{\eCor}{\end{cor}}
\newcommand{\prf}{\begin{pf}}
\newcommand{\eprf}{\qed\end{pf}}
\newcommand{\xam}{\begin{exmp}}
\newcommand{\exam}{\end{exmp}}

\newcommand{\A}{{\cal A}}
\newcommand{\B}{{\cal B}}
\newcommand{\C}{{\cal C}}

\newcommand{\F}{{\cal F}}

\newcommand{\I}{{\cal I}}

\newcommand{\K}{{\cal K}}
\newcommand{\M}{{\cal M}}

\renewcommand{\P}{{\cal P}}

\newcommand{\R}{{\cal R}}
\newcommand{\T}{{\cal T}}

\newcommand{\W}{{\cal W}}

\newcommand{\IN}{\mbox{$I\!\!N$}}
\newcommand{\cross}{\times}
\newcommand{\Circ}{\mbox{{\small $\bigcirc$}}}
\newcommand{\false}{\mbox{{\it false}}}
\newcommand{\true}{\mbox{{\it true}}}

\newcommand{\sat}{\models}
\newcommand{\rimp}{\Rightarrow}
\newcommand{\limp}{\Leftarrow}
\newcommand{\dimp}{\Leftrightarrow}
\newcommand{\union}{\cup}
\newcommand{\inter}{\cap}

\renewcommand{\L}{{\cal L}}
\renewcommand{\S}{{\cal S}}

\newcommand{\Sys}{\I}

\newcommand{\Next}{\Circ}

\newcommand{\Cond}{\mbox{\boldmath$\rightarrow$\unboldmath}}
\newcommand{\RCond}{>}
\newcommand{\Condi}{\Cond_i\,}

\newcommand{\Know}{K}
\newcommand{\Bel}{B}

\newcommand{\True}{\mbox{\it true}}
\newcommand{\False}{\mbox{\it false}}

\newcommand{\intension}[1]{[\![ #1 ]\!]}

\newcommand{\Pl}{\mbox{\rm Pl\/}}
\newcommand{\PL}{\mbox{\it PL}}
\newcommand{\PBox}{N}
\newcommand{\bottom}{\perp}

\newcommand{\Poss}{\mbox{Poss}}

\newcommand{\LCond}{\L^{C}}

\newcommand{\Tref}[1]{Theorem~\ref{#1}}
\newcommand{\Lref}[1]{Lemma~\ref{#1}}
\newcommand{\Pref}[1]{Proposition~\ref{#1}}
\newcommand{\Cref}[1]{Corollary~\ref{#1}}

\newcommand{\Sref}[1]{Section~\ref{#1}}
\newcommand{\BEL}{\mbox{Bel}}

\newcommand{\Sub}{\mbox{{\em Sub}}}
\newcommand{\Subp}{\mbox{{\em Sub}}^+}

\newcommand{\sSub}{\mbox{{\small {\em Sub}}}}

\newcommand{\Card}[1]{\left| #1\right|}

\newenvironment{RETHM}[2]{\it \trivlist \item[\hskip \labelsep{\bf #1 \ref{#2}:}]}{\endtrivlist}
\newcommand{\rethm}[1]{\begin{RETHM}{Theorem}{#1}}
\newcommand{\repro}[1]{\begin{RETHM}{Proposition}{#1}}
\newcommand{\relem}[1]{\begin{RETHM}{Lemma}{#1}}
\newcommand{\recor}[1]{\begin{RETHM}{Corollary}{#1}}

\newcommand{\erethm}{\end{RETHM}}
\newcommand{\erepro}{\end{RETHM}}
\newcommand{\erelem}{\end{RETHM}}
\newcommand{\erecor}{\end{RETHM}}

\newcommand{\MP}{MP}

\renewcommand{\Omega}{W}

\newcommand{\Intension}[1]{[#1]}

\newcommand{\type}{{\it type}}
\newcommand{\Dvalue}{{\it value}}
\newcommand{\strat}{{\it strat}}
\newcommand{\prior}{{\it prior}}
\newcommand{\action}{{\it act}}
\newcommand{\Do}{{\it do}}

\newcommand{\Epsilon}{\Sigma}

\newcommand{\Plass}{\P}

\newcommand{\QUAL}{\mbox{QUAL}}
\newcommand{\CONS}{\mbox{CONS}}
\newcommand{\NORM}{\mbox{NORM}}
\newcommand{\REF}{\mbox{REF}}
\newcommand{\SDP}{\mbox{SDP}}
\newcommand{\UNIF}{\mbox{UNIF}}

\newcommand{\RANK}{\mbox{RANK}}

\newcommand{\sCONS}{\mbox{\scriptsize CONS}}
\newcommand{\sNORM}{\mbox{\scriptsize NORM}}

\newcommand{\sKB}{\mbox{\scriptsize KB}}

\newcommand{\AX}{\mbox{AX}}
\newcommand{\AXT}{\mbox{AX$^T$}}
\newcommand{\sAXT}{\mbox{\scriptsize AX$^T$}}

\renewcommand{\Omega}{W}
\renewcommand{\MP}{\mbox{MP}}
\newcommand{\Diag}[1]{D_{#1}}

\newcommand{\obs}[1]{o_{(#1)}}
\newcommand{\Dobs}[1]{o_{#1}}

\newcommand{\fault}{\mbox{{\it fault}}}
\newcommand{\faulty}{\mbox{{\it faulty}}}
\newcommand{\hi}{\mbox{{\it hi}}}

\newcommand{\Sysclass}{\C}

\newcommand{\MapBack}{\mbox{prev}}

\newenvironment{cond}[1]{\begin{quote}{\bf #1} }{\end{quote}}

\newcommand{\diag}{{\mbox{\scriptsize\it diag}}}

\commentout{
\newcommand{\type}{{\it type}}
\newcommand{\fault}{{\it fault}}
\newcommand{\obs}{{\it obs}}
\newcommand{\strat}{{\it strat}}
\newcommand{\prior}{{\it prior}}
\newcommand{\action}{{\it act}}
\newcommand{\Do}{{\it do}}

\newcommand{\Epsilon}{\Sigma}

\newcommand{\QUAL}{\mbox{QUAL}}
\newcommand{\CONS}{\mbox{CONS}}
\newcommand{\NORM}{\mbox{NORM}}
\newcommand{\REF}{\mbox{REF}}
\newcommand{\SDP}{\mbox{SDP}}
\newcommand{\UNIF}{\mbox{UNIF}}
\newcommand{\RANK}{\mbox{RANK}}

\newcommand{\sCONS}{\mbox{\scriptsize CONS}}
\newcommand{\sNORM}{\mbox{\scriptsize NORM}}

\newcommand{\sKB}{\mbox{\scriptsize KB}}

\newcommand{\AX}{\mbox{AX}}

\renewcommand{\Omega}{W}
\renewcommand{\MP}{\mbox{MP}}
\newcommand{\Diag}[1]{D_{#1}}

\newenvironment{cond}[1]{\begin{quote}{\bf #1}: }{\end{quote}}
}
\renewcommand{\Cond}{\rightarrow}

\newcommand{\citeyear}{\cite}

\commentout{

REFEREE REPORT #1

Review report
=46or Artificial Intelligence Journal

Title: Modeling Belief in Dynamic Systems - Part I: Foundations

Authors: Nir Friedman and Joseph Y. Halpern

Comments:

General comments:

The paper presents a formalism to deal with knowledge, uncertainty, belief,=
 time and multi-agent notions. The main goal is to reason about belief=
 change in general. There are many interesting results and the paper is easy=
 to read up to section 4 where it becomes less fluent for me.

Unfortunately, I hadn't time to check all proofs.

In my opinion, there is no doubt about the quality of this work, but the=
 presentation has to be improved.

Clearly, it is central to this issue as far as belief revision is the major=
 subject. There are also some words on multi-agent systems, and the=
 formalism is clearly built to take multi-agents notions into account. I=
 only regret that this part hasn't been developed so much. Section 3 is=
 named 'Knowledge and Plausibility in Multi-agent Systems', but I don't see=
 any thing requiring multi-agent notions there. Even in the example there is=
 a single agent.

The paper is organized in five sections. Sections 1 and 5 are the=
 introduction and the conclusion. Section 2 is divided into 7 subsections.=
 There is no subsection for section 3 (only 2 pages including one for an=
 example). The section 3 contains only one subsection for 8 pages and many=
 different things: intuitive explanations; formal counterpart of the=
 intuitive explanations; comparison with other approaches; connection with=
 "typically A's are B's"; limits; what happens when relaxing some=
 hypothesis;... I would prefer a more structured plan as for section 2.=
 Furthermore, there is only one little example of three lines to say that=
 'It is easy to verify that...', but nothing on the use of Prior=
 plausibilities which is the subject of this section.=20

Detailed comments:

Page 4:
Semantical definition of knowledge and belief. At the end of the second=
 chapter, the brackets '(of believes)' are a little bit disturbing. Since=
 you deal with belief after the formal definition, you don't need to mix=
 belief and knowledge at this level.

Page 5:
'In this paper, we take knowledge to satisfy the axioms of S5'. Why this=
 choice? You justify most of the other choices you make, but not this one.

pointing to FHMV, but this breaks the flow. I really don't want to get
into the S4.x vs S5 debate here.

Page 7:
Second bullet, second line: 'either Poss([[...' I guess that it has to be=
 'either K([['

Page 9:
Proposition 2.2 is referenced as Theorem 2.2

Your introduction to example 2.4 is: it 'illustrates the use of plausibility=
 measures'. But in fact, you uses preference relations and PPD to formalize=
 the problem and then translate them into plausibility measures. As far as I=
 am concerned, this illustrate the fact that plausibility mesures are=
 powerful enough to express your example, but there is no use of=
 plausibility measures here.

example. The offending line is gone.

page 10:
footnote: Phi is missing at the end of the first line (between 'not' and 'ac=
cording').

Page 11:
Example 2.5: You wrote 'We consider only worlds where the test results are=
 consistent with the set of faulty components'. As far as I understand your=
 example, I think that it is faulty components which have to be consistent=
 with test results.nir22

Example 2.5, second paragraph: 'we can construct two possible=
 structures...'. Does that mean that there are only two possible structures,=
 those you present, or that you focus on these two for some good reasons? It=
 would be great to make things clearer.

is better explained. However, I think the reviewer is overdoing it
here.

Page 13:
Example 2.7: Once again, is the partition you propose the only one solution?=
 If yes, how do you handle the fact that Alice now knows that there are very=
 few Leftfoot on this island and then consider that Bob is plausibly a=
 Rightfoot, and possibly a Leftfoot?

Page 15:
'This discussion shows that our model of belief is more general that the=
 classical Kripke-structure account of beliefs, since there are models where=
 the agent's beliefs are not determined by a set of accessible worlds'. I=
 don' t get this.
As far as beliefs are concerned, you exhibit a model with an infinite set of=
 possible worlds (wi: i>=3D0) for which it's not possible to define an=
 accessibility relation to characterize belief. Does that mean that your=
 models are more general, i.e there is no Kripke structure at all satisfying=
 the same formulae? I don't think so. Especially considering the following=
 of the paragraph: 'this does not lead to new properties of beliefs...',=
 '...we have a finite model property...'.

Part 3:
Title 'Knowledge and plausibility in multi agent systems'. As I understand=
 this part, it has more to do with time than with multi-agent systems. Even=
 in your example 3.1 there is a single agent.

Example 3.1:
You wrote: 'we want to model diagnosis as a system.'. Why? What motivations=
 to model diagnosis as a system? Wasn't the previous formalization satisfact=
ory?

Part 4:
I find it difficult to read. If part one was clear and divided into many=
 subchapters, there is only one subsection here and the section is long and=
 contains many different things.

Part 5: Conclusion: What about the limitation(s) of your formalism?
As far as I see, even if the notion of plausibility is very general and=
 interesting in a theoretical point of view, it is difficult to use it=
 directly in order to represent problems. In all of your examples you use=
 preference ordering, PPD, K-ranking... to model and then you translate them=
 in your framework. Moreover, to reason, you had to consider some particular=
 models. Does that mean that logical deduction of your formalisms is not=
 powerful enough.
How do you address the simplifying assumptions:
        1 - 'there is a single agent'
                Even if your formalism has multi-agent formal settings,=
 there is no part of the paper where interaction between agents are used in=
 an example.
        2 - ...
        3 - 'the new information can be characterized in the language'
                You speak about that in your paper, but without saying how=
 your formalism addresses this problem, or I didn't understood.
You write: '...these simplifying assumptions are not suitable when we want=
 examine belief change in a more realistic settings (... example 3.1)'. But=
 as far as I can see, there is only a single agent in this example,=
 information is characterized in the language and event if the knowledge of=
 the agent increases, example 2.5 formalises the same problem without this f=
eature.
By these remarks, I don't claim that the formalism is not powerful nor=
 interesting.  I believe the converse. The only point is that the form of=
 the article can be improved.=20

=================

REFEREE REPORT #2

Review for the special issue of Artificial Intelligence on Principles of
Multiagent Systems

Paper:   Modeling Beleif in Dynamic Systems. Part I: Foundations
Authors: Nir Friedman and Joseph Halpern

The paper presents a "new" framework for belief change; one that is
sufficently general to capture almost all known frameworks.

Before reading the paper I was not familiar with the authors previous
results, however the paper appears to be a solid piece of previously
unpublished work and the results are sound.

The papers seems interesting enough to accept.

Some of the references point to work "in preparation" e.g. [FH95a] and
[FH95b]. Presumably the editors have a policy on this matter.

}

\begin{document}
\begin{frontmatter}
\title{
Modeling Belief in Dynamic Systems.\\
Part I: Foundations}
\thanks{Some of this work was done while both authors were at the IBM Almaden
Research Center.
The first author was also at Stanford while much of
the work was done.  IBM and Stanford's support are gratefully
acknowledged.   The work
was also supported in part by the Air Force Office of Scientific
Research (AFSC), under Contract F49620-91-C-0080 and grant
F94620-96-1-0323 and by NSF under grants
IRI-95-03109 and IRI-96-25901.
A preliminary version of this paper
appears in {\em Proceedings of the 5th
Conference on
Theoretical Aspects of Reasoning
About Knowledge}, 1994, pp.~44-64, under the title ``A knowledge-based
framework for belief change, Part I: Foundations''.  This version is almost identical to one that will appear in {\em Artificial Intelligence}.}

\author{Nir Friedman}
\address{Computer Science Division, 387 Soda Hall,
	University of California, Berkeley, CA 94720,
        nir@cs.berkeley.edu,
http://www.cs.berkeley.edu/\~nir
}

\author{Joseph Y.\ Halpern}
\address{Computer Science Department,
 Cornell University, Ithaca, NY 14853,
 halpern@cs.cornell.edu,
http://www.cs.cornell.edu/home/halpern
}

\begin{abstract}
Belief change is a fundamental problem in AI: Agents constantly have
    to
update their beliefs to
accommodate new observations.
In recent years, there has been much work on axiomatic characterizations
of belief change. We claim that
a
    better understanding of belief change can be gained from examining
    appropriate {\em semantic\/} models.
    In this paper we propose a general framework in which to model
    belief change.
We begin by defining belief in terms of knowledge and plausibility:
    an agent believes $\phi$ if he knows that
$\phi$ is more plausible than $\neg\phi$.
We then consider some properties defining the interaction between
    knowledge and plausibility, and show
    how these properties affect the properties of belief.  In particular,
    we show that by assuming two of the most natural properties, belief
    becomes a KD45 operator.  Finally, we add time to the picture.
    This gives us a framework in which we can talk about
    knowledge, plausibility (and hence belief), and time, which
    extends the
    framework of Halpern and Fagin
    for modeling knowledge
    in multi-agent systems.
We then examine the problem of ``minimal change''. This notion can be
    captured
by using {\em prior plausibilities},
an analogue to prior probabilities, which
can be
    updated by ``conditioning''.
We show by example that conditioning
on a plausibility measure can
capture many scenarios of interest.
In a companion paper, we show how the two
best-studied scenarios of
    belief change, {\em
    belief revision\/} and {\em belief update\/}, fit into our
    framework.
\end{abstract}
\end{frontmatter}

\section{Introduction}
In order to act in the world we must make assumptions,
    such as ``the corridor is clear''
or ``my car is parked where I     left it''.
These assumptions,
    however, are defeasible. We can easily imagine situations where
    the corridor is blocked, or where the car is stolen. We call the
    logical consequences of such defeasible assumptions {\em
    beliefs.\/} As time passes, we constantly obtain new information that might
    cause us to make additional assumptions or
    withdraw some of our previous assumptions. The problem of {\em
    belief change\/} is to understand how beliefs should change.

The study of belief change has been an active area in philosophy
and in  artificial intelligence \cite{Gardenfors1,KM91}.
In the literature, two instances of this general phenomenon have been
    studied in detail:
    {\em Belief revision\/} \cite{agm:85,Gardenfors1} attempts to
    describe how an agent should
    accommodate a new belief (possibly inconsistent with his other
    beliefs) about a static world. {\em Belief
    update\/} \cite{KM91}, on
    the other hand, attempts to describe how an agent should change his
    beliefs as a result of learning about a change in the world.
Belief revision and belief update describe only two of the many
ways in which beliefs can change.
Our goal is to construct a framework
to reason about
belief change in general. This paper describes the details of that
    framework.
In a companion paper \cite{FrH2Full} we consider the
    special cases of belief revision and update in more detail.

Perhaps the most straightforward approach to belief change is to
    simply represent an agent's beliefs as a closed set of formulas in
    some language and then put constraints on how
these beliefs can change.
This is essentially the approach taken in \cite{agm:85,Gardenfors1,KM91};
as their results
show, much can be done with this framework.
The main problem with this approach is that it does not provide
a good semantics for belief.
As we hope to show in this paper and in \cite{FrH2Full}, such a
semantics
    can give us a much deeper understanding of how and why beliefs
    change.
Moreover, this semantics provides the tools to deal with complicating
    factors such actions, external events, and multiple agents.

\commentout{
One standard approach to giving semantics to belief is to put a
plausibility ordering on a set of worlds (intuitively, the worlds
the agent considers possible).
Using plausibility orderings, we can interpret statements
such as
``it typically does not rain in San Francisco in the summer''.  Roughly
speaking, a statement such as
 ``$\phi$ typically implies $\psi$'' is true at a given world if
$\psi$ is true in the most plausible worlds where $\phi$ is true.
Various authors \cite{Boutilier92,Goldszmidt92,KM91,spohn:88} have
then interpreted ``the agent believes $\phi$'' as ``$\phi$ is true in
    the most plausible worlds that the agent
considers possible''.  Under this interpretation, the agent believes
$\phi$ if $\true$ typically implies $\phi$.%
\footnote{The technique of putting an ordering on worlds has also been
    used to model counterfactuals, conditionals and non-monotonic inference
    \cite{Lewis73,Shoham87,KLM,Pearl90}.  We focus here on its application to
modeling belief.}
}

One standard approach to giving semantics to beliefs is to put a {\em
    preference ordering\/} on
the set of worlds that the agent considers possible.
Intuitively, such an
    ordering captures the relative likelihood of worlds. Various
    authors  \cite{Boutilier92,Goldszmidt92,KM91,spohn:88} have then
    interpreted ``the agent believes $\phi$'' as ``$\phi$ is true in
    the most plausible worlds that the agent considers possible''. An
    alternative approach is to put a probability measure over the set
    of possible worlds. Then we can
interpret ``the agent believes
    $\phi$'' as ``the probability of $\phi$ is close to 1''
    \cite{Pearl90}.
We examine a new approach to modeling uncertainty based on {\em
    plausibility measures},
introduced in \cite{FrH7,FrH5Full},
    where a plausibility measure just
    associates with an event (\ie a set of possible worlds) its {\em
    plausibility}, an element in some partially ordered set. This
    approach is easily seen to generalize other approaches to modeling
    uncertainty, such as probability measures, belief functions, and
    preference orderings. We interpret the ``agent believes $\phi$''
    as ``the plausibility of $\phi$ is greater than that of
    $\neg\phi$''.
As we show, this is often (but not always) equivalent to ``$\phi$ is
true in the most plausible worlds''.

By modeling beliefs in this way, there is an
assumption
    that the
plausibility measure
    is part of the agent's epistemic state.  (This
    assumption is actually made explicitly in \cite{Boutilier92,KLM}.)
This implies that the plausibility measure is {\em subjective\/}, that
is, it
    describes the agent's estimate of
the plausibility of each event.
But actually, an even stronger assumption is being made:  namely,
    that the agent's epistemic state is characterized by a {\em single\/}
    plausibility measure.
We feel
that
this latter assumption makes the models less expressive than
    they ought to be.
In particular, they cannot represent a situation where the agent is
    not sure about what is plausible, such as ``Alice does not
     know that it typically does not rain in San Francisco in the
    summer''.  To capture this, we need to allow Alice to consider several
    plausibility measures possible; in some it typically does not rain
and in
    others it typically does.%
\footnote{In fact, this issue is discussed by Boutilier
    \cite{Boutilier92}, although his framework does not allow him to
    represent such a situation.}
As we shall see, this extra expressive power is necessary to capture some
interesting scenarios of belief change.

To deal with this, in addition to plausibility measures,
we add a standard accessibility relation to represent {\em knowledge}.
Once we have knowledge in the picture, we
define belief by saying that an agent {\em believes\/}
$\phi$ if she {\em knows\/} that $\phi$ is typically true. That is,
    according to all the plausibility measures she considers possible,
$\phi$ is
more plausible than $\neg \phi$.

The properties of belief depend on how the plausibility measure
interacts with the accessibility relation that defines knowledge.
We study these interactions, keeping in mind
that plausibility generalizes probability.
In view of this, it is perhaps not surprising that many of the issues
    studied by Fagin and Halpern \cite{FH3} when considering the
interaction of knowledge and probability also arise in our framework.
    There are, however, a number of new issues that arise in our
    framework due to the interaction between knowledge and belief.
As we shall see, if we take what are perhaps the most natural
restrictions on this interaction, our notion of belief is characterized
by the axioms of the modal logic KD45 (where an agent has complete
introspective knowledge about her beliefs, but may have false beliefs).
Moreover, the interaction between knowledge and belief satisfies
the standard properties considered by Kraus and Lehmann
    \cite{KrausL}.  Although our
    major goal is not an abstract study of
the properties of knowledge and belief, we view the fact that we have a
concrete interpretation under which these properties can be studied to
be an important side-benefit of our approach.

Having a notion of belief is not enough in order to study belief
    change. We want a framework that captures the beliefs of the agent
    before and after the change. This is achieved by introducing {\em
    time\/} explicitly into the framework.
    The resulting framework is an extension of the
    framework of Halpern and Fagin \cite{HF87} for modeling knowledge
    in multi-agent systems, and allows to talk about
    knowledge, plausibility (and hence belief), and time. This
    framework is analogous to combination of knowledge, probability
    and time studied in \cite{HT}.
As we show by example,
having
knowledge, plausibility, and time
    represented explicitly
gives us
a powerful and
    expressive framework for capturing belief change.

This framework is particularly suited to
    studying how plausibility changes over time.
One important intuition we would like to capture is
    that of {\em minimal change.\/} Suppose an agent gets new
    information at time $t$.
Certainly we would expect
 his
    plausibility assessment (and his beliefs) at time $t+1$
to
    incorporate this
    new information;
otherwise, we would expect his assessment at time $t+1$ to have changed
minimally
{f}rom his assessment
    at time $t$.
In probabilistic reasoning, it can be argued that {\em
    conditioning\/} captures this intuition.
Conditioning incorporates the new information by giving it probability
1.  Moreover, the relative probability of
all events consistent
    with the new information
is the same
before
    and after conditioning,
so, in this sense, conditioning changes things minimally.
We focus here on a plausibilistic analogue of conditioning and argue that
    it captures the intuition of minimal change in
    plausibilities. We can then proceed much in the spirit of the
    Bayesian approach, but starting with a {\em prior plausibility\/}
    and conditioning.
As we show, many situations previously
    studied in the literature, such as {\em diagnostic reasoning\/}
    \cite{reiter:87}, can be easily captured by using such prior
    plausibilities.
Moreover,
as we show in a companion
    paper \cite{FrH2Full},
belief revision and belief update---which both attempt to capture
intuitions involving minimal change in beliefs---can
be captured in our framework
by conditioning on an appropriate
prior plausibility measure.
Thinking in terms of priors also gives us insight into other
    representations of belief change, such as those 
of \cite{Boutilier94AIJ2,Goldszmidt92,LamarreShoham}.

The rest of this paper is organized as follows.  In the next section, we
review the syntax and semantics of the standard approach to modeling
knowledge using Kripke structures and show how plausibility can be
added to the framework.
Much of our technical discussion of axiomatizations and decision procedures
is closely related to that of \cite{FH3}.
In Section~\ref{systems}, we present our full framework which
adds plausibility to the framework of \cite{HF87} for modeling knowledge
(and time) in multi-agent systems.
In Section~\ref{priorplaus} we introduce prior plausibilities
    and show how they can be used. We conclude in
    Section~\ref{conclusion} with some discussion
of the general approach.
Proofs of theorems are given in Appendix~\ref{proofs}.

\section{Knowledge and Plausibility}\label{knowledgeandplaus}

In this section, we briefly review the standard models for knowledge
    and beliefs (see \cite{HM2} for further motivation and details),
    describe a notion of plausibility, and then show how to
    combine the two notions. Finally, we compare the derived notion of
    belief with previous
work
on the subject.

\subsection{The Logic of Knowledge}\label{knowledge}
We start by examining
the standard models for knowledge and belief.
The syntax for the logic of knowledge is simple: we start with
    primitive propositions and close off under conjunction, negation,
    and the modal operators $K_1, \ldots, K_n$.  A formula such as
    $K_i \phi$ is read ``agent $i$ knows $\phi$''. The logic of belief
    is the result of replacing the $\Know_i$ operator by $\Bel_i$. The
 formula, $\Bel_i\phi$ is read ``agent $i$ believes $\phi$''.
The resulting languages are denoted
$\L^{K}$ and $\L^{B}$, respectively.

The semantics for these languages is given by means of {\em Kripke
    structures\/}. A {\em Kripke structure for knowledge\/} (or
    {\em belief})
is a tuple $(W, \pi,\K_1, \ldots, \K_n)$, where $W$
    is a set of possible worlds, $\pi(w)$ is a truth assignment to the
    primitive propositions at world $w \in W$, and the $\K_i$'s are
    accessibility relations on the worlds in $W$. For convenience, we define
    $\K_i(w) = \{ w' : (w,w') \in \K_i\}$.
Intuitively, $\K_i(w)$ describes the set of worlds that agent $i$
    considers possible in $w$. We
say that agent $i$ knows (or
    believes) $\phi$ at
world
    $w$, if all the worlds $\K_i(w)$ satisfy $\phi$.

We assign truth values to formulas at each world in the structure. We
    write $(M,w) \models \phi$ if the formula $\phi$ is true at a
    world $w$ in the Kripke structure $M$.
\begin{itemize}\denselist
 \item $(M,w) \models p$ for a primitive proposition $p$ if $\pi(w)(p) = \True$,
 \item $(M,w) \models \neg\phi$ if  $(M,w)\not\models \phi$,
 \item $(M,w) \models \phi\land\psi$ if $(M,w)\models\phi$ and $(M,w)\models\psi$,
 \item $(M,w) \models \Know_i\phi$ if $(M,w')
    \models \phi$ for all $w' \in \K_i(w)$.
\end{itemize}
The last clause captures the intuition that $\phi$ is known exactly
    when it is true in all possible worlds.
When considering the language of beliefs $\L^{B}$, we typically
use $\B_i$ rather than $\K_i$ to denote
    the accessibility relations.
The
    truth condition for
    $\Bel_i\phi$ is exactly the same as for $\Know_i\phi$.

Let $\M_K$ be the class of Kripke structures described above. We say
that
    $\phi \in \L^K$ is valid in some $M\in\M_K$ if $(M,w)
    \sat \phi$ for all $w$ in $M$. We say that $\phi\in\L^K$ is {\em
    valid\/} in $\M_K$
    if it is valid in all models $M\in\M_K$. We say that $\phi$ is
    satisfiable in $\M_K$ if there is a model $M\in\M_K$ and world $w$ such
    that $(M,w) \sat \phi$.

The definition of Kripke structure does not put any restriction on the
    $\K_i$ relations.
By imposing conditions on the $\K_i$
    relations we get additional properties of knowledge (or belief).
    These properties are captured by systems of axioms that describe the
    valid formulas in classes of structures that satisfy various constraints
    of interest. We briefly describe these systems and the
    corresponding constraints on the accessibility relations. Consider
    the following axioms and rules:
\begin{description}\denselist
  \item[K1.] All substitution instances of propositional tautologies
  \item[K2.] $\Know_i\phi\land\Know_i(\phi\rimp\psi) \rimp \Know_i\psi$
  \item[K3.] $\Know_i\phi \rimp \phi$
  \item[K4.] $\Know_i\phi \rimp \Know_i\Know_i\phi$
  \item[K5.] $\neg\Know_i\phi \rimp \Know_i\neg\Know_i\phi$
  \item[K6.] $\neg\Know_i\False$
  \item[RK1.]  From $\phi$ and $\phi \rimp \psi$ infer $\psi$
  \item[RK2.] From $\phi$ infer $\Know_i\phi$
\end{description}
The system K contains the axioms K1 and K2 and the rules of inference
    RK1 and RK2. By adding axioms K4 and K5 we get system K45; if in
    addition we add axiom K6 we get system KD45; if instead we add
    axiom K3 to K45 we get the axiom system known as S5.

We now relate these axiom systems with restrictions on the
    accessibility relations. We start with some definitions.
 A relation $\R$ on $W$ is {\em Euclidean\/} if $(x,y),(x,z) \in \R$ implies
    that $(y,z) \in \R$, for all $x,y$ and $z$ in $W$; it is {\em
    reflexive\/} if  $(x,x) \in \R$ for all $x \in W$; it is {\em serial\/}
    if for all $x\in W$ there is a $y$ such that $(x,y) \in R$; and
it is {\em transitive\/} if $(x,y),(y,z) \in \R$
    implies that $(x,z) \in \R$, for $x,y$ and $z$ in $W$.
Let $\M_K^{et}$ be the set of Kripke structures with Euclidean and
    transitive accessibility relations, $\M_K^{est}$ be the subset of
    $\M_K^{et}$ where the accessibility relations are also serial,
    and $\M_K^{ert}$  be the subset of $\M_K^{et}$ where the
    accessibility relations are also transitive.
\Thm{\rm\cite{HM2}}\label{thm:S5}
The axiom system K (\resp K45, KD45, S5) is a sound and complete
    axiomatization of $\L^K$ with respect to $\M_K$ (\resp $\M_K^{et}$,
    $\M_K^{est}$, $\M_K^{ert}$).
\eThm

In this paper, we use the multi-agent systems formalism of \cite{FHMV}
to model knowledge; this means that knowledge satisfies the axioms of
S5.  (We provide some motivation for this choice below; see \cite{FHMV}
for further discussion.)

This
    implies that if an agent knows $\phi$, then $\phi$ is true (K3)
    and that the agent is introspective---he knows what he knows and
    does not know (K4 and K5). Belief, on the other hand, is typically
    viewed as defeasible. Thus, it does not necessarily satisfy K3.
    It may satisfy a weaker property, such as K6, which says that the
    agent does not believe inconsistent formulas. Like knowledge,
    belief is taken to be introspective, as it satisfies K4 and K5.
    Thus, in the literature, belief has typically been take to satisfy
    K45 or KD45; we do the same here. According to \Tref{thm:S5}, this
    means that the notion of knowledge we use is characterized by
    $\M_K^{ert}$ while belief is characterized by $\M_K^{et}$ or
    $\M_K^{est}$.%
\footnote{As is well known, a relation is reflexive, Euclidean and
    transitive if and only if it is an equivalence relation (\ie
    reflexive, symmetric and transitive). Thus, $\M_K^{ert}$ consists
    of these structures where the $\K_i$'s are equivalence relations.}

\subsection{Plausibility Measures}\label{plaus}
Most non-probabilistic approaches to belief change require (explicitly
    or implicitly) that the
    agent has some ordering over possible alternatives.
    For example, the agent might have a preference ordering over
    possible worlds \cite{Boutilier94AIJ2,Grove,KM92} or an
    entrenchment ordering over
    formulas
    \cite{MakGar}.
This ordering dictates how the agent's beliefs change. For example, in
    \cite{Grove}, the new beliefs are
    characterized by the most preferred worlds that are consistent
    with the new observation, while in \cite{MakGar} beliefs are
    discarded according to their degree of entrenchment until
it is consistent to add the new observation to the resulting set of
    beliefs.

Keeping this insight in mind, we now describe {\em plausibility
    measures\/}
\cite{FrH7,FrH5Full}.  This
    is a notion for
    handling uncertainty that generalizes previous approaches,
    including various notions of
    preference
    ordering.
We briefly review the
    relevant definitions and results here.

Recall that a probability space is a tuple $(W,\F,\Pr)$, where
$W$ is a set of worlds, $\F$ is an algebra of {\em measurable\/}
    subsets of $W$ (that is, a set of subsets closed under union
    and complementation to which we assign probability), and $\Pr$ is a
    {\em probability measure}, that is, a function mapping each set in
    $\F$ to a number in $[0,1]$ satisfying the well-known probability
    axioms ($\Pr(\emptyset) = 0$, $\Pr(W) = 1$, and $\Pr(A \union
    B) = \Pr(A) + \Pr(B)$, if $A$ and $B$ are disjoint).

A plausibility space is a direct generalization of a probability
    space. We simply replace the probability measure $\Pr$ by a {\em
    plausibility measure\/} $\Pl$, which, rather than mapping sets in
    $\F$ to numbers in $[0,1]$, maps them to elements in some
    arbitrary partially ordered set. We read $\Pl(A)$ as ``the
    plausibility of set $A$''.  If $\Pl(A) \le \Pl(B)$, then $B$ is at
    least as plausible as $A$. Formally, a {\em plausibility space\/}
    is a tuple $S = (W,\F,\Pl)$, where $W$ is a set of worlds, $\F$
    is an algebra of subsets of $W$,
    and $\Pl$ maps sets in $\F$ to some domain  $D$ of {\em
    plausibility values\/} partially ordered by a relation $\le_D$ (so
    that $\le_D$ is reflexive, transitive, and anti-symmetric).
We assume that $D$
is {\em pointed\/}: that is, it contains
two special elements $\top_D$ and $\bottom_D$ such that $\bottom_D \le_D d
    \le_D \top_D$ for all $d \in D$; we further assume that $\Pl(\Omega)
= \top_D$ and
    $\Pl(\emptyset) = \bottom_D$.
As usual, we define the ordering $<_D$ by taking $d_1 <_D d_2$ if $d_1
    \le_D d_2$ and $d_1 \neq d_2$.
We
omit the subscript $D$ from $\le_D$, $<_D$, $\top_D$ and $\bottom_D$
whenever it is clear from context.

Since we want a set to be
    at least as plausible as any of its subsets, we require
\begin{cond}{A1}
If $A \subseteq B$, then $\Pl(A) \le \Pl(B)$.
\end{cond}

Some brief remarks on this definition:  We have deliberately suppressed
the domain $D$
of plausibility values
from the tuple $S$, since
for the purposes of this paper, only the ordering induced by $\le$
    on the subsets in $\F$ is relevant.
The algebra $\F$ also does not play a significant role in this paper.
    Unless we say otherwise, we
assume $\F$ contains all
    subsets of interest and suppress mention of $\F$, denoting a
    plausibility space as a pair $(W,\Pl)$.

Clearly plausibility spaces generalize probability spaces.
We now briefly discuss a few other notions of uncertainty that they
    generalize:
\begin{itemize}
\item A {\em belief function\/} $\Bel$ on $W$ is a function
    $\Bel: 2^{W} \rightarrow [0,1]$ satisfying certain axioms
    \cite{Shaf}.  These axioms certainly imply property A1, so a
    belief function is a plausibility measure.

\item A {\em fuzzy measure\/} (or a {\em Sugeno measure\/}) $f$ on
    $\Omega$ \cite{WangKlir} is a function $f:2^{\Omega}\mapsto
    [0,1]$, that satisfies A1 and some continuity constraints.
A {\em possibility measure\/} \cite{DuboisPrade88} $\Poss$ is a fuzzy
measure such that $\Poss(W) = 1$, $\Poss(\emptyset) = 0$, and
$\Poss(A) = \sup_{w \in A}(\Poss(\{w\})$.

\item An {\em ordinal ranking\/} (or {\em $\kappa$-ranking\/}) on
    $\Omega$ (as defined by \cite{Goldszmidt92}, based on ideas that
    go back to \cite{spohn:88}) is a function $\kappa: 2^\Omega
    \rightarrow \IN^*$, where $\IN^* = \IN \union \{\infty\}$, such
    that  $\kappa(\Omega) = 0$, $\kappa(\emptyset) = \infty$, and
    $\kappa(A) = \min_{w\in A}(\kappa(\{w\}))$. Intuitively, an
    ordinal ranking assigns a degree of surprise to each subset of
    worlds in $\Omega$, where $0$ means unsurprising and higher
    numbers denote greater surprise. It is easy to see that if
    $\kappa$ is a ranking on $\Omega$, then $(\Omega, \kappa)$
    is a plausibility space, where $x \le_{\IN^*} y$ if and only if $y
    \le x$ under the usual ordering on the ordinals.

\item A {\em preference ordering\/} on $W$ is a partial order
    $\prec$ over $W$ \cite{KLM,Shoham87}. Intuitively, $w \prec w'$
    holds if $w$ is {\em
    preferred\/} to $w'$. Preference orders have been used to provide
    semantics for {\em default\/} (\ie conditional) statements.
    In \cite{FrH5Full} we show how to map preference orders on $\Omega$ to
    plausibility measures on $W$ in a way that preserves the
    ordering of events of the form $\{ w\}$ as well as the truth
    values of defaults. We review these results below.

\item A {\em parametrized probability distribution \/} (PPD) on $W$ is a
    sequence $ \{\Pr_i : i \ge 0\}$ of
    probability measures over $W$. Such structures provide semantics
    for defaults in {\em $\epsilon$-semantics\/}
    \cite{Pearl90,GMPfull}.  In \cite{FrH5Full} we show how to map
    PPDs into plausibility structures in a way that preserves the
    truth-values of conditionals (again, see discussion below).
\end{itemize}

\subsection{The Logic of Conditionals}\label{conditionals}

Our goal is to describe the agent's beliefs in terms of plausibility.
    To do this, we describe how to evaluate statements of the form
    $\Bel\phi$ given a plausibility space. In fact, we examine a
    richer logical language that also allows us to describe how the agent
    compares different alternatives. This is the
    logic of conditionals.
Conditionals are statements of the form $\phi\Cond\psi$,
    read ``given
    $\phi$, $\psi$ is plausible'' or ``given $\phi$, then by default
    $\psi$''. The syntax of the logic of
    conditionals is simple: we start with primitive propositions and
    close off under conjunction, negation and the modal operator
    $\Cond$. The resulting language is denoted $\LCond$.

Many semantics have been proposed in the literature for conditionals.
Most
of them involve structures of the form $(W,X,\pi)$, where $W$
    is a set of possible worlds, $\pi(w)$ is a truth assignment to
    primitive propositions, and $X$ is some ``measure'' on $W$ such as
    a preference ordering, a $\kappa$-ranking,
or a possibility measure.
We now describe some of the proposals in the literature, and
    then show how they can be viewed as using plausibility measures.
Given a structure $(W,X,\pi)$,
let $\intension{\phi} \subseteq W$ be
    the set of worlds
    satisfying $\phi$.
\begin{itemize}\denselist
 \item A {\em  possibility structure\/} is a tuple $(W,\Poss,\pi)$,
    where $\Poss$ is a possibility measure on $W$.
     It satisfies a conditional $\phi\Cond\psi$ if either
    $\Poss(\intension{\phi}) = 0$ or $\Poss(\intension{\phi\land\psi})
    > \Poss(\intension{\phi\land\neg\psi})$ \cite{DuboisPrade:Defaults91}.
That is, either $\phi$ is impossible, in which case the conditional holds
    vacuously, or $\phi\land\psi$ is more possible than
    $\phi\land\neg\psi$.

 \item A {\em $\kappa$-structure\/}  is a tuple $(W,\kappa,\pi)$, where
    $\kappa$ is an ordinal ranking on $W$. It satisfies a conditional
    $\phi\Cond\psi$ if either
$\kappa(\intension{\phi}) = \infty$ or
    $\kappa(\intension{\phi\land\psi}) <
    \kappa(\intension{\phi\land\neg\psi})$ \cite{Goldszmidt92}.

 \item A {\em preferential structure\/} is a tuple $(W,\prec,\pi)$,
    where $\prec$ is a partial order on $W$. The intuition
    \cite{Shoham87} is that a preferential structure
    satisfies a conditional $\phi\Cond\psi$ if all the most
    preferred worlds (\ie the minimal worlds according to $\prec$) in
    $\intension{\phi}$ satisfy $\psi$.  However,
there may be
no minimal worlds in $\intension{\phi}$. This can happen
    if $\intension{\phi}$ contains an infinite descending
    sequence $\ldots \prec w_2 \prec w_1$. What do we
    do in these structures?  There are
    a number of options: the first is to assume that,
for each formula $\phi$, there are minimal  worlds in
    $\intension{\phi}$; this is the
    assumption actually made in \cite{KLM}, where it is called the
    {\em smoothness\/} assumption.
A yet more general definition---one that works even if $\prec$ is
not smooth---is given in
    \cite{Lewis73,Boutilier94AIJ1}.
    Roughly speaking, $\phi \Cond \psi$ is true if, from a certain
    point on, whenever $\phi$ is true, so is $\psi$.
More formally,
\begin{quote}
    $(W,\prec,\pi)$ satisfies  $\phi\Cond\psi$, if
    for every
world $w_1 \in \intension{\phi}$,
    there is a world $w_2$ such that (a) $w_2 \preceq w_1$
(so that $w_2$ is at least as normal as $w_1$),
    (b) $w_2 \in \intension{\phi\land\psi}$, and (c) for all
    worlds $w_3 \prec w_2$, we have
     $w_3 \in \intension{\phi \rimp \psi}$ (so any world more
    normal than $w_2$ that satisfies $\phi$ also satisfies $\psi$).
\end{quote}
It is easy to verify that this definition is equivalent to the
earlier one if $\prec$ is smooth.

\item A {\em PPD structure\/} is a tuple $(W,\{ \Pr_i : i \ge 0
    \},\pi)$, where $\{\Pr_i\}$ is PPD over $W$.
    Intuitively, it satisfies a conditional $\phi\Cond\psi$ if the
    conditional probability $\psi$ given $\phi$ goes to $1$ in the
    limit. Formally, $\phi\Cond\psi$ is satisfied if $\lim_{i
    \rightarrow \infty}\Pr_i(\intension{\psi}|\intension{\psi}) = 1$
     \cite{GMPfull} (where
    $\Pr_i(\intension{\psi}|\intension{\phi})$ is taken to be 1 if
    $\Pr_i(\intension{\phi}) = 0$).
\end{itemize}

In \cite{FrH5Full} we use plausibility to provide semantics for
    conditionals and show that our
    definition generalizes the definition in the various approaches we
    just described.
We briefly review the definitions and results here.

A {\em  plausibility structure\/} is a tuple $\PL = (W, $\Pl$, \pi)$,
    where $\Pl$ is a plausibility measure on $W$.
 Conditionals are evaluated according to a rule that is essentially
    that used in possibility structures:
\begin{itemize}\denselist
 \item $\PL \sat \phi\Cond\psi$ if either
    $\Pl(\intension{\phi}) = \bottom$ or
    $\Pl(\intension{\phi\land\psi}) >
    \Pl(\intension{\phi\land\neg\psi})$.
\end{itemize}
Intuitively, $\phi \Cond \psi$ holds vacuously if $\phi$ is
    impossible; otherwise, it holds if $\phi \land \psi$ is more
    plausible than $\phi \land \neg \psi$. It is easy to see that this
    semantics for conditionals generalizes the semantics of
    conditionals in
    possibility structures and $\kappa$-structures.
The following result shows that
    it also generalizes the semantics of conditionals in preferential
    structures and PPD structures.
\Pro{\rm\cite{FrH5Full}}\label{pro:prec}
\begin{enumerate}
\item[(a)]
If $\prec$ is a preference ordering on $W$, then there is a
plausibility measure $\Pl_\prec$ on $W$ such that
$(W,\prec,\pi) \sat \phi\Cond\psi$ if and only if $(W,\Pl_\prec,\pi) \sat
    \phi\Cond\psi$.
\item[(b)]
If $PP = \{ \Pr_i \}$ is a PPD on $W$, then there is a plausibility
measure $\Pl_{PP}$ such that
$(W,\{\Pr_i\}, \pi) \sat
    \phi\Cond\psi$ if and only if  $(W,\Pl_{PP},\pi) \sat \phi\Cond\psi$.
\end{enumerate}
\ePro

We briefly describe the construction of $\Pl_\prec$ and $\Pl_{PP}$ here,
since we use them in the sequel.  Given a preference order $\prec$ on
$W$,
 let $D_0$ be the domain of plausibility values consisting of
one element $d_w$ for every element $w \in W$.
We define a partial order on $D_0$ using $\prec$:
$d_v < d_w$
if $w \prec v$. (Recall that $w \prec
    w'$ denotes that $w$ is preferred to $w'$.)  We then take $D$ to
    be the smallest set containing
$D_0$
that is
closed
under
least upper bounds (so that every set
    of elements in $D$
has a least upper bound in $D$).
For a subset $A$ of $W$, we can then define $\Pl_\prec(A)$ to be the
least upper bound of $\{d_w: w \in A\}$.  Since $D$ is closed under least
upper bounds, $\Pl_\prec(A)$ is well defined.  As shown in \cite{FrH5Full},
this choice of $\Pl_\prec$ satisfies Proposition~\ref{pro:prec}.

The construction in the case of PPD's is even more straightforward.
Given a PPD $PP = \{Pr_i\}$ on
$W$, we define $\Pl_{PP}$ as follows:
\begin{quote}
$\Pl_{PP}(A) \le \Pl_{PP}(B)$ if and only if $\lim_{i \rightarrow
    \infty}\Pr_i(B|A \union B) =  1$.
\end{quote}
A straightforward argument shows that
this choice of $\Pl_{PP}$ satisfies Proposition~\ref{pro:prec}.

These results show that our semantics for conditionals in plausibility
    structures generalizes the various approaches examined in the
    literature. Does it capture our intuitions about conditionals? In
    the AI literature,
there has been discussion of the right
    properties of default statements (which are essentially
    conditionals).
While there has been little consensus on what the ``right''
properties for defaults should be, there has been some consensus on
a reasonable ``core'' of inference rules for default reasoning.
This core, known as the KLM properties
\cite{KLM},
consists of
    the following axiom and rules of inference:
\begin{description}\denselist
 \item[LLE.] {From} $\phi \dimp \phi'$ and
 $\phi\Cond\psi$ infer
    $\phi'\Cond\psi$ (left logical equivalence)
 \item[RW.] {From} $\psi \rimp \psi'$ and
 $\phi\Cond\psi$ infer
    $\phi\Cond\psi'$ (right weakening)
 \item[REF.] $\phi\Cond\phi$ (reflexivity)
 \item[AND.] From $\phi\Cond\psi_1$ and $\phi\Cond\psi_2$ infer
    $\phi\Cond \psi_1 \land \psi_2$
 \item[OR.] From $\phi_1\Cond\psi$ and $\phi_2\Cond\psi$ infer
    $\phi_1\lor\phi_2\Cond \psi$
 \item[CM.] From $\phi\Cond\psi_1$ and $\phi\Cond\psi_2$ infer
    $\phi\land \psi_1 \Cond \psi_2$ (cautious monotonicity)
\end{description}
LLE states that the syntactic form of
    the antecedent is irrelevant. Thus, if $\phi_1$ and $\phi_2$ are
    equivalent, we can deduce $\phi_2\Cond\psi$ from
    $\phi_1\Cond\psi$.  RW describes a
    similar property of the consequent: If $\psi$ (logically) entails
    $\psi'$, then we can deduce $\phi\Cond\psi'$ from $\phi\Cond\psi$.
    This allows us to can combine default and logical reasoning.
    REF states that $\phi$ is always a default conclusion
    of $\phi$. AND states that we can combine two default conclusions:
    If we can conclude by default both $\psi_1$ and $\psi_2$ from
    $\phi$, we can also conclude  $\psi_1\land\psi_2$ from $\phi$.
    OR states that we are allowed to reason by cases: If the
    same default conclusion follows from each of two antecedents, then
it also follows from their disjunction. CM
    states that if $\psi_1$ and $\psi_2$ are two default
    conclusions of $\phi$, then discovering that $\psi_1$ holds
    when $\phi$
    holds (as would be
    expected, given the default) should not cause us to retract the
    default conclusion $\psi_2$.

Do conditionals in plausibility structures
    satisfy the KLM properties?
In general, the answer is no. It is almost immediate from the
    definition that a probability measure $\Pr$ is also a plausibility
    measure.
Notice
that $\Pr(\intension{\phi\land\psi} ) >
    \Pr(\intension{\phi\land\neg\psi})$ if and only if
    $\Pr(\intension{\psi} \mid \intension{\phi}) > 1/2$.
Expanding the semantics of conditionals, we get that
    $\phi \Cond \psi$ holds in $\Pr$ exactly if
    $\Pr(\intension{\phi}) = 0$ or     $\Pr(\intension{\psi} \mid
    \intension{\phi}) > 1/2$.  It is easy to see that this definition
    does not satisfy the
    AND rule: it is not in general the
    case that $\phi \Cond \psi_1$ and $\phi \Cond \psi_2$
together
    imply $\phi \Cond
    (\psi_1 \land \psi_2)$, since $\Pr(A_1 \mid B) > 1/2$ and $\Pr(A_2
    \mid B) > 1/2$ do not imply $\Pr(A_1 \inter A_2|B) > 1/2$. Since
    the AND rule is a fundamental feature of qualitative reasoning, we
    would like to restrict to plausibility structures where it holds.
    In \cite{FrH5Full} we show that the following condition is necessary
    and sufficient to guarantee that the And rule holds:
\begin{cond}{A2}
If $A$, $B$, and $C$ are pairwise disjoint sets,
 $\Pl(A \union B) > \Pl(C)$, and $\Pl(A \union C) > \Pl(B)$, then
 $\Pl(A) > \Pl(B \union C)$.
\end{cond}
It turns out that conditionals in plausibility structures that satisfy
    A2 also
satisfy
LLE, RW, and CM. They also satisfy OR when
    one of the conditionals $\phi_1\Cond\psi$ and $\phi_2\Cond\psi$ is
    satisfied non-vacuously (that is, in a plausibility measure $\Pl$
    such that either $\Pl(\intension{\phi_1}) > \bot$ or
    $\Pl(\intension{\phi_2}) > \bot$).
    To satisfy OR in general we need another
    condition:
\begin{cond}{A3}
If $\Pl(A) = \Pl(B) = \bottom$, then $\Pl(A \union B) = \bottom$.
\end{cond}
A3 also has a nice axiomatic characterization.  Let
    $\PBox \phi$ be an abbreviation for
 $\neg\phi\Cond\False$. (This operator is called the ``outer
    modality'' in \cite{Lewis73}.) Expanding the definition
    of $\Cond$, we get that $\PBox\phi$ holds at $w$ if and only
    if $\Pl(\intension{\neg\phi}) =
    \bottom$. Thus, $\PBox\phi$ holds if $\neg\phi$ is
    considered completely implausible. We can think of the $\PBox$
    modality as the
    plausibilistic version of necessity.  It is easy to show that
A3 corresponds to an AND rule for $\PBox$.  It holds exactly if
 $(\PBox \phi \land \PBox \psi) \rimp \PBox(\phi \land \psi)$.

A plausibility space $(W,\Pl)$ is {\em qualitative\/}
if it satisfies A2 and A3.
A plausibility structure $(W,\Pl,\pi)$ is qualitative
if $(W,\Pl)$ is a qualitative plausibility space.
    In \cite{FrH5Full} we show that, in a very
    general sense, qualitative plausibility structures capture
    default reasoning.
    More precisely, we show that the KLM properties
    are sound with
    respect to a class of plausibility structures if and only if the
    class consists of qualitative plausibility structures. We also
    show that a very weak condition is necessary and sufficient in
    order for the KLM properties to be complete axiomatization of the
    language of default entailment considered in \cite{KLM}.
These results help explain why so many different approaches to giving
semantics to conditionals are characterized by the KLM properties.  In
addition, as we shall see, it also shows that if we want belief to have
some reasonable properties, then we need to restrict to qualitative
plausibility measures.

\commentout{
The following example illustrates the use of plausibility measures.

\xam
\label{xam:diag-order}
The circuit diagnosis problem has been well studied in the literature
    (see \cite{davis} for an overview). Consider a circuit that
    contains $n$ logical components $c_1,\ldots,c_n$. Our target is
    to construct a plausibility measure over the possible failures of
    the circuit. A {\em  failure set\/} is taken to be a set of faulty
    components. We assume that failures of individual
    components are independent of one another. If we also assume that
    the likelihood of each component failing is the same,
    we can construct a plausibility measure as follows: If $f_1$ and
    $f_2$ are two failure sets, we say that $f_1$ is more plausible than
    $f_2$ if $\Card{f_1} < \Card{f_2}$, that is,
if $f_1$ consists of fewer
    faulty components than $f_2$. This definition leads to a
    preference ordering over failure sets. This preference ordering
    induces a plausibility measure using the construction of
\Pref{pro:prec}.
In this measure $\Pl(F_1) < \Pl(F_2)$ if $\min_{f \in F_1}(\Card{f})
    < \min_{f \in F_2}(\Card{f})$.

We can construct the same plausibility measure based on probabilistic
    arguments using PPDs.
Suppose that the probability of a single component failing is
$\epsilon$. Since we have assumed
    that failures are independent,
it follows that the
    probability of a failure set $f$ is
$\epsilon^{\Card{f}}(1-\epsilon)^{(n-\Card{f})}$, since
    there are $\Card{f}$ components that fail, and $n - \Card{f}$
    components that do not fail.
To model the behavior of small but unknown
failure probability, we can consider the PPD $(\Pr_0, \Pr_1, \ldots)$,
where in $\Pr_m$ the probability of a single failure is $1/(m+1)$.
It is not hard to check that $\lim_{m\rightarrow
    \infty}\Pr_m(F_2)/\Pr_m(F_1) =
    0$ if and only if
$\Pl(F_2) < \Pl(F_1)$ in the plausibility
    measure described above. Interestingly, this
    plausibility measure is almost identical to the $\kappa$-ranking
    in which $\kappa(\{f\}) = |f|$. The only difference is that
    if $\Card{f_1} = \Card{f_2}$, $\Pl(\{f_1\})$ is incomparable to
    $\Pl(\{f_2\})$ in the plausibility measure we constructed, while
they are equal according to the $\kappa$-ranking.

In some situations it might be unreasonable to assume that all components
    have the same probability of failure. Thus, we might assume that for
    each component $c_i$ there is a probability $\epsilon_i$ of
    failure. If we assume independence, then given $\vec\epsilon =
    (\epsilon_1,\ldots,\epsilon_n)$, the probability of a failure set
    $f$ is
$\Pi_{c_i\in
    f}\epsilon_i \,\Pi_{c_i\not\in f}(1 - \epsilon_i)$.
We can construct a PPD that captures the effect of the $\epsilon_i$'s
getting smaller, but at possibly different rates:
Suppose $g$ is a bijection from $\IN^m$ to $\IN$.  If $\vec{m} = (m_1,
\ldots, m_n)$, let $\Pr_{g(\vec{m})}$ be the distribution where the
probability of $c_i$ failing is $1/(m_i+1)$, for $i = 1, \ldots, n$.
In this case, we get that $\lim_{m \rightarrow \infty}
\Pr_m(f_2)/\Pr_m(f_1) = 0$ if
    only if $f_2$ is a strict subset of $f_1$, \ie if
    $f_1$ contains all the components in $f_2$ and
    more.
Since we do not assume any
    relations among the failure probabilities of different components,
    it is not possible
    to compare failure sets unless one is a subset of the other.
Thus, we
    can define $f \prec f'$ if $f \subset f'$.
Using the construction
    of \Pref{pro:prec}, we can again consider the plausibility measure
    $\Pl$ induced by $\prec$. It is not hard to see that $\Pl(F_1) \le
    \Pl(F_2)$ if for every failure set
    $f_1 \in F_1 - F_2$ there is some $f_2 \in F_2$ such that
$f_2 \prec f_1$.
As our construction shows, this plausibility measure can be induced by
either a preference ordering or a PPD; however, it cannot be captured by
a $\kappa$-ranking or a possibility measure, since the ordering on
failure sets is partial.
\exam
}

\subsection{Combining Knowledge and Plausibility}
\label{combining}
\label{COMBINING}

We now define a logic that combines knowledge and plausibility.
Let $\L^{KC}$ be the language obtained by starting with primitive
propositions, and closing off under conjunction, negation, and the
operators $K_i$ and $\Condi$, $i = 1, \ldots, n$.
Note that we have a different conditional operator for each agent. We
read $\phi \Condi \psi$ as
``according to agent $i$'s plausibility measure,
    $\phi$ typically implies $\psi$''.

A {\em (Kripke) structure (for knowledge and plausibility)\/} is
    a tuple $(W,\pi, \K_1, \ldots, \K_n, \Plass_1, \ldots, \Plass_n)$
where $W$, $\pi$ and
    $\K_i$ are just as in
    Kripke structures for knowledge,
while  $\Plass_i$ is a
{\em plausibility assignment}, a
function that
    assigns a
    plausibility space to agent $i$ at each world.  Intuitively,
    the structure
    $\Plass_i(w) = (\Omega_{(w,i)},\Pl_{(w,i)})$ captures
    agent $i$'s plausibility measure in the world $w$. For
    now we allow $\Omega_{(w,i)}$ to be an arbitrary subset
of $W$.
    We discuss some possible restrictions on $\Omega_{(w,i)}$
    below.
It is reasonable to ask at this point where the plausibility spaces
$\Plass_i(w)$ are coming from, and why we need a different one for each
agent at each world.  The answer to this question depends very much on
the intended application.  We defer further discussion of this
issue until later.

We can now give semantics to formulas in $\L^{KC}$ in Kripke structures
for knowledge and plausibility. This is done in a recursive way
    using the rules specified above for $\L^{K}$ and $\LCond$.
    Statements of the form $\Know_i \phi$ are evaluated according to
    $\K_i$:
\begin{itemize}\denselist
 \item $(M,w) \sat \Know_i\phi$ if $(M,w')
    \sat \phi$ for all $w' \in \K_i(w)$.
\end{itemize}
Statements of the form $\phi\Condi\psi$ are evaluated
    according to $\Plass_i$. Let $\intension{\phi}_{(w,i)} = \{ w' \in
    W_{(w,i)} : (M,w') \sat \phi \}$.
\begin{itemize}\denselist
 \item $(M,w) \sat \phi\Condi\psi$ if either
    $\Pl_{(w,i)}(\intension{\phi}_{(w,i)}) = \bottom$
    or
    $\Pl_{(w,i)}(\intension{\phi\land\psi}_{(w,i)}) >
    \Pl_{(w,i)}(\intension{\phi\land\neg\psi}_{(w,i)})$.
\end{itemize}

We now define beliefs. Recall that $\True\Condi\phi$ means that $\phi$
    is more plausible than
$\neg \phi$
according to agent's $i$ plausibility
    measure. We might say that in this case the agent believes $\phi$.
    However, recall that the agent can have different plausibility
    assessments at different worlds. Thus, there can be a model $M$,
    and worlds $w, w'$ such that $(w,w') \in
    \K_i$, but $(M,w) \sat \True\Condi\phi$ while $(M,w') \sat
    \neg(\True\Condi\phi)$.
(In Example~\ref{xam:Alice}, we show why this extra expressive power is
necessary.)
    That is, $\phi$ is more plausible than $\neg \phi$ in one of the
    worlds the agent considers possible, but not in another.
    Since our intention is that the agent should not
    distinguish between accessible worlds, we would like the
    agent to have the same beliefs in all the worlds he considers
    possible. We say that an agent {\em believes\/}
     $\phi$ if he knows that
$\phi$ is more plausible than $\neg \phi$ in all the worlds he considers
possible.
     Thus, we define $\Bel_i \phi$, read ``agent $i$ believes
    $\phi$'', as an abbreviation for $\Know_i(\True\Condi\phi)$.
\subsection{Example: Circuit Diagnosis}\label{sec:xam-diag}

The following example illustrates some of the expressive power
of this language.
Although it only involves one agent and only one plausibility measure in
any given structure, it can easily be extended to allow for many agents
with different plausibility measures.

\begin{figure}
\begin{center}
\input{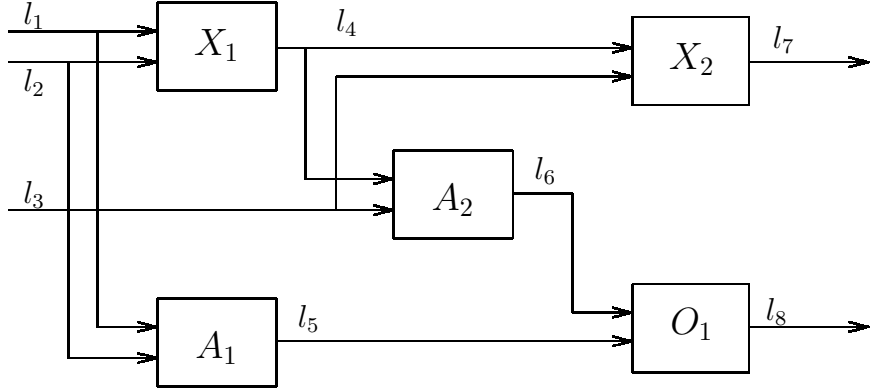}
\end{center}
\caption{A full adder. $X_1$ and $X_2$ are XOR gates, $A_1$ and $A_2$ are
AND gates, and $O_1$ is an OR gate.}
\label{fig:circuit}
\end{figure}

The circuit diagnosis problem has been well studied in the literature
(see \cite{davis} for an overview). Consider a circuit that contains
$n$ logical components $c_1,\ldots,c_n$ and $k$ lines $l_1, \ldots,
l_k$. As a concrete example, consider the circuit of
Figure~\ref{fig:circuit}.%
\footnote{The ``full adder'' example is often used in the diagnosis
literature. In our discussion here we loosely follow the examples of
Reiter \citeyear{reiter:87}.}
The diagnosis task is to identify which components are faulty.
The agent
can set the values of input lines of the circuit and observe the
output values. The agent then compares the actual output values to the
expected output values and attempts to locate faulty components.

We model this situation using the tools we presented in the
previous sections. We
start by describing the agent's knowledge using a Kripke
structure. We then construct two possible plausibility measures over
worlds in this Kripke structures, and examine the resulting
knowledge and belief.

\paragraph{Knowledge} We model the agent's knowledge about the circuit
using the Kripke structure $M^K_\diag =
(W_\diag,\pi_\diag,\K_\diag)$.  Each possible world $w \in W_\diag$ is
composed of two parts: $\fault(w)$, the {\em failure set\/}---that is,
the set of faulty components in $w$, and $\Dvalue(w)$, the value of
all the lines in the circuit.  We consider only worlds where the
components that are not in the failure sets perform as expected.
For example, in the circuit of Figure~\ref{fig:circuit},
if the AND gate $A_1$ is not faulty, then we require that $l_5$ has
value ``high'' if and only if both $l_1$ and $l_2$ have the value
``high''.  Most accounts of diagnosis assume that there is a logical
theory $\Delta$ that describes the properties of the device.
To capture our intuition, it must be the case that $w$ is a
possible world in $M$ if and only if $\fault(w)$ and $\Dvalue(w)$ are
together consistent with $\Delta$.

The most straightforward language for reasoning about faults is the
following: let $\Phi_\diag = \{ \faulty(c_1),
\ldots, \faulty(c_n), \hi(l_1), \ldots, \hi(l_k) \}$ be the set of
propositions,
where each $\faulty(c_i)$ denotes that component $i$ is faulty and $\hi(l_i)$
denotes that line $i$ in a ``high'' state. We then define the interpretation
$\pi_\diag$ in the obvious way: $\pi_\diag(w)(\faulty(c_i)) =$ {\bf true} if
$c_i \in \fault(w)$, and $\pi_\diag(w)(\hi(l_i)) =$ {\bf true} if $\<l_i,1\>
\in \Dvalue(w)$.

Next, we need to define the
agent's knowledge.  We define $\Dobs{w} \subseteq \Dvalue(w)$ to be the
values of those lines the agent sets or observes.  The agent knows
which tests he has performed and the results he observed.  Therefore,
we have $(w,w') \in
\K_\diag$ if $\Dobs{w} = \Dobs{w'}$.
For example, suppose the agent observes
$\hi(l_1)\land\hi(l_2)\land\hi(l_3)\land \hi(l_7)\land\hi(l_8)$. The
agent then considers possible all worlds where the same observations
hold. Since these observations are consistent with the correct
behavior of the circuit, one of these worlds has an empty failure
set. However, other worlds are possible. For example, it might be that
the AND gate $A_2$ is faulty. This would not affect the outputs in
this case, since if $A_1$ is non-faulty, then its output is ``high'',
and thus, $O_1$'s output is ``high'' regardless of $A_2$'s output.

Now suppose that the agent observes
$\hi(l_1)\land\neg\hi(l_2)\land\hi(l_3)\land\hi(l_7)\land\neg\hi(l_8)$. These
observations imply that the circuit is faulty.
(If $l_1$ and $l_3$ are ``high'' and $l_2$ is ``low'', then the
correct values for
$l_7$ and $l_8$ should be ``low'' and ``high'', respectively.) In this
case there are several possible failure sets, including $\{ X_1 \}$,
$\{X_2, O_1\}$, and $\{X_2,A_2\}$.

In general, there is more than one explanation for the observed faulty
behavior. Thus, the agent can not {\em know\/} exactly which
components are faulty, but he may have {\em beliefs\/} on that score.

\paragraph{Plausibility}
To model the agent's beliefs, we need to decide on the plausibility
measure the agent has at any world. We assume that only
failure sets are relevant for determining a world's
plausibility. Thus, we start by constructing a plausibility measure
over possible failures of the circuit. We assume that failures of
individual components are independent of one another. If we also
assume that the likelihood of each component failing is the same, we
can construct a preference ordering on failure set as follows:
If $f_1$ and $f_2$
are two failure sets, we say that $f_1$ is
preferred to $f_2$
if $\Card{f_1} <
\Card{f_2}$, that is, if $f_1$ consists of fewer faulty components
than $f_2$.
This preference ordering induces a plausibility measure
using the construction of
\Pref{pro:prec}.
In this measure
$\Pl(F_1) <
\Pl(F_2)$ if $\min_{f \in F_1}(\Card{f}) < \min_{f \in
F_2}(\Card{f})$.

We can construct the same plausibility measure based on probabilistic
arguments using PPDs.  Suppose that the probability of a single
component failing is $\epsilon$. Since we have assumed that failures
are independent, it follows that the probability of a failure set $f$
is $\epsilon^{\Card{f}}(1-\epsilon)^{(n-\Card{f})}$, since there are
$\Card{f}$ components that fail, and $n - \Card{f}$ components that do
not fail.  To model the behavior of small but unknown failure
probability, we can consider the PPD $(\Pr_0, \Pr_1, \ldots)$, where
in $\Pr_m$ the probability of a single failure is $1/(m+1)$.  It is
not hard to check that $\lim_{m\rightarrow \infty} \Pr_m(F_2) /
\Pr_m(F_1) = 0$ if and only if $\Pl(F_2) < \Pl(F_1)$ in the
plausibility measure described above. Interestingly, this plausibility
measure is almost identical to the $\kappa$-ranking in which
$\kappa(\{f\}) = |f|$. The only difference is that if $\Card{f_1} =
\Card{f_2}$, $\Pl(\{f_1\})$ is incomparable to $\Pl(\{f_2\})$ in the
plausibility measure we constructed, while they are equal according to
the $\kappa$-ranking.

In some situations it might be unreasonable to assume that all
components have equal failure probability. Thus, we might assume that
for each component $c_i$ there is a probability $\epsilon_i$ of
failure. If we assume independence, then given $\vec\epsilon =
(\epsilon_1,\ldots,\epsilon_n)$, the probability of a failure set $f$
is $\Pi_{c_i\in f}\epsilon_i \,\Pi_{c_i\not\in f}(1 - \epsilon_i)$.
We can construct a PPD that captures the effect of the $\epsilon_i$'s
getting smaller, but at possibly different rates:
Suppose $g$ is a bijection from $\IN^m$ to $\IN$.  If $\vec{m} = (m_1,
\ldots, m_n)$, let $\Pr_{g(\vec{m})}$ be the distribution where the
probability of $c_i$ failing is $1/(m_i+1)$, for $i = 1, \ldots, n$.
In this case, we get that $\lim_{m \rightarrow \infty}
\Pr_m(f_2)/\Pr_m(f_1) = 0$ if
and
only if $f_2$ is a strict subset of
$f_1$, \ie if $f_1$ contains all the components in $f_2$ and more.
Since we do not assume any relations among the failure probabilities
of different components, it is not possible to compare failure sets
unless one is a subset of the other.  Thus, we can define $f \prec f'$
if $f \subset f'$. Using the construction of
\Pref{pro:prec}, we can again consider the plausibility
measure $\Pl$ induced by $\prec$. It is not hard to see that $\Pl(F_1)
\le \Pl(F_2)$ if for every failure set $f_1 \in F_1 - F_2$ there is
some $f_2 \in F_2$ such that $f_2
\prec f_1$. As our construction shows, this plausibility measure can
be induced by either a preference ordering or a PPD; however, it
cannot be captured by a $\kappa$-ranking or a possibility measure,
since the ordering on failure sets is partial.

\paragraph{Beliefs}
We now have the required components to examine the agent's beliefs.
Using the two plausibility measures we just described, we can
construct two possible structures $M_{\diag,1}$ and $M_{\diag,2}$. In
both structures we set $\Omega_{(w,1)} = \K_\diag(w)$, and in both
$M_{\diag,1}$ and $M_{\diag,2}$ the plausibility measure is induced
from a preference ordering on failures (using the construction of
\Pref{pro:prec}).
In $M_{\diag,1}$, we take the plausibility measure to be such
that $\Pl_{(w,1)}(\{w\}) \ge \Pl_{(w,1)}(\{w'\})$ if and only if
$|\fault(w)|
\le |\fault(w')|$, and in $M_{\diag,2}$ so that $\Pl_{(w,1)}(\{w\}) \ge
\Pl_{(w,1)}(\{w'\})$ if and only if $\fault(w) \subseteq \fault(w')$.
It is easy to see that, in both structures, if there is a world $w$ in
which these observations occur and where $\fault(w) = \emptyset$, then
the agent believes that the circuit is faultless. If the agent detects
an error, he believes that it is caused by one of the {\em minimal
explanations\/} of his observations, where the notion of minimality
differs in the two structures.  We now make this statement more
precise. Let $f$ be a failure set.
Let $\Diag{f}$ be the formula that
denotes that $f$ is the failure set, so that $(M,w) \sat \Diag{f}$ if
and only if $\fault(w) = f$.  The agent believes that $f$ is a
possible {\em diagnosis\/} (\ie an explanation of his observations) if
$\neg\Bel_1\neg\Diag{f}$.  The set of diagnoses the agent considers
possible is $\BEL(M,w) = \{ f : (M,w) \sat \neg\Bel_1\neg\Diag{f} \}$.
We say that a failure set $f$ is {\em consistent\/} with an
observation $o$ if it is possible to observe $o$ when $f$ occurs, \ie
if there is a world $w$ in $W$ such that $\fault(w) = f$ and $\Dobs{w}
= o$.%
\footnote{
Note that if $\Delta$ is a theory that describes the properties of
circuit, then a failure $f$ is consistent with observation $o$, if
and only if $f$ and $o$ are consistent according to $\Delta$. }

\Pro
\begin{enumerate}\denselist
 \item[(a)] $\BEL(M_{\diag,1},w)$ contains all failure sets $f$ that are
    consistent with $\Dobs{w}$ such that there is no failure set $f'$
    with $|f'| < |f|$ which is consistent with $\Dobs{w}$.
 \item[(b)] $\BEL(M_{\diag,2},w)$ contains all failure sets $f$ that are
    consistent with $\Dobs{w}$ such that there is no failure set $f'$
    with $f' \subset f$ which is consistent with $\Dobs{w}$.
\end{enumerate}
\ePro
\prf
Straightforward; left to the reader.
\eprf

Thus, both $\BEL(M_{\diag,1},w)$ and $\BEL(M_{\diag,2},w)$ consist of
minimal sets of failure sets consistent with $\Dobs{w}$, for different
notions of minimality.  In the case of $M_{\diag,1}$, ``minimality''
means ``of minimal cardinality'', while in the case of $M_{\diag,2}$,
it means ``minimal in terms of set containment''.  This proposition
shows that $M_{\diag,1}$ and $M_{\diag,2}$ capture standard
assumptions made in model-based diagnosis; $M_{\diag,1}$ captures the
assumptions made in \cite{dekleer}, while $M_{\diag,2}$ captures the
assumptions made in \cite{reiter:87}. More concretely, in our example,
if the agent observes
$\hi(l_1)\land\neg\hi(l_2)\land\hi(l_3)\land\hi(l_7)\land\neg\hi(l_8)$,
then in $M_{\diag,1}$ she would believe that $X_1$ is faulty, since
$\{X_1\}$ is the only diagnosis with cardinality one. On the other
hand, in $M_{\diag,2}$ she would believe that one of the three minimal
diagnoses occurred: $\{X_1\}$, $\{X_2,O_1\}$ or $\{X_2,A_2\}$.

\subsection{Properties of Knowledge and Plausibility}
Kripke structures for knowledge and plausibility are quite similar
to the Kripke structures for knowledge and probability introduced by
    Fagin and Halpern \citeyear{FH3}.  The only difference is that in
    Kripke structures for
knowledge and probability, $\Plass_i(w)$ is a probability space rather than
a plausibility space.  Fagin and Halpern explore various natural
    restrictions on the interactions between the probability spaces
    $\Plass_i(w)$ and the accessibility relations $\K_i$.
    Here we investigate restrictions on the interaction
    between the plausibility spaces and the accessibility relations.
    Not surprisingly, some of these conditions are exact analogues to
    conditions investigated by Fagin and Halpern.
Given our interest in the KLM properties, we will be interested
in structures that satisfy the following condition:
\begin{cond}{QUAL}
$\Plass_i(w)$ is qualitative
for all worlds $w$
    and agents $i$.
\end{cond}

The same arguments that show that A2 gives us
the AND rule also show that it gives us property K2 for beliefs.  More
precisely, we have the following result.

\Thm\label{QUALthm}
If $M$ satisfies QUAL, then for all worlds $w$ in $M$, we have
\begin{enumerate}
\item[(a)] $(M,w) \sat ((\sigma \Cond_i \phi) \land (\sigma \Cond_i
\psi))
\rimp (\sigma \Cond_i (\phi \land \psi))$
\item[(b)] $(M,w) \sat \Bel_i\phi \land \Bel_i \psi\rimp
 \Bel_i(\phi
\land \psi)$
\item[(c)] $(M,w) \sat \Bel_i\phi \land \Bel_i(\phi \land \psi) \rimp
\Bel_i\psi$.
\end{enumerate}
\eThm
\prf
Straightforward; left to the reader.
\eprf

In view of this result, we typically assume that QUAL holds whenever we
want to reason about belief.

The set $\Omega_{(w,i)}$ consists of all worlds to which agent $i$
    assigns some degree of plausibility in world $w$.
We would not expect the agent to place a positive probability on
    worlds that he considers impossible.
Similarly, he would not want
    to consider as plausible (even remotely) a world he knows to be
    impossible.
This intuition leads us to the following condition,
    called CONS for {\em consistency\/} (following \cite{FH3}):
\begin{cond}{CONS}
$\Omega_{(w,i)} \subseteq \K_i(w)$ for all worlds $w$
and all agents $i$.%
\footnote{We remark that CONS is inappropriate if we use
$\Cond$ to model, not plausibility, but counterfactual conditions, as
is done by Lewis \citeyear{Lewis73}.  If CONS holds, then it is easy to
    see that $K_i \phi \rimp K_i(\neg \phi \Condi \psi)$ is valid, for
    all $\psi$.  That is, if agent~$i$ knows $\phi$, then he knows
    that in the most plausible worlds where $\neg \phi$ is true,
    $\psi$ is vacuously true, because there are no plausible worlds
    where $\neg\phi$ is true.  On the other hand, under the
    counterfactual reading, it makes perfect sense to say ``I know the
    match is dry, but it is not the case that if it were wet, then it
    would light if it were struck.''
}
\end{cond}

A consequence of assuming CONS is a stronger
    connection between knowledge and belief. Since
    CONS implies that the most plausible worlds are in
$\K_i(w)$, it follows that if the agent knows $\phi$ he also believes
    $\phi$.  (Indeed, as we shall see, this condition characterizes
    CONS.)

In probability theory, the agent assigns probability 1 to the set of all
worlds.  Since $1 > 0$, this means the agent assigns non-zero
probability to some sets of worlds.  It is possible to have
$\top = \bot$
in plausibility spaces.  If this happens, the agent considers all
sets to be completely implausible.
The following condition, called NORM for {\em normality\/} (following
\cite{Lewis73}),
says this does not happen:
 \begin{cond}{NORM}
$\Plass(w,i)$ is normal, that is,
$\top_{(w,i)} > \bottom_{(w,i)}$, for all worlds $w$
and all agents $i$.
\end{cond}

We can strengthen this condition somewhat to one that says that
the agent never considers the real world implausible.
This suggests
    the following condition:  $\Pl_{(w,i)}(\{w\}) >
    \bottom$. Stating this condition, however, leads to a technical
    problem. Recall that $\Pl_{(w,i)}$
    is defined over the set of measurable subsets of
    $\Omega_{(w,i)}$.  In general, however, singletons may not be
    measurable. Thus, we examine a
    slightly weaker condition which we call REF for {\em
    reflexive\/} (following \cite{Lewis73}):
\begin{cond}{REF}
For all worlds $w$
and all agents $i$,
\begin{itemize}\denselist
 \item  $w \in \Omega_{(w,i)}$, and
 \item $\Pl_{(w,i)}(A) >
    \bottom$  for all $A \in
    \F_{(w,i)}$ such that $w \in A$.
\end{itemize}
\end{cond}

As we said in the introduction, much of the previous work using
    conditionals assumed (implicitly or explicitly) that the
agent considers only one plausibility measure possible. This amounts
    to assuming
    that the plausibility measure is a function of the agent's
    epistemic state.
This is captured by an assumption
called SDP (following  \cite{FH3})
 for {\em state determined plausibilities\/}:
\begin{cond}{SDP}
For all worlds $w$ and $w'$
and all agents $i$,
if $(w,w') \in \K_i$ then $\Plass_i(w) =
    \Plass_i(w')$.
\end{cond}
It is easy to see that SDP implies that an
    agent knows his plausibility measure.  In particular,
    as we  shall see, with SDP we have that  $\phi \Condi \psi$
    implies $K_i (\phi \Condi \psi)$.

It is easy to verify that the structures described in the diagnosis
example of Section~\ref{sec:xam-diag} satisfy CONS, REF, and SDP.  As
mentioned in the introduction, SDP is not appropriate in all
situations; at times we may want to allow the agent to consider
possible several plausibility measures. To capture this, we need to
generalize SDP.  The following example might help motivate the formal
definition.
\xam
\label{xam:Alice}
This is a variation of the Liar's Paradox. On a small Pacific island
    there are two tribes, the Rightfeet and the Leftfeet.
    The Rightfeet
    are known to usually tell the truth, while the Leftfeet are known to
     usually lie. Alice is a visitor to the island. She encounters
    a native, Bob, and discusses with him various aspects of life on the
    island. Now, Alice does not know to what tribe Bob belongs.
     Thus, she considers it possible both that Bob is a Rightfoot and
     that he is a Leftfoot. In the first case, she should believe
     what he tells her and in the second she should be skeptical.

One possible way of capturing this situation is
by partitioning the worlds Alice considers possible
into two sets, according to Bob's tribe.
Let $W_R$ (\resp $W_L$) be the set of worlds that Alice considers
possible where Bob is a Rightfoot (\resp Leftfoot).  As the discussion
above suggests, Alice's plausibility measure at the worlds of $W_R$
gives greater plausibility to worlds where Bob is telling the truth than
to worlds where Bob is lying; the opposite situation holds at worlds of
$W_L$.
In such a structure, the formula
    $\neg\Know_{Alice} \neg (tell(\phi) \Cond_{Alice}\; \neg \phi) \land
    \neg\Know_{Alice}\neg(tell(\phi) \Cond_{Alice}\; \phi)$ is
    satisfiable,
where $tell(\phi)$ is the formula that holds when Bob tells Alice
$\phi$.
On the other hand,
    in structures satisfying SDP, this formula is satisfiable only when
$tell(\phi)$ has plausibility $\bottom$ in all the worlds that Alice
considers possible.
\exam

While this example may seem contrived, in many situations
    it is possible to extract parameters
such as Leftfoot and Rightfoot
    that determine which
    conditional statements are true. For example, when we introduce
    time into the picture (in Section~\ref{systems}), these parameters
    might be the agent's own actions in the future. Such a
partition allows us
to make statements such as ``I do not know whether
    $\phi$ is plausible or not, but I know that if I do $a$, then
    $\phi$ is plausible'', where $\phi$ is some statement about the
    future.
If the agent does not know the value of these parameters, she will not
necessarily know which conditionals are true at a given world (as
was the case in the example above).

Example~\ref{xam:Alice} motivates the condition called
{\em uniformity}.
\begin{cond}{UNIF}
For all worlds $w$ and agents $i$, if $w' \in \Omega_{(w,i)}$ then
$\Plass_i(w) = \Plass_i(w')$.%
\footnote{This condition is not the same as uniformity
    as defined in \cite{Lewis73}; rather, it corresponds in the Lewis
    terminology to absoluteness.}
\end{cond}
It is not hard to show that
UNIF holds if and only if, for each agent $i$, we can partition
the set of possible
worlds in such a way that for each cell $C$ in the partition, there is a
plausibility space $(W_C,\Pl_C)$ such that $W_C \subseteq C$ and
$\Plass_i(w) = (W_C,\Pl_C)$ for all worlds $w \in C$.
Moreover, if CONS also holds, then this partition refines the
    partition induced by the agent's knowledge, \ie if $C$ is a cell
    in the partition and $w$ is some world $C$, then $C \subseteq \K_i(w)$.
It easily follows that SDP and CONS together imply UNIF.

When we model uncertainty about the relative plausibility of different
     worlds this way it is reasonable
     to demand that
the plausibility measure totally orders all events; \ie it is a {\em
ranking}.
The RANK assumption is:
 \begin{cond}{RANK}
 For all worlds $w$ and agents $i$, $\Plass_i(w)$ is a ranking, that is,
for all sets $A, B \subseteq W_w$
either $\Pl_w(A) \le \Pl_w(B)$ or $\Pl_w(B) \le \Pl_w(A)$, and
$\Pl_w(A\union B) = \max(\Pl_w(A),\Pl_w(B))$.
 \end{cond}

Note that $\kappa$-rankings and possibility measures are two examples
of rankings. Additionally,
{\em rational}
preference orderings of
\cite{KLM} are essentially rankings in the sense that for each rational
preference ordering we can construct a ranking that satisfies exactly
the same conditional statements \cite{FrThesis,FrH5Full}.

While rankings are quite natural, they have often been
rejected as being too inexpressive \cite{ginsberg:86}.
In a
ranking there is a total order on events.
The standard argument for partial orders is as follows: In general, an
agent may not be able to determine the relative plausibility of $a$
and $b$.  If the plausibility measure is a ranking, the agent is forced
to make this determination; with a partial order, he is not. This
argument loses much of its force in our framework, once we combine
knowledge and plausibility.  As we said above, the agent's ignorance
can be modeled by allowing him to consider (at least) two
rankings
possible, one in which $a$ is more plausible than $b$, and one in
which $b$ is more plausible that $a$.  The agent then believes neither
that $a$ is more plausible than $b$ nor that $b$ is more plausible
than $a$.

\subsection{Knowledge and Belief}
\label{kbsec}
\label{KBSEC}

How reasonable is the notion of belief we have defined?  In this
    section, we compare it to other notions considered in the
    literature.

Recall that $\L^{B}$ be the language where
    the only modal operators are $B_1,
    \ldots, B_n$. Let $\L^{KB}$ be the language where we have
    $\Know_1, \ldots, \Know_n$ and $\Bel_1, \ldots, \Bel_n$ (but no $\Condi$
    operators).
It is not hard to see (and will follow from our proofs below)
that to get belief to satisfy even minimal
such as K2, we need the AND rule to hold.  Thus, in this section, we
restrict attention to Kripke structures for knowledge and plausibility
that satisfy QUAL.  We then want to investigate the impact of adding
additional assumptions.
Let $\M$ be the set of all Kripke structures
    for knowledge and plausibility
    that satisfy QUAL,
    and let $\M^{\sCONS}$ (\resp $\M^{\sCONS,\sNORM}$) be the structures
    satisfying QUAL and CONS (\resp QUAL, CONS and NORM).

Work on belief and knowledge in the literature
\cite{HM2,Hi1,Lev2} has focused on the modal systems S5, KD45, K45,
    and K with semantics based on Kripke structures as described in
    \Sref{knowledge}.
Before we examine the properties of belief in our approach, we relate our
    semantics of belief (in terms of plausibility)
to the more standard Kripke approach, which presumes that belief is
defined in terms of a binary relation $\B_i$.
Can we define a relation $\B_i$ in
    terms of $\K_i$ and $\Plass_i$ such that $(M,w) \sat \Bel_i\phi$ if
    and only if $(M,v) \sat \phi$ for all $v \in \B_i(w)$?
We show that this is possible in some structures, but not in general.

Let $S = (W,\Pl)$ be a qualitative plausibility space. We say that $A
    \subseteq W$ is a set of {\em most plausible\/} worlds if $\Pl(A) >
    \Pl( \overline{A})$ (where $\overline{A}$ is the complement of
    $A$, \ie $W-A$) and for all $B \subset A$, $\Pl(B) \not>
    \Pl(\overline{B})$.
    That is, $A$ is a minimal set of worlds that is more plausible
    than its complement. It is easy to verify that if such a set
    exists, then it must be unique. To see this, suppose that
    $A$ and $A'$ are both sets of most plausible worlds. We now show
    that $\Pl(A \inter A') > \Pl(\overline{A \inter A'})$.
Since $A$ and $A'$ are both most plausible sets of worlds,
this will show
that we must have $A = A'$.
To see that $\Pl(A \inter A') > \Pl(\overline{A \inter A'})$,
first note that $A \inter A'$, $A -
    A'$ and $\overline{A}$ are pairwise disjoint.
Since $A$ and $A'$ are most plausible sets of worlds, we have that
$\Pl((A \inter A') \union (A - A')) = \Pl(A) >
    \Pl(\overline{A})$ and $\Pl((A \inter A') \union \overline{A}) \ge \Pl((A
    \inter A') \union (A' - A)) = \Pl(A') > \Pl(\overline{A'}) \ge
    \Pl(A - A')$. We can
    apply A2 to get that $\Pl(A \inter A') > \Pl((A - A') \union
    \overline{A}) = \Pl(\overline{A \inter A'})$.

In finite plausibility structures (that is, ones with only finitely many
worlds), it is easy to see that there is always a (unique) set of most
plausible worlds.
In general,
    however, a set of most plausible worlds does not necessarily exist.
    For example, consider the space $S_0 = (W,\Pl)$, where $W = \{ w_i :
    i \ge 0 \}$
    and  $\Pl$ is
defined
as follows: $\Pl( A ) = \infty$ if $A$ contains an infinite
    number of worlds, and $\Pl(A) = \max_{w_i \in A}(i)$ otherwise.
    Suppose that $\Pl(A) > \Pl(\overline{A})$.
$\overline{A}$ must be finite, for otherwise
    $\Pl(\overline{ A}) = \infty$. Thus, $A$ must be infinite.
    Suppose $w_i \in A$. It is easy to see that $A - \{w_i\}$ is
infinite
    and $\overline{A - \{w_i\}}$ is finite. Thus, $\Pl(A - \{ w_i \}) >
    \Pl(\overline{A - \{w_i\}})$. This shows
    that there does not exist a set of most plausible worlds in $S$.

If there is no set of most plausible worlds, then we may not be able to
find a relation $\B_i$ that characterizes agent $i$'s beliefs.  For
example, consider the structure $M =
    (W,\pi,\K_1,\Plass_1)$, where $W = \{ w_i : i \ge
    0 \}$ is the set of worlds described in $S_0$ above; $\pi$ assigns
    truth values to primitive propositions
$p_1,p_2,\ldots$ in such a way that
    $\pi(w_i)(p_j) = $ {\bf true} if and only if $j \ge i$; $\K_1$ is
    the complete accessibility relation $\K_1 = W \cross W$; and
    $\Plass_1(w_i)$ is the space $S_0$ described above. It is not hard
    to verify that
    $(M,w_0) \sat \Bel_1\phi$ if and only if
    $\intension{\neg\phi}_{(w_0,i)}$ is a finite set, \ie there is an
    index $i$ such that for all $j \ge i$, we have $(M,w_j) \sat \phi$.
Thus,
$(M,w_0)
    \sat \Bel_1 p_j$ for all $j \ge 0$. Yet there are no worlds in the model
    that satisfy all the propositions $p_j$ at once. Thus, there is
no accessibility relation $\B_1$ that characterizes agent 1's beliefs in
$w_0$.

On the other hand, we can show that if there is always a set of most
plausible worlds, then we can characterize the agents' beliefs by an
accessibility relation.
Let $S = (W,\Pl)$ be a plausibility space.
Define $\MP(S)$ to be the set of most plausible
worlds
    in $S$ if it exists, and  $\emptyset$ if $\Pl(W) =\, \bottom$.
    Otherwise $\MP(S)$  is not defined.
\Pro
Let $M$ be a
Kripke structure
for knowledge and plausibility. If
$\MP(\Plass_i(w'))$
    is defined for all $w' \in \K_i(w)$,
    then $(M,w) \sat \Bel_i\phi$ if and only if $(M,w'') \sat
    \phi$ for all $w'' \in \union_{w' \in \K_i(w)} \MP(\Plass_i(w'))$.
\ePro
\prf
Straightforward; left to the reader.
\eprf

This proposition implies that,
if most plausible sets of worlds always exist in $M$, then we can
set $\B_i(w) =
\union_{w' \in
    \K_i(w)}\MP(\Plass_i(w'))$
and recover the usual Kripke-style semantics for belief.

This discussion shows that our model of belief is more general than
    the classical Kripke-structure account of beliefs, since there are
    models where the agent's beliefs are not determined by a set of
    accessible worlds. However, as we shall see, this does not lead to
    new properties of beliefs in $\L^B$. Roughly speaking, this is because we
    have a finite model property: a formula in $\L^B$
    is satisfiable if and only if it is satisfiable in a finite
    model (see \Tref{thm:small-model} below). It is easy to verify
    that in a finite model
    $\MP(\Plass_i(w))$ is  always defined.
We note, however, that this finite model property is no longer true
when we consider the interaction of beliefs with other modalities,
such as time, or when we examine the first-order case. In these
situations, the two models of beliefs are not equivalent.
Plausibility is strictly more expressive;
see \cite{FrHK1}.

We now examine the formal properties of belief and knowledge in
    structures of knowledge and plausibility.
We start by restricting our attention to $\L^{B}$. As we show below,
    the modal system K precisely characterizes the valid formulas of
    $\L^{B}$ in the class $\M$.
However, in the literature, belief has typically been taken to be
    characterized by the modal system K45 or KD45, not K.
We get K45 by restricting to models that satisfy CONS, and KD45 by
    further restricting to models that satisfy NORM.  Thus, the two
    requirements that are most natural, at least if we have a
    probabilistic intuition for plausibility, are already enough to
    make $\Bel_i$ a KD45 operator.
\Thm
\label{thm:LB}
K (\respc K45, KD45) is a sound and complete axiomatization
for $\L^B$ with respect to $\M$ (\respc $\M^{\sCONS}$, $\M^{\sCONS,\sNORM}$).
\eThm
\prf
See Appendix~\ref{prf:kbsec}.
\eprf

We now consider knowledge and belief together. This
    combination has been investigated in the literature
    \cite{KrausL,Voorbraak}. In particular, Kraus and Lehmann
\citeyear{KrausL}
    define Kripke structures for knowledge and belief that have two
accessibility relations, one characterizing the worlds that are
knowledge-accessible and one characterizing worlds that are
belief-accessible. $\Know_i$ and $\Bel_i$ are defined, as usual, in
terms of these relations. They argue that the two accessibility
relations must be coherent in the sense that the agent knows what she
believes and believes what she knows to be true.
Kraus and Lehmann describe restrictions on the interaction
between the two relations that force this coherence. They show that in
the resulting structures, the interactions between knowledge and
belief are characterized by the following axioms.
\begin{description}\denselist
 \item[KB1.] $\Bel_i\phi \rimp \Know_i\Bel_i\phi$
 \item[KB2.] $\Know_i\phi \rimp \Bel_i\phi$
\end{description}
It turns out that KB1 holds in $\M$ and KB2 is a consequence of CONS.
To see this, recall that $\Bel_i\phi \equiv \Know_i(\True\Cond\phi)$.
    Using positive introspection for knowledge (axiom K4), we derive
    that $\Bel_i\phi\rimp
    \Know_i\Know_i(\True\Cond\phi)$.  This is equivalent to axiom KB1.
When
    $M$ satisfies CONS, we have that $\Omega_{(w,i)} \subseteq
    \K_i(w)$. If $(M,w) \sat \Know_i\phi$, then all worlds in $\K_i(w)$
    satisfy $\phi$. This implies that there are no worlds satisfying
    $\neg\phi$  in $\Omega_{(w,i)}$, and thus $\Bel_i\phi$ must hold.
Thus, KB2 must hold.

We now state this formally. Let $\AX^{\sKB}$
consist of
the S5 axioms for the
    operators $K_i$, the K axioms for the operators $B_i$, together
    with KB1; let $\AX^{\sKB,\sCONS}$ consist of $\AX^{\sKB}$ together
with the K4
    and K5 axioms for $\Bel_i$ and KB2; and let $\AX^{\sKB,\sCONS,\sNORM}$
    consist of $\AX^{\sKB,\sCONS}$ together with the K6 axiom for
$\Bel_i$.
\Thm\label{thm:LKB}
$\AX^{\sKB}$ (\respc $\AX^{\sKB,\sCONS}$, $\AX^{\sKB,\sCONS,\sNORM}$)
    is a sound and
    complete axiomatization of $\L^{KB}$ with respect to $\M$ (\respc
    $\M^{\sCONS}$, $\M^{\sCONS,\sNORM}$).
\eThm
\prf
See Appendix~\ref{prf:kbsec}.
\eprf

As an immediate corollary, we get
that there is a close relationship between our framework
    and that of \cite{KrausL}. Let $KL$ be the logic of Kraus and Lehmann:
\Cor
 For any $\phi \in \L^{KB}$, $KL \sat \phi$ if and only if $\M^{\sCONS,\sNORM}
     \sat \phi$.
\eCor

We now relate to three other notions of beliefs in the
literature---those of
    Moses and Shoham \citeyear{MosesShoham}, Voorbraak \citeyear{Voorbraak},
    and Lamarre and Shoham \citeyear{LamarreShoham}.

Moses and Shoham \citeyear{MosesShoham} also view belief as being derived
 from knowledge. The intuition that they try to capture is that once
    the agent makes a
    defeasible assumption, the rest of his beliefs should follow from
    his knowledge. In this sense, Moses and Shoham can be viewed as
    focusing on the
    implications of an assumption and not on how it was
    obtained.
We can understand their notion as saying that
 $\phi$ is believed if it is known to be true in the most plausible
  worlds.  But for them, plausibility is not defined by an ordering.
Rather, it is defined in terms of a formula, which can be thought of as
  characterizing the most plausible worlds.
  More formally, for a fixed formula $\alpha$, they
  define $\Bel_i^\alpha\phi$ to be an abbreviation for $K_i (\alpha \rimp
  \phi$).%
  \footnote{Shoham and Moses also
  examine two variants of this definition. These mainly deal with the
    cases where $\alpha$ is inconsistent with the agent's knowledge.
For simplicity,
    we assume here that $\alpha$ is consistent with the agent's knowledge.}
The following result relates our notion of belief to that of Moses and
    Shoham.
\Lem\label{lem:MosesShoham}
 Let $M$ be a propositional Kripke structure of knowledge and
    plausibility satisfying CONS and SDP.
Suppose that $w$, $i$, and $\alpha$ are such that the most plausible
    worlds in $\Plass_i(w)$ are exactly those worlds in $\K_i(w)$
    that satisfy $\alpha$, \ie
    $\MP(\Plass_i(w)) = \{ w' \in \K_i(w) : (M,w') \sat \alpha \}$.
Then for any formula $\phi\in\L^{KB}$ that includes only the
    modalities $\Know_i$ and $\Bel_i$, $(M,w)
     \models \phi$ if and only if $(M,w) \models
     \phi^*$, where $\phi^*$ is the result of recursively replacing each
    subformula of the form
    $\Bel_i\psi$ in $\phi$ by $\Know_i(\alpha\rimp\psi^*)$.
\eLem
\prf
See Appendix~\ref{prf:kbsec}.
\eprf

Voorbraak \citeyear{Voorbraak} distinguishes two notions of knowledge: {\em
    objective knowledge\/} and {\em true justified belief\/}. He then
    studies the
    interaction of both notions of knowledge with beliefs. The
    intuition we assign to knowledge is similar to Voorbraak's
    intuition for objective knowledge. However, Voorbraak
    objects to the axiom $\Know_i\phi\rimp\Bel_i\phi$, and suggests
    $\Bel_i\phi\rimp\Bel_i\Know_i\phi$.
The difference lies in the interpretation of belief.
Voorbraak's notion of belief is stronger than ours.  His view is that
the agent cannot distinguish what he believes from what he knows
(indeed, he believes that what he believes is the same as what he knows).
Our notion of belief is weaker, in that we allow agents to be aware of
the defeasibility of their beliefs.

Lamarre and Shoham \citeyear{LamarreShoham} investigate the notion of
    knowledge as justified true belief using a framework that is very
    similar to ours. They start with an explicit preference ordering
    over possible worlds, and then define $\Bel^\alpha\phi$ to
    read ``given evidence $\alpha$, $\phi$ holds in the most plausible
    $\alpha$-worlds''. Their formal account of $\Bel^\alpha\phi$ is
    exactly $\alpha\Condi\phi$ in our notation.
    Unlike us, they examine a notion of knowledge as
    ``belief stable under incorporation of correct facts'', which is
    rather different then our notion of objective knowledge. Thus,
    while the technical construction is similar, the resulting
    framework is substantially different.
Lamarre and Shoham take plausibility to be the only primitive, and use
it to determine both knowledge and belief.
We take
both
knowledge and plausibility to be primitive, and use them to
define belief.

\subsection{Axiomatizing the Language of Knowledge and
Plausibility}\label{axiomfull}
\label{sec:axiomfull}
\label{SEC:AXIOMFULL}

Up to now, we have considered just the restricted language $\L^{KB}$.
We now present sound and complete axiomatizations for the full language
    $\L^{KC}$.
The technical details are much in the spirit of
the axiomatizations presented in \cite{FH3} for knowledge
and probability.
Our complete axiomatization for $\M$
consists of two ``modules'':
a complete axiomatization for knowledge (\ie S5) and
a complete axiomatization for conditionals.
In the general case,  there are no axioms connecting knowledge and
    plausibility.
For each of the conditions we consider, we provide an axiom that
characterizes it.
The axioms characterizing NORM, REF, RANK, and UNIF are taken from
    \cite{Lewis73}
    and \cite{Burgess81}
(see also \cite{FrThesis,FrH5Full}),
while the
axioms for  CONS and SDP (and also UNIF) correspond directly to the
    axioms suggested in \cite{FH3} for their probabilistic counterparts.
We also provide complete characterizations of
the complexity of the validity problem for all the logics considered,
    based on complexity results for knowledge \cite{HM2} and for
    conditionals \cite{FrH3Full}.

The axiom system can be modularized into three
    components: propositional reasoning, reasoning about knowledge,
    and reasoning about conditionals.
The component for propositional reasoning consists of K1 and RK1
(from Section~\ref{knowledge}); the component for reasoning about
knowledge consists of K2--K5 and RK2 (from Section~\ref{knowledge});
the component for reasoning about conditionals consists of the
    standard axioms and rules for conditional logic
C1--C4, RC1, and RC2 described in
\cite{FrThesis,FrH5Full} following \cite{Burgess81,Lewis73}:
\begin{description}\denselist
 \item[C1.] $\phi \Cond \phi$
 \item[C2.] $((\phi \Cond \psi_1)\land(\phi\Cond\psi_2)) \rimp
    (\phi\Cond(\psi_1\land\psi_2))$
 \item[C3.] $((\phi_1\Cond\psi)\land(\phi_2\Cond\psi)) \rimp
    ((\phi_1\lor\phi_2) \Cond\psi)$
 \item[C4.] $((\phi_1\Cond\phi_2)\land(\phi_1\Cond\psi)) \rimp
    ((\phi_1\land\phi_2) \Cond \psi)$
 \item[R1.] From $\phi$ and $\phi \rimp \psi$ infer $\psi$
 \item[RC1.] From $\phi \dimp \phi'$ infer $(\phi\Cond\psi) \rimp
(\phi'\Cond\psi)$
 \item[RC2.] From $\psi \rimp \psi'$ infer $(\phi\Cond\psi) \rimp
(\phi\Cond\psi')$
\end{description}

Let \AX\ consist of K1--K5, C1--C4, RK1, RK2, RC1, and RC2.
\Thm
\label{thm:ax}
\AX\ is a sound and complete axiomatization for $\L^{KC}$
    with respect to $\M$.
\eThm
\prf
See  Appendix~\ref{prf:axiomfull}.
\eprf

We now capture the conditions described
above---CONS, NORM, REF, SDP, UNIF, and RANK---%
    axiomatically.
\commentout{
RANK, NORM, REF, and UNIF correspond the axioms C5--C8, respectively,
described in \Sref{conditionallogic}.
}
RANK, NORM, REF, and UNIF correspond the axioms C5--C8, respectively,
from \cite{FrThesis,FrH5Full}:
\begin{description}\denselist
 \item[C5.] $\phi\Cond\psi \land \neg(\phi\Cond\neg\xi) \rimp
 \phi\land\xi \Cond \psi$
 \item[C6.] $\neg(\true\Cond\false)$.
 \item[C7.] $\PBox\phi \Cond \phi$
 \item[C8.] $[(\phi \Cond \psi) \rimp \PBox(\phi \Cond \psi)] \land
            [\neg(\phi \Cond \psi) \rimp \PBox \neg (\phi \Cond
\psi)]$
\end{description}
CONS and SDP correspond to the
following axioms, respectively;
\begin{description}\denselist
 \item[C9.] $\Know_i\phi \rimp \PBox_i\phi$
 \item[C10.] $(\phi\Condi\psi) \rimp \Know_i(\phi\Condi\psi)$
\end{description}

It is interesting to note that the axioms for CONS and UNIF are
   derived from  the axioms defined in \cite{FH3} by
    replacing $w(\phi) = 1$ (the probability of $\phi$ is 1) by
    $\PBox_i\phi$, which has a
similar reading.
We show
that adding the appropriate axioms to AX gives a
    sound and complete axiomatization of the logic with respect to the
    class of structures satisfying the corresponding conditions.
\Thm\label{thm:ax-extend}
Let $\A$ be a subset of $\{ \RANK, \NORM, \REF, \UNIF, \CONS, \SDP \}$
    and let $A$
    be the corresponding subset of $\{$C5, C6, C7, C8, C9, C10$\}$. Then
    $\mbox{AX} \cup A$ is a sound and
    complete axiomatization with respect to the structures in $\M$
    satisfying $\A$.
\eThm
\prf
See Appendix~\ref{prf:axiomfull}.
\eprf

We now consider the complexity of the validity problem.  Our results
are based on a combination of results for complexity of epistemic
logics \cite{HM2} and conditional logics \cite{FrH3Full}. Again, the
technical details are much in the spirit of those in \cite{FH3}.
\commentout{
 We start with an overview of the complexity-theoretic
    notions we need.  For a more detailed treatment of the topic, see
    \cite{GarJoh,HU}.

    Complexity theory examines the difficulty of
    determining membership in a set as a function of the input size.
    In our case we check if a formula
    $\phi$ is in the set of valid formulas. Difficulty is
    measured in terms of the time or space required to decide if a
    formula $\phi$ is valid as a function of
$|\phi|$, the length of the formula. The complexity
    classes we are interested in are NP, PSPACE, and EXPTIME.
    These
    classes consist of these sets for which deciding membership can be done in
    nondeterministic polynomial time, polynomial space,
    and exponential time,
    respectively. We are also interested in the complementary class
    co-NP, which consists of sets for which deciding non-membership can be done
    in nondeterministic polynomial time. For example, checking
    validity of $\phi$ is in co-NP, if checking non-validity of $\phi$
    (\ie checking the satisfiability of $\neg\phi$) is in NP. (Note
    that PSPACE and EXPTIME are  the complement of themselves, \eg
    co-PSPACE $=$ PSPACE.)

To show that a set is in a complexity class we usually describe a
    procedure that determines membership in the set and conforms to the
    time or space restriction of the class. Usually, we also want to
    show that a set is not in an easier class. To do this we show that the
    set is {\em hard\/} in the class. Roughly speaking, A set $A$ is
    hard in a class $\C$ if
    for every set $B \in \C$, an algorithm deciding membership in $B$
    can be effectively obtained from an algorithm deciding membership in $A$.
    A set is {\em complete\/} with respect to a complexity class $\C$
    if it is both in $\C$ and $\C$-hard.
}

We start with few results that will be useful in our discussion of
complexity.  As is often the case in modal logics, we can prove a
``small model property'' for our logic: if a formula is satisfiable at
all, it is satisfiable in a small model. Let $\Sub(\phi)$ be the set
of subformulas in $\phi$. It is easy to see that an upper bound on
$\Card{\Sub(\phi)}$ is the number of symbols in $\phi$.

\Thm\label{thm:small-model}
Let $\A$ be a subset of $\{ \CONS, \NORM, \REF, \SDP, \UNIF, \RANK
\}$. The formula $\phi$ is satisfiable in a Kripke structure
satisfying $\A$ if and only if it is satisfiable in a Kripke structure
with at most $2^{\Card{\sSub(\phi)}}$ worlds.
\eThm
\prf
See  Appendix~\ref{prf:axiomfull}.
\eprf

This shows that if $\phi$ is satisfiable, then it is satisfiable in a
model with at most exponential number of worlds. Such a ``small
model'' result is useful when we consider upper bound on the
complexity of checking satisfiable. Roughly speaking, if there is a
small model, then we can construct this model in time, say,
exponential in the size of the formula. However, there is one problem
with the result we have just proved.  This ``small'' number of worlds
does not necessarily mean that we can compactly describe the Kripke
structure. Recall that $\Pl_{(w,i)}$ describes an ordering over
subsets of $\Omega_{(w,i)}$. Thus, in the worst case, we need to
describe an ordering on $2^{\Card{\Omega_{(w,i)}}}$ sets of
worlds. Thus, the representation of a structure might be exponential
in the number of worlds. Fortunately, we can show that a satisfiable
formula is satisfiable in a small model with a compact
representation.

We start with a definition. We say that $M =
(W,\pi,\K_1,\ldots, \K_n,
\Plass_1,\ldots, \Plass_n)$ is a {\em preferential (Kripke)
structure\/} if for each $\Plass_i(w)$, there is a preference
ordering $\prec_{(w,i)}$ on $\Omega_{(w,i)}$ that induces
$\Pl_{(w,i)}$ using the construction of
\Pref{pro:prec}.
Recall that a preference ordering is a binary relation on the set of
possible worlds. Thus,
if $W$ is finite, we can describe the relations $\K_i$ and
the preference orderings $\le_{(w,i)}$ using tables of size at most
$\Card{W}^2$. So the representation of such
structures is polynomial in $|W|$.  Is it possible to find a small
preferential Kripke structure satisfying $\phi$? Indeed we can. Using
results of \cite{FrH3Full}, we immediately get the following lemma:

\Lem\label{lem:small-pref-model}
Let $\A$ be a subset of $\{ \CONS, \NORM, \REF, \SDP, \UNIF, \RANK
\}$. If a formula $\phi$ is satisfiable in a Kripke structure
satisfying $\A$ with $N$ worlds, then $\phi$ is satisfiable in a
preferential Kripke structure with at most ${\Card{\Sub(\phi)}}N$
worlds.
\eLem

Combining this with \Tref{thm:small-model}, we conclude that if $\phi$
is satisfiable, then it is satisfiable in a structure of exponential
size with an exponential description.  It can be shown that this
result is essentially optimal (see \cite{HM2,FrH3Full}). However, if
there is only one agent and we assume CONS and either UNIF or SDP,
then we can get polynomial-sized models.

\Thm\label{thm:small-model-one-agent}
Let $\A$ be a subset of $\{ \CONS, \NORM, \REF, \SDP, \UNIF, \RANK \}$
containing CONS and either SDP or UNIF. If $\phi$ talks about the
knowledge and plausibility of only one agent, then $\phi$ is
satisfiable in a Kripke structure satisfying $\A$ if and only if it
is satisfiable in a preferential Kripke structure satisfying $\A$ with
at most $\Card{\Sub(\phi)}^3$ worlds.
\eThm
\prf
See Appendix~\ref{prf:axiomfull}.
\eprf

We now consider the complexity of decision procedure for the validity
    problem. The difficulty of deciding whether $\phi$ is valid
is
a function of the length of $\phi$, written $\Card{\phi}$.

\Thm\label{thm:complex}
Let $\A$ be a subset of $\{ \CONS, \NORM, \REF, \SDP, \UNIF, \RANK
\}$. If $\CONS \in \A$, but it is not the case that UNIF or SDP is in
$\A$, then the validity problem with respect to structures satisfying
$\A$ is complete for exponential time. Otherwise, the validity problem
is complete for polynomial space.
\eThm
\prf
See Appendix~\ref{prf:axiomfull}.
\eprf

If we restrict attention to the case of one agent and structures
satisfying CONS and either UNIF or SDP, then we can do better.
\Thm\label{thm:complex-one-agent}
Let $\A$ be a subset of $\{ \CONS, \NORM, \REF, \SDP, \UNIF, \RANK \}$
containing CONS and either UNIF or SDP. For the case of one agent, the
validity problem in models satisfying $\A$ is co-NP-complete.
\eThm
\prf
See Appendix~\ref{prf:axiomfull}.
\eprf

\section{Adding Time}

In the previous section, we developed a model of knowledge and
beliefs.  Having a good model of knowledge and belief is not
enough in order to study how beliefs change. Indeed, if we are mainly
interested in agents' beliefs, the additional structure of
plausibility spaces does not play a significant role in a static
setting.  However, if we introduce an explicit notion of time, we
expect the plausibility measure to (partially) determine how agents
change their beliefs. As we shall see, this gives a reasonable notion
of belief change.

In this section, we introduce time into the framework. We then examine
how time, knowledge, and plausibility interact. In particular, we
suggest a notion of {\em conditioning\/} that captures the intuition
that plausibility changes in the minimal way that is required by
changes to the agent's knowledge.

\subsection{Knowledge and Plausibility in Multi-Agent Systems}
\label{systems}
\label{SYSTEMS}

A straightforward approach to adding time is by introducing another
accessibility relation on worlds, which characterizes their temporal
relationship (see, for example, \cite{KrausL}). We
    introduce more structure into the description by adopting the
    framework of Halpern and Fagin \citeyear{HF87} for modeling
    multi-agent systems. This
    structure gives a natural definition of knowledge and an intuitive
    way to describe agents' interactions with their environment. We
    start by describing the framework of Halpern and Fagin, and then
    add plausibility.

The key assumption in this framework is that we can characterize the
    system by describing it in terms of a {\em state\/} that changes
    over time. This is a powerful and natural way to model systems.
Formally, we assume that at each point in time, each agent is in some
    {\em local state}. Intuitively, this local state encodes the
    information that is available to the agent at that time. In addition,
    there is an {\em environment}, whose state encodes
    relevant aspects of the system that are not part of the agents'
    local states.
For example, if we are modeling a robot that navigates in some office
    building, we  might encode the robot's sensor input as part of the
    robot's local state. If the robot is uncertain about his
    position, we would encode this position in the environment state.

A {\em global state\/} is a tuple $(s_e, s_1, \ldots, s_n)$ consisting
    of the environment state $s_e$ and the local state $s_i$ of each
    agent $i$.
    A {\em run\/} of the system is a function from time
    (which, for ease of exposition, we assume ranges over the natural
    numbers) to global states. Thus, if $r$ is a run, then $r(0),
    r(1), \ldots$ is a sequence of global states that, roughly
    speaking, is a complete description of what happens over time in
    one possible execution of the system. We take a {\em system\/} to
    consist of a set of runs.  Intuitively, these runs describe all
    the possible sequences of events that could occur in a system.

Given a system $\R$, we refer to a pair $(r,m)$ consisting of a run $r
    \in \R$ and a time $m$ as a {\em point}. If $r(m) = (s_e,
    s_1,\ldots,s_n)$, we define $r_i(m) = s_i$;
    thus, $r_i(m)$ is agent $i$'s local state at the point $(r,m)$.
    Finally, to reason in a logical language about such a system, we
    need to assign truth values to primitive propositions.
An {\em interpreted system\/} is a tuple $(\R,\pi)$
    consisting of a system $\R$ together with a
    mapping $\pi$ that associates a truth assignment with the primitive
    propositions at each state of the system.

An interpreted plausibility system can be viewed as a Kripke structure
    for knowledge. We say two points $(r,m)$ and $(r',m')$ are {\em
    indistinguishable\/} to agent $i$, and write $(r,m) \sim_i
    (r',m')$, if $r_i(m) = r'_i(m')$, \ie if the agent has the same
    local state at both points. This is
consistent
 with the intuition
    that an agent's local state encodes all the information available to
    the agent.
    Taking $\sim_i$ to define the $\K_i$ relation, we get a Kripke
    structure over points.%
\footnote{It is straightforward to extend these definitions
    to deal with continuous time. This is done, for example, in \cite{BLMSFull}.}

This definition of knowledge has proved useful in many applications in
    distributed systems and AI (see \cite{FHMV} and the references
    therein). As argued
    above, we want to
    add the notion of plausibility so that we can model the agent's
    beliefs. It is straightforward to do so by adding a plausibility
    assessment for each agent at each point. Formally, an {\em
    interpreted plausibility system\/} is a tuple
    $\Sys = (\R,\pi,\Plass_1,\ldots,\Plass_n)$, where $\R$ and $\pi$
    are as before,
    and
the plausibility assignment
$\Plass_i$ maps each point $(r,m)$ to a plausibility space
    $\Plass_i(r,m) = (\Omega_{(r,m,i)}, \Pl_{(r,m,i)})$.
In order to reason about the temporal aspects of the system,
we add to the language temporal modalities in the standard fashion (see
    \cite{GPSS}). These include $\Next\phi$ for
    ``$\phi$ is true at the next time step''
We call this language $\L^{KCT}$. Evaluation of temporal
    modalities at a point $(r,m)$ is done by examining the future
    points on the run $r$:
Given a point $(r,m)$ in an interpreted
    system $\Sys$, we have that

\begin{itemize}\denselist

 \item $(\Sys,r,m)\sat \Next\phi$ if
    $(\Sys,r,m+1)\sat \phi$.%
\footnote{It is easy to add other temporal modalities such as {\em until,
    eventually, since\/}, etc. These do not play a role in this work.}
\end{itemize}
This framework is clearly a temporal extension of the logic of
    knowledge and plausibility described in the previous section.

\subsection{Example: Circuit Diagnosis
    Revisited}\label{sec:xam-diag-sys}

We now show how the framework can be used to extend the example of
\Sref{sec:xam-diag} to incorporate time, allowing the agent to perform
a sequence of tests.

We want to model the {\em process\/} of diagnosis. That is, we want to
model the agent's beliefs about the circuit while it performs a
sequence of tests, and how the observations at each step affects her
beliefs. Thus, we want to model the agent and the circuit as part of a
system.
To do so, we need to describe the agent's local state and the
state of the environment. The construction we used in
Section~\ref{sec:xam-diag} provides a natural division between the
two: The agent's state is the sequence of input--output relations
observed, while the environment's state describes the faulty
components of the circuit and the values of all the lines. This
corresponds to our intuitions, since the agent can observe only the
input--output relations.  Each run describes the results of a specific
series of tests the agent performs and the results he observes.  We
make two additional assumptions: (1) the agent does not forget what
tests were performed and their results, and (2) the faults are
persistent and do not change over time. Formally, we define the
agent's state $r_1(m)$ to be $\<
\obs{r,0}, \ldots, \obs{r,m} \>$, where $\obs{r,m}$ describes the
input--output relation observed at time $m$.  We define the environment
state $r_e(m) = (\fault(r,m),\Dvalue(r,m))$ to be the failure set at
$(r)$ and the values of all the lines. We capture the assumption that
faults do not change by requiring that $\fault(r,m) =
\fault(r,0)$. The system $\R_\diag$ consists of all runs $r$ satisfying
these requirements in
which $\Dvalue(r,m)$ is consistent with $\fault(r,m)$ and $\obs{r,m}$
for all $m$.

Given the system $\R_\diag$, we can define two interpreted
plausibility systems corresponding to the two plausibility measures we
considered in Section~\ref{sec:xam-diag}. In both systems,
$\Omega_{(r,m,1)} = \K_i(r,m)$. In $\Sys_{\diag,1}$, we compare two
points $(r_1,m)$ and $(r_2,m)$ by comparing the size of
$\fault(r_1,m)$ and $\fault(r_2,m)$, while in $\Sys_{\diag,2}$ we
check whether one failure set is a subset of the other.  At a point
$(r,m)$, the agent considers possible all the points where he
performed the same tests up to time $m$ and observed the same
results. As before, the agent believes that the failure set is one of
the minimal explanations of his observations.  As the agent performs
more tests, his knowledge increases and his beliefs might change.

We define $\BEL(\Sys,r,m)$ to be the set of failure sets
(\ie diagnoses) that the agent considers possible at $(r,m)$.  Belief
change in $\Sys_{\diag,1}$ is characterized by the following
proposition.
\Pro\label{pro:diag-sys-1}
If there is some $f \in \BEL(\Sys_{\diag,1},r,m)$ that is consistent
with the new observation $\obs{r,m+1}$, then
$\BEL(\Sys_{\diag,1},r,m+1)$ consists of all the failure sets in
$\BEL(\Sys_{\diag,1},r,m)$ that are consistent with $\obs{r,m+1}$.  If
all $f \in \Bel(\Sys_{\diag,1},r,m)$ are inconsistent with
$\obs{r,m+1}$, then $\Bel(\Sys_{\diag,1},r,m+1)$ consists of all
failure sets of cardinality $j$ that are consistent with
$\obs{r,m+1}$, where $j$ is the least cardinality for which there is
at least one failure set consistent with $\obs{r,m+1}$.
\ePro
\prf
Straightforward; left to the reader.
\eprf

Thus, in $\Sys_{\diag,1}$, a new observation consistent with the
current set of most likely explanations reduces this set (to those
consistent with the new observation).  On the other hand, a surprising
observation (one inconsistent with the current set of most likely
explanations) has a rather drastic effect.  It easily follows from
Proposition~\ref{pro:diag-sys-1} that if $\obs{r,m+1}$ is surprising,
then $\BEL(\Sys_{\diag,1},r,m) \inter \BEL(\Sys_{\diag,1},r,m+1) =
\emptyset$, so the agent discards all his current explanations in this
case.  Moreover, an easy induction on $m$ shows that if
$\BEL(\Sys_{\diag,1},r,m) \inter \BEL(\Sys_{\diag,1},r,m+1) =
\emptyset$, then the cardinality of the failure sets in
$\BEL(\Sys_{\diag,1},r,m+1)$ is greater than the cardinality of
failure sets in $\BEL(\Sys_{\diag,1},r,m)$.  Thus, in this case, the
explanations in $\BEL(\Sys_{\diag,1},r,m+1)$ are more complicated than
those in $\Bel(\Sys_{\diag,1},r,m)$.  Notice that if we can
characterize the observation $\obs{r,m+1}$ in our language---that is,
if we have a formula $\phi$ such $(\Sys,r',m') \sat \phi$ if and only
if $\obs{r',m'} = \obs{r,m+1}$---then we can also express the fact
that agent $i$ considers it surprising: This is true precisely if
$(\Sys_{\diag,1},r,m) \sat \Bel_i \neg\Next\phi$.

Belief change in $\Sys_{\diag,2}$ is quite different, as the following
proposition shows.  Given a failure set $f$, we define {\it ext\/}$(f)
= \{ f' : f \subseteq f' \}$.  Thus, {\it ext\/}$(f)$ consists of all
the failure sets that extend $f$.

\Pro\label{pro:diag-sys-2}
$\BEL(\Sys_{\diag,2},r,m+1)$ consists of the minimal (according to
$\subseteq$) failure sets in $\union_{f \in \BEL(\Sys_{\diag,2},r,m)}
${\it ext}$(f)$ that are consistent with $\obs{r,m+1}$.
\ePro
\prf
Straightforward; left to the reader.
\eprf

We see that, as with $\Sys_{\diag,1}$, failure sets that are
consistent with the new observation are retained. However, unlike
$\Sys_{\diag,1}$, failure sets that are discarded are replaced by more
complicated failure sets even if some of the explanations considered
most likely at $(r,m)$ are consistent with the new observation.
Moreover, while new failure sets in $\BEL(\Sys_{\diag,1},r,m+1)$ can
be unrelated to failure sets in $\BEL(\Sys_{\diag,1},r,m)$, in
$\Sys_{\diag,2}$ the new failure sets must be extensions of some
discarded failure sets.  Thus, in $\Sys_{\diag,1}$ the agent does not
consider new failure sets as long as the observation is not
surprising. On the other hand, in $\Sys_{\diag,2}$ the agent has to
examine new candidates after each test.  The latter behavior is
essentially that described by Reiter \citeyear[Section~5]{reiter:87}.

\subsection{Axiomatizing the Language of Knowledge, Plausibility and
    Time}
\label{sec:axiomtime}
\label{SEC:AXIOMTIME}

We now present sound and complete axiomatization for the language
$\L^{KCT}$. The technical details are much in the spirit of the
results of \Sref{sec:axiomfull}, with two exceptions. First, we need
to deal also with the temporal modality $\Next$. Second, instead of
dealing with worlds, we are dealing with systems that have some
structure, \ie the distinction between agents' local state and the
environment's state. As we shall see, both issues can be dealt with
in a straightforward manner.

The axiom system \AXT\ consists of the axioms and rule in the axiom
system AX of \Sref{sec:axiomfull} and the following axioms and rule
the describe the properties of $\Next$:.
\begin{description}\denselist
 \item[T1.] $\Next \phi \land \Next( \phi\rimp\psi) \rimp \Next \psi$
 \item[T2.] $\Next \phi \equiv \neg \Next \neg \phi$
 \item[RT1.] From $\phi$ infer $\Next\phi$.
\end{description}

Let $\Sysclass$ be the set of all plausibility interpreted systems.
\Thm\label{thm:ax-time}
The axiom system \AXT\ is a sound and complete axiomatization of
$\L^{KCT}$ with respect to $\Sysclass$.
\eThm
\prf
See Appendix~\ref{prf:axiomtime}.
\eprf

We can also prove a result analogous to \Tref{thm:ax-extend} that
describes a complete axiomatization for the classes of systems
satisfying some of the assumptions we examined in
\Sref{combining}.

\Thm\label{thm:ax-time-extend}
Let $\A$ be a subset of $\{ \RANK, \NORM, \REF, \UNIF, \CONS, \SDP \}$
and let $A$ be the corresponding subset of $\{$C5, C6, C7, C8, C9,
C10$\}$. Then $\AXT \cup A$ is a sound and complete axiomatization
with respect to systems in $\Sysclass$ satisfying $\A$.
\eThm
\prf
See Appendix~\ref{prf:axiomtime}.
\eprf

\section{Prior Plausibilities}
\label{priorplaus}
\label{PRIORPLAUS}

The formal framework of knowledge, plausibility and time described in the
    previous section raises a serious problem: While it is easy to see
    where the $\sim_i$ relations that define knowledge come from, the
    same cannot be said for the plausibility spaces $\Plass_i(r,m)$. We
    now present one possible answer to this question, inspired by
    probability theory.

Up to now, we have allowed
the plausibility assessment at each point to
be
    almost arbitrary. In particular, the plausibility space $\Plass_i(r,m)$
    can be quite different from $\Plass_i(r,m+1)$.
Typically, we would expect there to be some relationship between these
successive plausibility assessments.  For example, it seems reasonable
to expect that the new plausibility assessment should incorporate
whatever was learned at $(r,m+1)$, but otherwise involve minimal changes
{f}rom $\Plass_i(r,m)$.

One way of doing this in probability theory is by {\em
    conditioning}.
If we start with a probability function $\Pr$ and observe $E$, where
$\Pr(E) > 0$,
then the conditional probability function $\Pr_E$ is defined so that
$\Pr_E(A) = \Pr(A \inter E)/\Pr(E)$.  Typically $\Pr_E(A)$ is denoted
$\Pr(A|E)$.  Notice that $\Pr_E$ incorporates the new information $E$ by
giving it probability 1.  It also is a minimal change from $\Pr$ in the
sense that
if $A,B
    \subseteq E$, then $\Pr(A)/\Pr(B) = \Pr(A|E)/\Pr(B|E)$:
the relative probability
of events consistent with $E$ is not changed by conditioning.%
\footnote{
There is another sense in which $\Pr_E$ represents the minimal change
{f}rom $\Pr$.  If we measure the ``distance'' of a probability
distribution $\Pr'$ from $\Pr$ in terms of the {\em cross-entropy\/}
of $\Pr'$ relative to $\Pr$, then it is well known that $\Pr_E$ is the
distribution that minimizes the relative cross-entropy from $\Pr$ among
all distributions $\Pr'$ such that $\Pr'(E) = 1$ \cite{X.entropy}.
Indeed, this holds true for other distance measures as well
\cite{DZ1}.}

Conditioning is a standard technique in probability theory, and can be
justified in a number of ways, one of which is the notion of ``minimal
change'' we have just described.  Another justification is a ``Dutch
book'' argument \cite{DeFinetti72,Ram}, which
shows that if an agent uses some other method of updating
probabilities, then it is possible to construct a betting game in
which he will always lose.  Probability measures are particular
instances of plausibility measures. Can we generalize the notion of
conditioning to plausibility measures?
It immediately follows from the definitions that the ordering
of the likelihood of events induced by $\Pr_E$ is determined by the
ordering induced by $\Pr$:
 \begin{quote}
$\Pr(A|E) \le \Pr(B|E)$ if and only if $\Pr(A \inter E)\le \Pr(B
    \inter E)$.
\end{quote}
We want the analogous property for plausibility:
\begin{cond}{COND}
$\Pl(A|C) \le \Pl(B|C)$ if and only if $\Pl(A \inter C)\le \Pl(B
    \inter C)$.
\end{cond}
This rule determines the order induced by posterior
plausibilities. Since we are interested only in this aspect of
plausibility, any method of conditioning that satisfies COND will do
for our present purposes.
(See
\cite{FrH7} for an examination of other properties we might require of
conditioning.)  Notice that any two methods for conditioning are
isomorphic in the following sense:
Let $S_1 = (W_1,\Pl_1)$ and $S_2 = (W_2,\Pl_2)$ be two
    plausibility spaces.
We say that $S_1$ and $S_2$ are {\em (order) isomorphic\/} if there
    is a bijection $h$ from $W_1$ to $W_2$ such that,
for $A, B \subseteq W_1$, we have $\Pl_1(A) \le \Pl_1(B)$ if and only if
$\Pl_2(h(A)) \le \Pl_2(h(B))$.  Any two definitions of conditioning
that satisfy COND result in order-isomorphic plausibility spaces (see
    \cite{FrH7}).

This discussion suggests that we define $\Pl_{(r,m+1,i)}$ to be the
    result of conditioning $\Pl_{(r,m,i)}$ on the new knowledge gained
    by agent
    $i$ at $(r,m+1)$. This, however, leads to the following technical
    problem. If the agent
    gains new knowledge at $(r,m+1)$, then $r_i(m) \neq r_i(m+1)$.
    This implies that the sets of points the agent considers possible
    are disjoint, \ie $\K_i(r,m) \inter \K_i(r,m+1) = \emptyset$.
But then CONS implies
that $\Pl_{(r,m,i)}$ and
    $\Pl_{(r,m+1,i)}$ are defined over disjoint spaces, so we cannot
    apply COND.

We circumvent this difficulty by working at the level of runs. The
    approach we propose resembles the Bayesian approach to
    probabilities. Bayesians assume that agents start with priors on
    all possible events. If we were thinking probabilistically, we
    could imagine the agents in a multi-agent system starting with
    priors on the runs in the  system.  Since a run describes a
    complete history over time, this means that the agents are putting
    a prior probability on the sequences of events that could happen.
    We would then expect the agent to modify his prior by conditioning
    on whatever information he has learned.  This is essentially the
    approach taken in \cite{HT} to defining how the agents'
    probability distribution changes in a multi-agent system.
We can do the analogous thing with plausibility.

We start by making the simplifying assumption that we are dealing with
    {\em synchronous\/} systems where agents have {\em perfect
    recall\/} \cite{HV2}. Intuitively, this means that the
    agents know what the time is and
do not forget the observations they have made.
Formally, a system is synchronous if for any $i$,
    $(r,m) \sim_i (r',m')$ only if $m = m'$.
Notice that by restricting to synchronous systems, if we further
    assume that the plausibility measure $\Plass_i(r,m)$ satisfies CONS,
    we never have to compare the plausibilities of two different
    points on the same run.
In synchronous systems, agent $i$ has perfect recall
if $(r',m+1) \sim_i
    (r,m+1)$ implies $(r',m) \sim_i (r,m)$.
Thus, agent $i$ considers run $r$ possible at the point $(r,m+1)$ only
if he also considers it possible at $(r,m)$.  This means that any runs
considered impossible at $(r,m)$ are also considered impossible at
$(r,m+1)$; an agent does not forget what he knew.

\begin{figure}
\begin{center}
\input{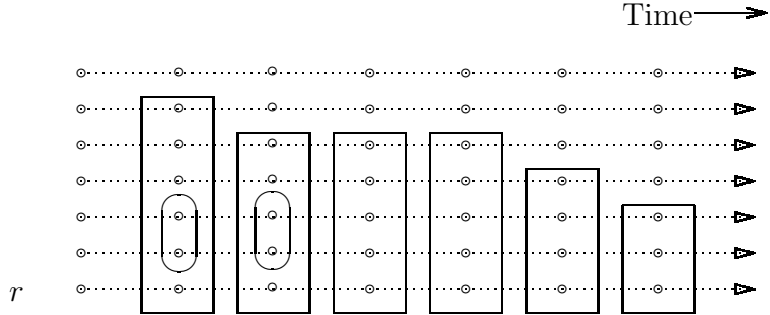}
\end{center}
\caption{Schematic description of how the agent's knowledge evolves
in time in synchronous systems with perfect recall. The boxes
represent the set of points in $\K_i(r,m)$. Since the system is
synchronous, at each time point, the agent consider possible points at
the same time. Since the agent has perfect recall, as time progresses,
the agent considers smaller and smaller sets of runs possible. The
ovals represent two disjoint events that correspond to the same set
of runs.}
\label{fig:prior-runs}
\end{figure}

Just as with probability, we assume that an agent has a prior
plausibility measure on runs, that describes
    his prior assessment on the possible executions
    of the system. As the agent gains knowledge, he updates
    his prior by conditioning.
    More precisely, at each point $(r,m)$, the agent
    conditions his previous assessment on the set of runs considered
    possible at $(r,m)$.
This is process is shown in Figure~\ref{fig:prior-runs}.
This results in an updated assessment
(posterior) of the plausibility of runs.
    This posterior induces, via a projection from runs to points, a
    plausibility measure on points.
We can think of agent $i$'s
    posterior at time $m$ as simply his
    prior conditioned on his knowledge at time $m$.

To make this precise,
let $S = (W,\Pl)$ be a plausibility space. Define the {\em
    projection\/} of $S$ on $E$ as
    $S|_E = (W|_E, \Pl|_E)$, where $W|_E = W \inter E$ and $\Pl|_E$ is
    the restriction of $\Pl$ to $W|_E$.
Projection is similar to conditioning: for
    any definition of conditioning that satisfies COND if $A, B
    \subseteq E$, then
    $\Pl(A|E) \le \Pl(B|E)$ if and only if $\Pl|_E(A) \le \Pl|_E(B)$.
Indeed, $S|_E$ is essentially isomorphic to any conditional plausibility
measure that results from conditioning on $E$.%
\footnote{
To make this precise,
we need a notion that is slightly more general than isomorphism.
    Let $P = (W,\Pr)$ be a probability space.  A set $A$ is called a
    {\em support\/} of $P$ if $\Pr(\overline{A}) = 0$. We can
    define a similar notion for plausibility spaces. Let $S = (W,\Pl)$ be a
    plausibility space. We say that $A \subseteq W$ is a {\em
    support\/} of $S$, if for all $B \subseteq W$, $\Pl(B) = \Pl(B
    \inter A)$. Thus, only $B \inter A$ is relevant for determining
    the plausibility of $B$. This certainly implies that
    $\Pl(\overline{A}) = \bot$, since we must have $\Pl(\overline{A})
    = \Pl(A \inter \overline{A}) = \Pl(\emptyset)$, but the converse
    does not hold in general. In probability spaces,
    $\Pr(\overline{A}) = 0$ implies that $\Pr(B) = \Pr(B \inter A)$
    for all $B$, but the analogous condition does not hold for
    arbitrary plausibility spaces.
We say that two plausibility spaces $S_1$ and $S_2$ are
{\em essentially (order) isomorphic\/} if there are supports $C_1$ and
    $C_2$ of $S_1$ and $S_2$, respectively,
such that $S_1|_{C_1}$ is isomorphic to $S_2|_{C_2}$.
It is easy to see that, as expected, essential isomorphism
    defines an equivalence relation among plausibility spaces.
Finally, it is easy to see that if $S = (W,\Pl)$, then $(W,\Pl(\cdot |
    E))$ is essentially
    isomorphic to $S|_E$ when we use any conditioning method
    that satisfies COND.
}

\begin{figure}
\begin{center}
\input{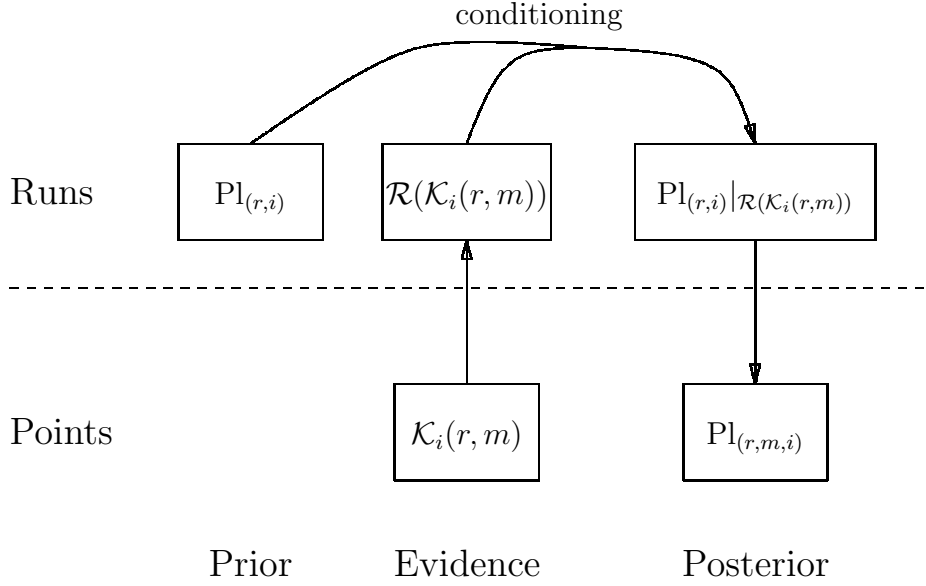}
\end{center}
\caption{Schematic description of the entities involved in the
definition of priors. Note some are defined over runs and some over points.}
\label{fig:conditioning}
\end{figure}

We can now define what it means for a
plausibility measure on points to be generated by a prior.  Suppose
that agent $i$'s prior plausibility at run $r$ is $\Plass_{(r,i)} =
(\R_{(r,i)},\Pl_{(r,i)})$, where $\R_{(r,i)} \subseteq \R$. Our
intuition is that the agent conditions the prior by his knowledge at
time $(r,m)$. In our framework, the agent's knowledge at time $m$ is
the set of point $\K_i(r,m)$. We need to convert this set of points to
an event in terms of runs.
If $A$ is a set of points, we define $\R(A) = \{ r : \exists m( (r,m)
\in A)\}$ to be the set of runs on which the points in $A$ lie.  Using
this notation, the
set of runs agent $i$ considers possible at $(r,m)$ is simply
$\R(\K_i(r,m))$. Thus, after conditioning on this set of runs, we get
agent $i$'s {\em posterior\/} at $(r,m)$, which is simply the
projection of the prior on the observation:
$\Pl_{(r,i)}|_{\R(\K_i(r,m))}$. We now use this plausibility measure,
which is a measure on a set of runs,
to define $\Plass_i(r,m)$, which is a measure on a set of points. We do so
in the most straightforward way: we project each run to a point that
lies on it.
Formally, we say that $\Plass_i(r,m)$ is the {\em time $m$ projection\/} of
$\Plass_{(r,i)}|_{\R(\K_i(r,m))}$ if $\Plass_i(r,m) =
(\Omega_{(r,m,i)},\Pl_{(r,m,i)})$, where
$\Omega_{(r,m,i)} = \{(r',m) \in \K_i(r,m) : r' \in \R_{(r,i)} \}$
and for all $A \subseteq \Omega_{(r,m,i)}$, we have that
$\Pl_{(r,m,i)}(A) = \Pl_{(r,i)}|_{\R(\K_i(r,m))}(\R(A))$.
$\Pl_{(r,m,i)}$ is the agent's plausibility measure at $(r,m)$. This
process is described in Figure~\ref{fig:conditioning}. The main
complications are due to the transition back and forth between entities
defined  over runs and ones defined over points.

We remark that if the system satisfies perfect recall as well as synchrony,
our original intuition that $\Plass_i(r,m+1)$ should be the result of
conditioning
$\Plass_i(r,m)$ on the knowledge that agent $i$ acquires at $(r,m+1)$
can be captured more directly. We can
in fact construct $\Plass_i(r,m+1)$ from $\Plass_i(r,m)$ by what can
be viewed as conditioning on the agent's new information: We take
$\Plass_i(r,m)$ and project it one time step forward by replacing each
point $(r',m)$ by $(r',m+1)$. We then condition on $\K_i(r,m+1)$ (\ie
the agent's knowledge at $(r,m+1)$) to get $\Plass_i(r,m,i+1)$.

\Pro\label{pro:local-change}
Let $\Sys$ be a  synchronous system satisfying perfect recall such
that $\Pl_{(r,m,i)}$ is the time $m$ projection of a prior
$\Pl_{(r,i)}$ on runs for all runs $r$, times $m$, and agents $i$.
Let $\MapBack(A) = \{ (r,m) : (r,m+1) \in A \}$.
Then $\Pl_{(r,m+1)}(A) \le \Pl_{(r,m+1)}(B)$ if and only if
$\Pl_{(r,m)}(\MapBack(A)) \le \Pl_{(r,m)}(\MapBack(B))$, for all runs
$r$, times $m$, and sets $A,B \in \Omega_{(r,m+1)}$.
\ePro
\prf
Straightforward; left to the reader.
\eprf

We say that $\Sys = (\R,\pi,\P)$ satisfies PRIOR if $\Sys$ is
    synchronous and for each run $r$ and agent $i$ there is a prior
    plausibility $\Plass_{(r,i)}$ such that
for all $m$,
$\Plass_i(r,m)$ is
the time $m$ projection of $\Plass_{(r,i)}$.

\commentout{
Notice that if $\Sys$ satisfies PRIOR and
$\Plass_{(r,i)}$ is
    independent of $r$, so that agent $i$'s prior is independent of
    the run he is in, then
$\Sys$ satisfies SDP.
A somewhat weaker assumption---that
    the set of runs can be partitioned into disjoint subsets
    $\R_1, \ldots, \R_k$ such that for $r, r' \in \R_j$, we have
    $\Plass_{(r,i)} = \Plass_{(r',i)} = (\R_j,\Pl_j)$---ensures that
$\Sys$ satisfies UNIF.
Intuitively, the sets $\R_j$ correspond to different settings of
parameters.  Once we set the parameters, then we fix the plausibility
measure (and it is the same at all runs that have the same setting of
the parameters).
Finally, it is easy to see that if each plausibility space $\Plass_{(r,i)}$
satisfies REF, then $\Sys$ satisfies REF.
}

\xam\label{xam:diag-prior}
It is easy to verify that the two systems we consider in
Section~\ref{sec:xam-diag-sys} satisfy PRIOR.  In both systems, the
prior $\Plass_{(r,i)}$ is independent of the run $r$, and is determined by
the failure set in each run.
\exam

\commentout{
The approach we propose resembles the Bayesian approach to probabilities.
Bayesians assume that agents start with priors on events.
If we were thinking probabilistically, we could imagine the agents
in a multi-agent system starting with priors on the runs in the
system.  Since a run describes a complete history over time, this
means that the agents are putting a prior probability on the
sequences of events that could happen.  We would then expect the
agent to modify his prior by conditioning on whatever information
he has learned.  This is essentially the approach taken
in \cite{HT} to
defining how the agents' probability distribution changes in a multi-agent
system.
}

By using prior plausibility measures, we have reduced the question
of where the plausibility measure at each point comes from to
the simpler question of where the prior comes from.
While this question is far from trivial, it is analogous to a
question that needs to be addressed by anyone using a Bayesian
approach.
Just as with probability theory, in many applications there is a
natural prior (or class of priors) that we can use.
By conditioning on plausibility rather than probability, we can deal
with a standard problem in the Bayesian approach,
    that of conditioning on an event of measure 0:
    Notice that
    whenever a prior assigns an event a probability measure of 0 it is
    not possible to condition on that event.
The standard solution in
    the Bayesian school is to give every event of interest,
    no matter how unlikely,
    a small positive probability.%
\footnote{Of course, this requires that there be only countably
many events of interest.}
We may well discover that a formula $\phi$ that we believed to be
true, \ie one that was true in all the
most plausible worlds, is in fact false.
Under the probabilistic interpretation of plausibility, this means
that we are essentially conditioning on an event ($\neg \phi$) of
measure 0.  The plausibility approach has no problem with this:
the conditioning process described
above still makes perfect sense.

\subsection{Conditioning as Minimal Change of
Belief}
\label{sec:cond-as-minimal-change}
\label{SEC:COND-AS-MINIMAL-CHANGE}

In this section we examine the properties of conditioning as an
approach to minimal change of beliefs
and relate
our approach to others in the literature.

Recall that QUAL guarantees that belief is closed under logical
implication and conjunction (Theorem~\ref{QUALthm}).  In a synchronous
system where the prior satisfies QUAL, it is not hard to see that
conditioning preserves QUAL.  Thus, we get the following result.
\Pro
Let $\Sys$ be a synchronous system satisfying
perfect recall
and PRIOR.
If the prior $\Pl_{(r,i)}$ satisfies A2 for all runs $r$ and agents
$i$, then
axiom K2 is valid in $\Sys$ for $\Bel_i$.
\ePro
\prf
Straightforward; left to the reader.
\eprf

This result shows that condition A2 is sufficient to get beliefs that
satisfy K2.
Is it also necessary? In general, the answer is no.  However, A2 is
the most natural condition that ensures that K2 is satisfied. To see
this,
note that if K2 is valid in $\Sys$ then A2 holds for all pairwise
disjoint subsets $A_1$, $A_2$ and $A_3$ of points in $\Sys$
definable in the language
such that $\R(\K_i(r,m)) = A_1 \union A_2
\union A_3$ for some run $r$, agent $i$, and time $m$.
Thus, if we assume that the language is rich enough so that all subsets
of $\Sys$ are definable (in that, for each subset $A$ and agent $i$,
there is a formula $\phi$ and point $(r,m)$ such that $A =
\intension{\phi}_{(r,m,i)}$), then K2 forces A2.

In view of this discussion, we focus
in this section on synchronous systems with a qualitative prior.

Next, we examine how changes in beliefs are determined by the prior.
Using Proposition~\ref{pro:local-change}, we now show that we can
characterize, within our language, how the agent's beliefs change via
conditioning, provided that we can describe in the language what
knowledge the agent acquired.  We say that a formula $\phi$ {\em
characterizes agent $i$'s knowledge at $(r,m+1)$\/} with respect to
his knowledge at $(r,m)$ if, for all $(r',m) \in \K_i(r,m)$, we have
$(r',m+1) \sat \phi$ if and only if $(r',m+1) \in \K_i(r,m+1)$. That
is, among the points that succeed points that are considered possible
at time $m$, exactly these satisfying $\phi$ are considered possible
at time $m+1$. Of course, it is not always possible to characterize
the agent's new knowledge by a formula in our language. However, in
many applications we can limit our attention to systems where it is
possible.  (This is the case, for example, in our treatment of
revision and update in
\cite{FrThesis,FrH2Full}.)
In such systems, we can characterize within the agent's belief change
process in the language.
\Pro\label{pro:bel-change}
Let $\Sys$ be a synchronous system satisfying perfect recall and
PRIOR.  If $\phi$ characterizes agent $i$'s knowledge at $(r,m+1)$
with respect to his knowledge at $(r,m)$, then $(\Sys,r,m+1) \sat \psi
\Condi \xi$ if and only if $(\Sys,r,m) \sat \Next(\phi\land\psi)
\Condi \Next\xi$.
\ePro
\prf
See Appendix~\ref{prf:cond-as-minimal-change}.
\eprf

\Cor
Let $\Sys$ be a  synchronous system satisfying perfect recall and PRIOR.
If $\phi$ characterizes agent $i$'s knowledge at $(r,m+1)$ with
respect to his knowledge at $(r,m)$, then
$(\Sys,r,m+1) \sat \Bel_i\psi$ if and only if $(\Sys,r,m) \sat \Know_i
    (\Next\phi \rimp (\Next\phi\Condi\Next\psi))$.
Moreover, if $\Sys$ also satisfies SDP, then
$(\Sys,r,m+1) \sat \Bel_i\psi$ if and only if $(\Sys,r,m) \sat
    \Next\phi\Condi\Next\psi$.
\eCor

We now use this result to relate our approach to other approaches for modeling
conditionals in the literature.
Boutilier \citeyear{Boutilier92}, Goldszmidt and Pearl
\citeyear{Goldszmidt92}, and Lamarre and Shoham \citeyear{LamarreShoham}
give conditional statements similar semantics
(using a preference
ordering), but $\phi\Cond \psi$ is read ``after learning $\phi$,
$\psi$ is believed''.  Two crucial assumptions are made in these
papers.  The first is that the agent considers only one plausibility
assessment, which in our terminology amounts to SDP.  The second is
that propositions are static, \ie their truth value does not change
along a run.%
\footnote{This assumption is only implicit, since none of these papers
have an explicit representation of time.  Nevertheless, it is clear
that this assumption is being made.}
Formally, a system is {\em static\/} if $\pi(r(m)) = \pi(r(0))$ for
all runs $r$ and times $m$. This implies that for any propositional
formula $\phi$, we have that $\phi \equiv \Next\phi$.
These two assumptions lead to a characterization of belief change.
\Cor\label{connection}
Let $\Sys$ be a synchronous static system satisfying PRIOR, SDP, and
perfect recall, and let $\phi$ and $\psi$ be propositional formulas.
If $\phi$ characterizes agent $i$'s knowledge at $(r,m+1)$ with
respect to his knowledge at $(r,m)$, then $(\Sys,r,m+1) \sat
\Bel_i\psi$ if and only if $(\Sys,r,m) \sat \phi\Condi\psi$.
\eCor

While this result shows that, in certain contexts, there is a
connection between a statement such as ``typically $\phi$'s are
$\psi$'s'' (which is how we have between interpreting $\phi \Condi
\psi$) and ``after learning $\phi$, $\psi$ is believed'' (which is how
it is interpreted in \cite{Boutilier92,Goldszmidt92,LamarreShoham}),
the two readings are in general quite different.  For one thing,
notice that Corollary~\ref{connection} assumes that $\phi$ and $\psi$
are propositional formulas. This is a necessary assumption.  If $\phi$
and $\psi$ contain modal formulas, then $\phi\Cond\psi$ does not
necessarily imply that the agent believes $\psi$ at the next time
step. For example, if $(\Sys,r,m) \sat \Bel_i\psi$, then for any
formula $\phi$, we have $(\Sys,r,m) \sat \phi\Condi \Bel_i\psi$,
regardless of whether $\Bel_i\psi$ is believed at $(r,m+1)$.  In
\cite{FrH4}, we examine conditionals of the form $\phi\RCond\psi$
intended to capture the second interpretation ``$\psi$ is believed
after learning $\phi$''. The semantics
for these conditionals involves examining future time points, just as our
intuitive reading dictates. As we have just seen, $\RCond$ and $\Cond$
are quite different when we consider modal formulas in the scope of
these conditionals.

This discussion shows one of the benefits of representing time
explicitly. In our framework we can distinguish between agents'
plausibility assessment and their belief dynamics. Of course, we would
like agents to be persistent in their assessment, which is exactly
what conditioning captures.  In the presence of several
assumptions, we get
a close connection between agents'
conditional beliefs and how their beliefs change. This allows us to
identify some of the assumptions implicitly made in previous
approaches. For example, all of the approaches we mentioned above
would not apply when we consider a changing environment, since they
cannot reason about how the environment changes between one time point
and the next.

\commentout{
The assumption that the system is static is also crucial in
Corollary~\ref{connection}.
In
the following chapters
we examine the
connection between these two readings of conditionals more carefully.
The explicit representation of time in our framework makes it easier to
do so.
}

Finally, we examine the work of Battigalli and Bonanno
\citeyear{BatBon97}. They consider a logic of knowledge,
belief, and time,
and attempt to capture properties of ``minimal change'' of beliefs.
Their language is slightly different from ours. Instead of introducing
a temporal modality, they define a different belief and knowledge
modality for each time step: $\Bel^t\phi$ reads ``the agent believes
$\phi$ at time $t$''. Battigalli and Bonanno also assume that
propositions are static and do not change in time. Thus, the only
changes are in terms of the agent's knowledge and belief. Battigalli and
Bonanno propose an axiom system similar to the axioms of Kraus and
Lehmann (that is, they use K5 for knowledge is K5, KD45 for belief,
and take axioms KB1 and  KB2 of Section~\ref{kbsec} to characterize
the connection between knowledge and belief) that also includes two
additional axioms that can be written in our language as
\begin{description}\denselist
 \item[BT1.] $\Bel_i\Next\Bel_i\phi \rimp \Bel_i\phi$
 \item[BT2.] $\Bel_i\phi \rimp \Bel_i\Next\Bel_i\phi$
\end{description}
Battigalli and Bonanno claim that these axioms capture the principle
that the agent does not change her mind unless new knowledge forces
her to do so.
Intuitively, this principle also applies to conditioning,
and thus it is instructive to understand when these axioms are
satisfied in our framework.

It turns out that RANK combined with a minimal assumption implies both
BT1 and BT2.
We say that a system
has {\em  finite branching\/} if it allows only
finitely many
``branches'' at each local state of an agent (that is there are only
finitely many observations that an agent can make at each point).
\Lem\label{lem:BatBon}
Let $\Sys$ be a synchronous static system satisfying PRIOR, RANK, SDP, and
perfect recall that has finite branching. Then $(\Sys,r,m) \sat
\Bel_i\phi \dimp \Bel_i\Next\Bel_i\phi$ for all propositional
formulas $\phi$.
\eLem
\prf
See
Appendix~\ref{prf:cond-as-minimal-change}.
\eprf

Are these conditions necessary to characterize BT1 and BT2? The answer
is no. First, the proof of Lemma~\ref{lem:BatBon} applies to systems
with infinite branching, if the agents' prior satisfies an infinitary
version of A2. As shown in \cite{FrHK1}, this infinitary version is
satisfied by
$\kappa$-rankings and preference orderings that are {\em well founded\/}
(that is, they have no infinite descending sequences $\cdots \prec w_3 \prec
w_2 \prec w_1$).
Thus, any system
with
static propositions
whose prior is induced by a well-founded preference order
satisfies BT1 and BT2.
Note that BT1 and BT2 do not characterize RANK,
since they put restrictions
only on certain events (ones definable by a conjunction of a
formula and the agent's new knowledge at some
time point). However, RANK is the most natural restriction that
implies these axioms.

Thus, we see that Battigalli and Bonanno essentially require
systems with minimal change to satisfy conditioning
with a prior that is a ranking.
As we shall see in the next
section, similar requirements are made by the AGM formulation of
belief revision \cite{agm:85}.

\subsection{Properties of Prior Plausibilities}
\label{sec:prior-prop}
\label{SEC:PRIOR-PROP}

If we take the plausibilities in a system to be generated by a prior,
then many of the conditions we are interested in, such as QUAL and REF,
can be viewed as being as being induced by the analogous property
on the prior.
We have considered these properties only in the
context of Kripke structures for knowledge and probability, so to make
sense of the prior having the ``analogous property'',
we have to be able to view the
set of runs as a Kripke structure for knowledge and probability.
    Let $\Sys$ be a synchronous system
    satisfying perfect recall and PRIOR. Define $M^r_\Sys =
    (\R,\pi^r,\K^r_1,\ldots,\K^r_n,\Plass^r_i,\ldots,\Plass^r_n)$, where
    $\pi^r$ is an arbitrary truth assignment, $\K^r_i$ is the full
    relation, \ie $\R\cross\R$, and $\Plass^r_i(r) = \Plass_{(r,i)}$, the prior
    of agent $i$ at run $r$.
\Pro\label{pro:prior-prop}
Let $\Sys$ be a synchronous system satisfying perfect recall and
    PRIOR. If $M^r_\Sys$ satisfies QUAL, REF, SDP, UNIF or RANK, then
    so does $\Sys$.
\ePro
\prf
Straightforward; left to the reader.
\eprf

Thus, by constructing  priors that satisfy various properties, we can
 ensure
    that the resulting system also satisfies them.
In particular, \Pref{pro:prior-prop} implies that if $\Plass_{(r,i)}$ is
    independent of $r$, so that agent $i$'s prior is independent of
    the run he is in, then $\Sys$ satisfies SDP.
A somewhat weaker assumption---that
    the set of runs can be partitioned into disjoint subsets
    $\R_1, \ldots, \R_k$ such that for $r, r' \in \R_j$, we have
    $\Plass_{(r,i)} = \Plass_{(r',i)} = (\R_j,\Pl_j)$---ensures that
$\Sys$ satisfies UNIF.
Intuitively, the sets $\R_j$ correspond to different settings of
parameters.  Once we set the parameters, then we fix the plausibility
measure (and it is the same at all runs that have the same setting of
the parameters).

We conclude this section by examining
whether assuming conditioning limits the
    expressiveness of our belief change operation.
A well-known result of Diaconis and Zabell
    \citeyear{DZ1} that shows that, in a precise sense, any form
    of {\em coherent\/} probabilistic belief change can be described by
    conditioning.
In particular, they show that, given two probability distributions $\Pr$
and $\Pr'$ on a finite space $W$ that are coherent in the sense that
$\Pr(A) = 0$ implies that $\Pr'(A) = 0$,
there is a space $W^*$ of the form $W \times X$, a subset $E$ of
$W^*$, and a distribution $\Pr''$ on $W^*$ such that, for all
$A \subseteq W$, we have
$\Pr''(A \times X) = \Pr(A)$ (so that $\Pr''$ can be viewed as an
extension of $\Pr$) and $\Pr'(A) = \Pr''(A \times X|E)$.

We can prove a result in a somewhat similar spirit in our framework.
The first step is to
define a plausibilistic analogue of coherence in systems.

Let $\Sys$ be a synchronous system.
    We say that $\Sys$ is {\em coherent\/} if the
    following condition is satisfied for all $r$ and $m$:
Suppose $R \subseteq \R$, $A^m \subseteq \Omega_{(r,m,i)}$, $\R(A^m) =
    R \inter \R(\Omega_{(r,m,i)})$, $A^{m+1}
    \subseteq \Omega_{(r,m+1,i)}$, and $\R(A^{m+1}) = R \inter
    \R(\Omega_{(r,m+1,i)})$.  If $\Pl_{(r,m,i)}(A^m) = \bot$, then
    $\Pl_{(r,m+1,i)}(A^{m+1}) = \bot$.
Despite the different formulation, this condition is
    analogous to the probabilistic coherence of Diaconis and Zabell.
    Roughly speaking,
    if a set of runs has plausibility $\bot$ (which
    is analogous to probability 0 for Diaconis and Zabell) at time
    $m$, then it is required to have plausibility $\bot$ at time $m+1$.
More precisely,
coherence of a system ensures that sets of runs that were considered
    implausible at $(r,m)$, either by being outside $\Omega_{(r,m,i)}$ or
    by being given plausibility $\bottom_{(r,m,i)}$, are also considered
    implausible at $(r,m+1)$.
Note, this condition does not put any
    constraints on how the runs that are considered possible are
    ordered.
It is easy to verify that the following axiom is valid in coherent
    systems:
\begin{description}\denselist
 \item[COH.] $\PBox_i\Next\phi \rimp \Next\PBox_i\phi$
\end{description}
\Pro\label{COH}
If $\Sys$ is a synchronous and coherent system, then COH is valid in
    $\Sys$.
\ePro
\prf
Straightforward; left to the reader.
\eprf

There is a sense in which the converse to Proposition~\ref{COH} holds as
well: Given a synchronous system that is not coherent, we can define a
truth assignment $\pi$ in this system for which COH does not hold.%
\footnote{We remark that COH is analogous to the axiom
$\Know_i\Next\phi \rimp \Next\Know_i\phi$ that characterizes perfect
recall in synchronous systems \cite{FHMV}.  Roughly speaking, this is
because
coherence  ensures that the agent does not forget what
she ruled
out as
    implausible.}

It is easy to see that coherence is a necessary condition for
    satisfying PRIOR.
\Pro
If $\Sys$ is a synchronous system satisfying perfect recall and
    PRIOR, then $\Sys$ is coherent.
\ePro
\prf
Straightforward; left to the reader.
\eprf

Thus, PRIOR forces systems to be coherent, and hence to satisfy COH.
It also forces systems to satisfy CONS, and hence C5.  As we shall see,
it also forces some other semantic properties.  Nevertheless,
we can show that for coherent systems that satisfy CONS, PRIOR
does not force any additional properties, by proving
an analogue to the Diaconis and Zabell result in our framework.

    We say that a formula $\phi\in\L^{KCT}$ is {\em temporally
    linear\/} if temporal modalities in $\phi$ do not appear in the
    scope of the $\K_i$ or $\Condi$ modalities.
Thus, for example, a formula such as $(\phi \Condi \psi) \rimp
    \Next\Bel_i\psi$ is temporally linear, while $K_i (\Next\phi
    \Condi \Next\psi) \rimp \Next\Bel_i\psi$ is not.
Temporal linearity ensures that
    all the temporal connectives in $\phi$ are evaluated with respect
    to a single run.
The following result says that,
at least for temporally linear formulas,
we can
view belief change in a
coherent
system $\Sys$ as coming from conditioning on a
prior, in the sense that we can embed $\Sys$ into a larger system where
this is the case.
\Thm\label{thm:coherence}
Let $\A$ be a subset of $\{ \QUAL, \NORM, \REF, \RANK \}$ and
let $\Sys$ be a coherent synchronous system satisfying perfect recall, CONS,
    and $\A$. Then there is a synchronous system $\Sys'$
    satisfying perfect recall, PRIOR, and $\A$, and a mapping $f:\R
\mapsto
    \R'$ such that for all temporally linear formulas $\phi\in\L^{KCT}$,
    we have $(\Sys,r,m)\sat \phi$ if and only if $(\Sys',f(r),m) \sat \phi$.%
\footnote{We note that this result is, in a sense, stronger than
    Diaconis and Zabell's.
    They examine only the
   probability of events, which are essentially propositional formulas
(\ie formulas without modal operators).}
\eThm
\prf
See Appendix~\ref{prf:prior-prop}.
\eprf

Notice that formulas that just compare an
agent's beliefs (or knowledge) at successive time points are temporally
linear.   All the AGM postulates and the KM postulates (when translated
to our language) are of this form.  Not surprisingly, as we show in
\cite{FrThesis,FrH2Full},
these postulates can be captured by systems with the
appropriate prior plausibility.

Can we extend \Tref{thm:coherence} to the full language?
We conjecture that \Tref{thm:coherence} actually holds for all
    $\phi\in\L^{KCT}$, not just temporally linear formulas.
This conjecture implies that a formula is valid with
    respect to synchronous systems satisfying perfect recall, CONS,
    and PRIOR
    if and only if it is valid with respect to synchronous coherent
    systems satisfying CONS
and perfect recall. That is, except for
COH and C9, we do not get any new properties by assuming PRIOR
and CONS.

Note that the construction described by \Tref{thm:coherence} does not
    necessarily preserve SDP
    or UNIF in the transformation from $\Sys$ to $\Sys'$. This is due to
    the fact that in the presence of SDP or UNIF, PRIOR forces new
semantic
    properties. Recall that UNIF implies that there is a partition of
    possible points such that two points $(r,m)$ and $(r',m')$ are in
    the same cell if and only if $\Plass_i(r,m) = \Plass_i(r',m')$. Let
    PERSIST be the requirement that this partition changes minimally in
    time. More precisely, we say that a system satisfies PERSIST if
    for all runs $r,r' \in \R$ and $m$ such that $(r,m+1) \sim_i
    (r',m+1)$, we have that $\Plass_i(r,m+1) = \Plass_i(r',m+1)$ if and only
    if $\Plass_i(r,m) = \Plass_i(r',m)$. Intuitively, PERSIST (in the presence
    of synchrony, perfect recall, and CONS)
    implies that the partition of points at time $m+1$ is determined by
    the partition of corresponding points at time $m$ and the
    knowledge relation at time $m+1$.
\commentout{
\Pro
If $\Sys$ is a synchronous system that satisfies perfect recall,
    PRIOR, and either SDP or UNIF, then $\Sys$ satisfies PERSIST.
\ePro

Moreover, it turns out that PERSIST always holds in systems that
    satisfy SDP.
\Pro
If $\Sys$ is a synchronous system that satisfies perfect recall and
    SDP, then $\Sys$ satisfies PERSIST.
\ePro
}
\Pro
If $\Sys$ is a synchronous system that satisfies perfect recall and
either PRIOR and UNIF, or SDP, then $\Sys$ satisfies PERSIST.
\ePro
\prf
Straightforward; left to the reader.
\eprf

It is not clear to us at this stage whether PERSIST forces new
    properties in our language. However, if we assume that PERSIST holds,
    we can get a result analogous to \Tref{thm:coherence}.
\Thm\label{thm:coherence-UNIF}
Let $\A$ be a subset of $\{ \QUAL, \NORM, \REF, \SDP, \UNIF, \RANK \}$
and let $\Sys$ be a coherent synchronous system satisfying perfect recall,
CONS,
    PERSIST,
    and $\A$. Then there is a
    synchronous system $\Sys'$ satisfying perfect recall, PRIOR, and
    $\A$, and a mapping $f:\R \mapsto \R'$ such that for all
    temporally linear formulas $\phi\in\L^{KCT}$, $(\Sys,r,m)\sat
    \phi$ if and only if $(\Sys',f(r),m) \sat \phi$.
\eThm
\prf
See Appendix~\ref{prf:prior-prop}.
\eprf

Thus, the question of whether PRIOR forces new properties in the
presence of UNIF reduces to the question of whether PERSIST
forces new properties.
Finally, since SDP implies PERSIST, PRIOR does not force new
    properties in the presence of SDP.

Our discussion of conditioning and priors up to now assumed
    synchrony and perfect recall. Can we make sense of
    conditioning when we relax these assumptions?
Note that the definition of PRIOR does not rely on perfect recall.
    PRIOR is well defined even in systems where agents can
    forget. However, in such systems,
the intuitions that motivated the use of PRIOR are no longer valid.
In particular, PRIOR does not imply coherence and
the analogue to Proposition~\ref{pro:local-change} does
not hold:  we no longer can construct
    $\Plass_i(r,m+1)$ from $\Plass_i(r,m)$ since runs that are considered
    impossible at time $m$ might be considered possible at time $m+1$.%
\footnote{We could, of course, redefine PRIOR so as to guarantee that
\Pref{pro:local-change} holds, but this leads to other complications.}
Dropping the assumption of synchrony also leads to problems, even in the
presence of perfect recall.
In an
asynchronous setting, an agent might consider several points on
the same run possible.  The question then arises as to how (or whether)
we should distribute the plausibility of a run over these points.
Two approaches are considered in a probabilistic setting in
\cite{Piccione}, in the context of analyzing games with imperfect
recall.  It would be of interest to see to what extent these
approaches can be carried over to the plausibilistic setting.

\section{Conclusion}\label{conclusion}

We have proposed a framework  for belief dynamics
that combines knowledge, time,
    and plausibility (and hence beliefs), and investigated a number of
properties of the framework, such as complete axiomatizations for
various sublanguages and various properties of the
relationships between the modal operators.
Of course, the obvious question is why we should consider this framework
at all.

There are two features that distinguish our approach from others.
The first is that we use plausibility to model uncertainty, rather than
other approaches that have been mentioned in the literature, such as
preference orderings on worlds or $\epsilon$-semantics.
The second is that we include knowledge and time, as well as
belief, explicitly in the framework.

    We could have easily
    modified the framework to use other ways of modeling uncertainty.
    Indeed, in a preliminary version of this paper
    \cite{FrH1}, we used preference orderings.
We have chosen to use plausibility measures for
    several reasons.
First, plausibility measures generalize
    all approaches to representing uncertainty that we are aware of.
    The use of plausibility makes it easier to compare our
approach, not only to
    preference-based approaches (\eg \cite{Boutilier92}), but also to
approaches
    based on $\kappa$-rankings (\eg \cite{Goldszmidt92}), probably measures
    (\eg \cite{HT}), or any other measure of uncertainty.
More
importantly, it makes it easier for us to incorporate intuitions from
other approaches.  We have already seen one example of this
phenomenon in the present paper: we defined
a plausibilistic analogue of conditioning, and used it
to
 model minimal change.  As we show in \cite{FrH2Full}, we can represent
the standard approaches to minimal change---belief revision and belief
update---in terms of conditioning.  Moreover, the semantic
characterization of conditioning should allow us to apply it
more easily to deal with complications that arise when the language lets
us reason about multiple agents, actions, and beliefs about beliefs.
Another example of adopting probabilistic intuitions is given in
\cite{FrThesis,FrH7,FrH6}, where
plausibilistic analogues of independence and Markov chains are
described and used to define a novel approach to belief change.
We believe that these notions will have applications elsewhere as well.
    Finally,
plausibility measures have the advantage of greater expressive power
than other approaches.
For example,
work on defaults has mainly focused on properties of structures with a
    finite number of worlds. In our framework, however, even a simple system
    with two global states might have
an
    uncountable number of runs.
    As shown in
   \cite{FrHK1}, once we examine structures with infinitely many worlds,
   qualitative plausibility measures
    can capture natural ordering of events that cannot be captured by
    preference orderings, possibility measures, or
    $\kappa$-rankings.

As we have tried to argue throughout the paper, the
explicit representation of knowledge and time makes it much easier to
study belief dynamics.
    Most current work in the area
    examines only the beliefs of an agent and how they change
after
    incorporating a new belief.  Many simplifying assumptions are made:
    that there is a single agent, that the agent's
    knowledge does not change, that new information can be characterized
in
    the language, and so on.
    It is useful to study this simple setting in
order to get at the basic issues of belief change.
    However, these simplifying assumptions are not
    suitable when we want examine belief change in more realistic
    settings (such as the diagnosis example of
\Sref{sec:xam-diag-sys}).
This means that most of the results in the
    current belief change literature are not directly applicable in
many standard
    AI problems.  Our
    framework dispenses
    with most of the simplifying assumptions made in the literature, and
thus can be viewed as a first step towards providing a model of more
realistic settings of belief change.

We have focused here on the foundations of the framework.
In the future, we hope to apply the framework to examine
more realistic problems.  We have already begun to do this.
For example, in
    \cite{FrH1} we provide a detailed
    analysis of iterated prisoner dilemma games between two agents. It
    is well-known that the players cannot cooperate when they have common
    knowledge of rationality. However, we show that they can cooperate
    when they have common belief of rationality.  A
    recent proposal by van~der~Meyden \cite{Meyden94a} for multi-agent
    belief change can easily be  embedded in our framework
    \cite{MeydenPersonal94}.  We hope to use our framework to study some
of the problems considered by van~der~Meyden, such as speech-act
semantics.
Another natural application area is reasoning about
    actions and planning in the presence of uncertainty.
We believe
that the flexibility and expressive power of the framework will help to
clarify what is going on in all these areas.

\subsection*{Acknowledgements}
The authors are grateful to Piepaolo Battigalli,
Craig Boutilier,
Ronen Brafman, Ron Fagin,
Moises Goldszmidt, Ron van~der~Meyden, Yoav Shoham,
and particularly Daphne Koller
    and Moshe Vardi for comments on previous versions of this paper and useful
    discussions relating to this work.

\appendix
\section{Proofs}
\label{proofs}

\subsection{Proofs for Section~\protect\ref{kbsec}}
\label{prf:kbsec}

\rethm{thm:LB}
K (\respc K45, KD45) is a sound and complete axiomatization
for $\L^B$ with respect to $\M$ (\respc $\M^{\sCONS}$, $\M^{\sCONS,\sNORM}$).
\erethm

\prf
As usual, soundness is straightforward, so we focus on completeness.
    We prove completeness by showing  that for
$M \in \M_K$ (\resp $\M_K^{et}$, $\M_K^{est}$)
    there is a structure $M^+ \in \M$ (\resp $M^{\sCONS}$,
    $\M^{\sCONS,\sNORM}$) such that for all $\phi \in \L^B$, we have
    $(M,w) \sat \phi$ if and only if $(M^+,w) \sat \phi$. Completeness
    then follows from \Tref{thm:S5}.

Let $M = (W,\pi,\B_1,\ldots, \B_n)$ be a Kripke structure for belief.
    We construct a Kripke structure for knowledge and plausibility $M^+
    = (W, \pi, \K_1, \ldots, \K_n, \Plass_1, \ldots, \Plass_n)$ as follows.
    We set $\K_i(w)$ to be the set of worlds where agent $i$'s beliefs
are the same as in $w$.
Formally, $(w,v) \in \K_i$
    if $\B_i(w) = \B_i(v)$. It is easy to verify that $\K_i$ is an
    equivalence relation. We define $\Plass_i(w) =
    (\Omega_{(w,i)},\Pl_{(w,i)})$, where $\Omega_{(w,i)} =
    \B_i(w)$ is the set of worlds
agent $i$ considers possible,
$\Pl_{(w,i)}(\emptyset) = 0$,
and $\Pl_{(w,i)}(A)$ is $1$ if $A
    \subseteq W_{(w,i)}$ is not empty
It is easy to verify that
    these (trivial) plausibility measures are qualitative.

We now prove that $(M,w) \sat \phi$ if and only $(M^+,w) \sat \phi$ for any $\phi
    \in \L^{B}$. This is shown by induction on the structure of
    $\phi$. The only interesting case is if $\phi$ is of the form
    $\Bel_i\phi'$.
Assume $(M,w) \sat \Bel_i\phi'$.
We want to show that $(M^+,w) \sat \Know_i(\True\Condi\phi')$. We
    start by noting that $(w,v) \in \K_i$ if and only if $\B_i(v) =
    \B_i(w)$. This implies that $\Plass_i(v) = \Plass_i(w)$. Thus, $(M^+,
    v) \sat \True\Condi\phi'$ if and only if $(M^+,
    w) \sat \True\Condi\phi'$. Thus, it suffices to show that $(M^+,
    w) \sat \True\Condi\phi'$, since this implies that $(M^+,
    w) \sat \Know_i(\True\Condi\phi')$, \ie $(M^+,w) \sat \Bel_i\phi'$.
There are two cases. If $\B_i(w) = \emptyset$, then $\Omega_{(w,i)} =
    \emptyset$. This implies that $\True\Condi\phi'$ holds vacuously.
If $\B_i(w)$ is not empty, then using the
    induction hypothesis we conclude that $\intension{\phi'}_{(w,i)} =
    \B_i(w)$. From the definition of $\Pl_{(w,i)}$ we conclude that
    $\Pl_{(w,i)}(\intension{\phi'}_{(w,i)}) = 1$ and that
    $\Pl_{(w,i)}(\intension{\neg\phi'}_{(w,i)}) = 0$. Thus, $(M^+,w) \sat
    \True\Condi \phi'$ and hence $(M^+,w) \sat
    \Know_i(\True\Condi\phi')$.
Now assume $(M,w) \sat \neg\Bel_i\phi'$. Then there is some $v \in
    \B_i(w)$ such that $(M,v) \sat\neg\phi'$. Using the induction
    hypothesis we  conclude that
    $\Pl_{(w,i)}(\intension{\neg\phi'}_{(w,i)}) = 1$. Hence, $(M^+,w) \sat
    \neg(\True\Condi \phi')$ and therefore, $(M^+,w) \sat
    \neg\Know_i(\True\Condi\phi')$.

It remains to show that if $M \in \M_K^{et}$ then $M^+$ satisfies CONS,
    and if $M \in \M_K^{est}$, then $M^+$ also satisfies NORM. Assume
    $\B_i$ is transitive and Euclidean. Let $w$ and $v$ be worlds such
    that $(w,v) \in \B_i$.
We claim that $\B_i(w) = \B_i(v)$.
If $(w,t) \in \B_i$, then since $\B_i$ is
    Euclidean  we get that $(v,t) \in \B_i$.
    If $(v,t) \in \B_i$, then since $\B_i$ is transitive we get
    that $(w,t) \in \B_i$. Thus, $\B_i(v) = \B_i(w)$,
as desired.
Recall that  if $\B_i(v) = \B_i(w)$,
    then our construction ensures that $v \in \K_i(w)$. Hence, $\B_i(w)
    \subseteq \K_i(w)$ and
    $M^+$ satisfies CONS. Assume that $\B_i$ is serial. This implies
    that for
all
$w$, $\B_i(w)$ is not empty. Thus, our construction
    guarantees that
    $\Omega_{(w,i)}$ is not empty and  $\Pl_{(w,i)}(\Omega_{(w,i)}) >
    \bottom$.
\eprf

\rethm{thm:LKB}
$\AX^{\sKB}$ (\respc $\AX^{\sKB,\sCONS}$, $\AX^{\sKB,\sCONS,\sNORM}$)
    is a sound and
    complete axiomatization of $\L^{KB}$ with respect to $\M$ (\respc
    $\M^{\sCONS}$, $\M^{\sCONS,\sNORM}$).
\erethm

\prf
Again soundness is straightforward, so we focus on completeness.
    We sketch a completeness proof following the usual Makinson
    \cite{Mak} style of proof. We describe only the
    parts that are different from the standard proofs. See, for
    example, Halpern and Moses \cite{HM2} for details.

In order to prove completeness, we need only show that if the formula
    $\phi$ is consistent with the axiom system (\ie $\AX^{\sKB},
    \AX^{\sKB,\sCONS}$ or $\AX^{\sKB,\sCONS,\sNORM}$) then $\phi$ is
    satisfiable in a Kripke structure of the appropriate class (\ie
    $\M$, $\M^{\sCONS}$, or $\M^{\sCONS,\sNORM}$,
    respectively).

Let $V$ be a set of formulas and $\AX$ an axiom system.
    We say that $V$ is {\em $\AX$-consistent\/} if for all
    $\phi_1,\ldots \phi_n \in  V$, it is not the case that $\AX \vdash
    \neg(\phi_1\land\ldots\land\phi_n)$.
    The set $V$ is a {\em maximal\/} consistent set if it is
    consistent, and for each formula $\phi$, either $\phi \in V$ or
    $\neg\phi \in V$.

We now build a {\em canonical model\/} $M^{\sKB}$ for $\AX^{\sKB}$, in which every
    $\AX^{\sKB}$-consistent formula is satisfiable. $M^{\sKB}$ has a world $w_V$
    corresponding to every maximal $\AX^{\sKB}$-consistent set
$V$ of formulas; we show that $(M^{\sKB},w_V) \sat \phi$ if and only if
$\phi\in V$.

We proceed as follows.
If $V$ is a set of formulas, define
    $V/\Know_i = \{ \phi : \Know_i\phi \in V \}$ and
    $V/\Bel_i = \{ \phi : \Bel_i\phi \in V \}$. Let $M^{\sKB} = (W, \pi,
    \K_1, \ldots, \K_n, \Plass_i, \ldots, \Plass_n)$, where
\begin{itemize}\denselist
 \item $W = \{ w_V : $ V is a maximal $\AX^{\sKB}$-consistent set of
    formulas$ \}$
 \item $\pi(w_V)(p) = $ {\bf true} if and only if $p \in V$
 \item $\K_i = \{ (w_V, w_U) : V/K_i \subseteq U \}$
 \item $\Plass_i(w_V) = (\Omega_{(w_V,i)},\Pl_{(w_V,i)})$, where
    $\Omega_{(w_V,i)} =
    \{ w_U : V/\Bel_i \subseteq U \}$, $\Pl_{(w_V,i)}(\emptyset) =
    0$, and $\Pl_{(w_V,i)}(A) = 1$ for $A \neq \emptyset$.
\end{itemize}

Using standard arguments, it is easy to
    show that the $\K_i$'s are equivalence relations (see \cite{HM2}).
    Using a standard induction argument, we can verify that
    $(M^{\sKB},w_V) \sat \phi$ if and only if $\phi \in V$.

This construction proves completeness for $\AX^{\sKB}$. To prove
    completeness for the other two variants we use the same
    construction, setting $W$ to correspond to the maximal
    $\AX^{\sKB,\sCONS}$-consistent sets (resp.
    $\AX^{\sKB,\sCONS,\sNORM}$-consistent sets). We must show
    that the
    resulting canonical models satisfy CONS and NORM, respectively.

Let $M^{\sKB,\sCONS}$ be the canonical model constructed
for
    $\AX^{\sKB,\sCONS}$. To show that
    $M^{\sKB,\sCONS}$  satisfies CONS, it is enough to show that
    $V/\Know_i \subseteq
    V/\Bel_i$. To show this, assume $\phi \in V/\Know_i$. Then
    $\Know_i\phi \in V$.
    Since KB2 $\in \AX^{\sKB,\sCONS}$, we conclude that $\Bel_i\phi\in
    V$, and thus
    $\phi \in V/\Bel_i$.

Let $M^{\sKB,\sCONS,\sNORM}$ be the canonical model constructed
for
    $\AX^{\sKB,\sCONS,\sNORM}$. The argument above shows
    that $M^{\sKB,\sCONS,\sNORM}$ satisfies
    CONS.  To show that it satisfies NORM, \ie
    $\Pl_{(w,i)}(\Omega_{(w,i)}) > \bottom$,
it is enough to
    show that $V/\Bel_i$ is consistent, for then there must be some
    $U$ such that $V/\Bel_i \in U$. Assume, by way of contradiction,
    that $V/\Bel_i$ is
    inconsistent. Then there are formulas $\phi_1,\ldots,\phi_m \in
    V/\Bel_i$ such that $\vdash \neg(\phi_1\land\ldots\land\phi_m)$.
    Since $\phi_1,\ldots,\phi_n \in V/\Bel_i$, we conclude that
    $\Bel_i\phi_1, \ldots, \Bel_i\phi_m \in V$. Using the
    K45 axioms for $\Bel_i$, standard arguments show that
    $\Bel_i(\phi_1,\ldots,\phi_n) \in V$, and hence that
    $\Bel_i(\False) \in V$, which contradicts the consistency of $V$.
\eprf

\relem{lem:MosesShoham}
 Let $M$ be a propositional Kripke structure of knowledge and
    plausibility satisfying CONS and SDP.
Suppose that $w$, $i$, and $\alpha$ are such that the most plausible
    worlds in $\Plass_i(w)$ are exactly those worlds in $\K_i(w)$
    that satisfy $\alpha$, \ie
    $\MP(\Plass_i(w)) = \{ w' \in \K_i(w) : (M,w') \sat \alpha \}$.
Then for any formula $\phi\in\L^{KB}$ that includes only the
    modalities $\Know_i$ and $\Bel_i$, $(M,w)
     \models \phi$ if and only if $(M,w) \models
     \phi^*$, where $\phi^*$ is the result of recursively replacing each
    subformula of the form
    $\Bel_i\psi$ in $\phi$ by $\Know_i(\alpha\rimp\psi^*)$.
\erelem

\prf
We prove by induction that for any $w' \in \K_i(w)$, $(M,w')
    \sat\phi$ if and only if $(M,w') \sat \phi^*$. The only
    interesting case is if $\phi$ has the from $\Bel_i\phi'$.
Suppose that $(M,w') \sat \Bel_i\phi'$. This
    implies that $(M,w') \sat \True\Condi \phi'$, \ie for
    all $w'' \in \MP(\Plass_i(w'))$ we have $(M,w'') \sat \phi'$. Now let $w'' \in
    \K_i(w')$. If $(M,w'') \sat \neg\alpha$, then $(M,w'') \sat
    \alpha\rimp (\phi')^*$. If $(M,w'') \sat \alpha$ then, by definition,
    $w'' \in \MP(\Plass_i(w))$, and  since we assumed
    SDP, $\MP(\Plass_i(w')) = \MP(\Plass_i(w))$. Thus, we conclude that
    $(M,w'') \sat \phi'$, and using the induction hypothesis we get that
    $(M,w'') \sat (\phi')^*$. We
    conclude that all worlds in $\K_i(w')$ satisfy $\alpha\rimp(\phi')^*$,
    and thus $(M,w') \sat \Know_i(\alpha\rimp(\phi')^*)$.
Now assume that $(M,w') \sat \Know_i(\alpha\rimp(\phi')^*)$. Let $w''$ be
    any world in $\K_i(w')$. Since we assumed SDP, we have that
    $\MP(\Plass_i(w'')) = \MP(\Plass_i(w))$ is the set of worlds in $\K_i(w)$
    that satisfy $\alpha$. We conclude, using our induction
    hypothesis, that all worlds in $\MP(\Plass_i(w''))$ satisfy $\phi'$.
    Hence, $(M,w'') \sat \True\Condi\phi'$. Since this is true for all
    $w'' \in \K_i(w')$ we conclude that $(M,w') \sat \Bel_i\phi'$.
\eprf

\subsection{Proofs for Section~\protect\ref{sec:axiomfull}}
\label{prf:axiomfull}

\rethm{thm:ax}
\AX\ is a sound and complete axiomatization for $\L^{KC}$
    with respect to $\M$.
\erethm

\prf
Again, we just describe the
    completeness proof. This proof draws on the usual completeness
    proofs for S5 modal logic, and the completeness proof for
    conditional logic
described in
\cite{FrThesis,FrH5Full}.

\renewcommand{\Intension}[1]{[#1]_{(w_V,i)}}

We proceed as follows.
If $V$ is a set of formulas, define
    $V/\Know_i = \{ \phi : \Know_i\phi \in V \}$ and
    $V/\PBox_i = \{ \phi : \PBox_i\phi \in V \}$.
We define a canonical model $M^c = (W, \pi,
    \K_1, \ldots, \K_n, \Plass_i, \ldots, \Plass_n)$ as follows:
\begin{itemize}\denselist
 \item $W = \{ w_V : $ V is a maximal AX-consistent set of formulas$ \}$
 \item $\pi(w_V)(p) = $ {\bf true} if and only if $p \in V$
 \item $\K_i = \{ (w_V, w_U) : V/K_i \subseteq U \}$
 \item $\Plass_i(w_V) = (\Omega_{(w_V,i)}, \F_{(w_V,i)},\Pl_{(w_V,i)})$,
where
\begin{itemize}\denselist
 \item $\Omega_{(w_V,i)} = \{ w_U : V/\PBox_i \subseteq U \}$,
 \item $\F_{(w_V,i)} = \{ \Intension{\phi} :  \phi \in
    \L^{KC}\}$ where
    $\Intension{\phi} = \{ w_U \in W_{(w_V,i)} : \phi \in U \}$, and
 \item $\Pl_{(w_V,i)}$ is such that $\Pl_{(w_V,i)}(\Intension{\phi})
    \le \Pl_{(w_V,i)}(\Intension{\psi})$ if and only if
    $(\phi\lor\psi) \Condi \psi \in V$.
\end{itemize}
\end{itemize}

We need to verify that $M^c$ is indeed a structure in $\M$. Using
standard
    arguments it is easy to show that the $\K_i$ relations are
    equivalence relations. In
\cite{FrThesis,FrH5Full}
we prove that $\Plass_i(w_V)$ is
    a well-defined qualitative plausibility space.

Finally, we have to show that $(M^c,w_V) \sat \phi$ if and only if $\phi
    \in V$. As usual, this is done by induction on the structure of
    $\phi$. We use the standard argument
    for formulas of the form $\Know_i\phi$ and arguments from
    \cite{FrThesis,FrH5Full}
for formulas of the from $\phi\Condi\psi$.
We omit the details here.
\eprf

\rethm{thm:ax-extend}
Let $\A$ be a subset of $\{ \RANK, \NORM, \REF, \UNIF, \CONS, \SDP \}$
    and let $A$
    be the corresponding subset of $\{$C5, C6, C7, C8, C9, C10$\}$. Then
    $\mbox{AX} \cup A$ is a sound and
    complete axiomatization with respect to the structures in $\M$
    satisfying $\A$.
\erethm

\prf
Yet again, we focus on completeness.
We obtain completeness in each case by modifying the proof of
    \Tref{thm:ax}. We construct a canonical model as in that proof,
    checking consistency with the extended axiom system. The resulting
structure
    is in $\M$ and has the property that $(M,w_V) \sat \phi$ if
    and only if $\phi\in V$. We just need to show that this
structure also
    satisfies the corresponding semantic restrictions.

First, we consider CONS and axiom C9. Assume that C9 is included as an
    axiom. It is easy to see that this implies that $V/\PBox_i
    \subseteq V/\Know_i$. This implies that $\Omega_{(w_V,i)}
    \subseteq \K_i(w_V)$ in our construction.

Now consider the relationship between SDP and C10. Assume that C10 is
    included as an axiom. We need to show that if $w_U \in \K_i(w_V)$,
    then $\Plass_i(w_U) = \Plass_i(w_V)$. It is enough to show that
    $\phi\Condi\psi \in V$ if and only if $\phi\Condi\psi \in U$,
    since these statements determine $\Plass_i$ in our construction.
    Assume $\phi\Condi\psi \in V$. Then, according to C10,
    $\Know_i(\phi\Condi\psi) \in V$, and thus $\phi\Condi\psi\in
    V/\Know_i$. Recall that $w_U \in \K_i(w_V)$ only if $V/\Know_i
    \subseteq U$. We conclude that $\phi\Condi\psi \in U$. The other
    direction follows from the fact that $\K_i$ is symmetric in our
    construction, and thus $w_V \in \K_i(w_U)$.

The desired relationship between RANK, NORM, REF, and UNIF
    and the axioms C5, C6, C7, and C8 is proved in
    \cite{FrThesis,FrH5Full},
for a logic that does not mention knowledge.
    Since these  conditions put restrictions on $\Plass_i(w)$ and do
not involve
    knowledge, the proof of
\cite{FrThesis,FrH5Full}
goes through unchanged; we
    do not repeat it here.
\eprf

\rethm{thm:small-model}
Let $\A$ be a subset of $\{ \CONS, \NORM, \REF, \SDP, \UNIF, \RANK \}$. The
    formula $\phi$ is satisfiable in a Kripke structure satisfying $\A$
    if and only if it is satisfiable in a Kripke structure
    with at most $2^{\Card{\sSub(\phi)}}$ worlds.
\erethm

\prf
The proof of this theorem relies on techniques from \cite{FrH3Full}. We
    sketch only the main steps here. The proof is based on a standard
    filtration argument.

Suppose there is a structure $M$ and a world $w$ in $M$ such that $(M,w)
\sat \phi$.
    Let $\Subp(\phi) =
    \Sub(\phi) \union \{ \neg\phi : \phi \in \Sub(\phi) \}$. We say
    that $V \subseteq \Subp(\phi)$ is an {\em atom \/} if for each
    $\phi \in \Sub(\phi)$, either $\phi \in V$ or $\neg\phi \in V$. We
    say that a world $w$ in $M$ satisfies an atom $V$ if for all
    $\phi \in V$, we have $(M,w) \sat \phi$. It is easy to see that each
    world satisfies exactly one atom. Given a world $w'$, we define
    $[w]$ to be the equivalence class containing all worlds that satisfy
    the same atom as $w$. For each equivalence class $[w]$, we
    arbitrarily choose a {\em representative world\/} $w_{[w]} \in [w]$.
We define $M' =
    (W',\pi',\K'_1,\ldots \K'_n,\P'_1, \ldots, \P'_n)$, where $W' =
    \{ [w] : w \in W \}$, $\pi'([w])
    = \pi(w_{[w]})$, $\K'_i = \{
    ([w],[w']) : (w,w') \in \K_i \}$, and $\P'_i([w]) =
    (\Omega'_{([w],i)}, \Pl'_{([w],i)})$, where $\Omega'_{([w],i)} =
    \{ [w'] : w' \in \Omega_{(w_{[w]},i)} \}$ and $\Pl'_{([w],i)}(A)
    \le \Pl'_{([w],i)}(B)$ if $\Pl_{(w_{[w]},i)}( A^* \inter
    \Omega_{(w_{[w]},i)}) \le \Pl_{(w_{[w]},i)}( B^* \inter
    \Omega_{(w_{[w]},i)})$, where $A^* = \{ w'' : \exists [w'] \in A,
    w'' \in [w'] \}$.
Arguments essentially identical to those of \cite{FrH3Full}
show that $(M',[w]) \sat \psi$ if and only if $(M,w)
    \sat \psi$ for all $\psi \in \Sub(\phi)$;
we omit details here.

We now have to describe how to modify this argument to ensure that
    $M'$ satisfies $\A$.
The modifications for $\NORM,\REF,\UNIF$ and $\RANK$ are described in
    \cite{FrH3Full}.
Suppose that $M$ satisfies
    CONS. Let $[w'] \in \Omega'_{([w],i)}$. By definition, $w' \in
    \Omega_{(w_{[w]},i)}$. But since $M$ satisfies CONS, we have that $w'
    \in \K_i(w_{[w]})$. By definition, we get
    that $[w'] \in \K'_i([w])$. We conclude that $M'$ satisfies CONS.
Finally,
suppose that $M$ satisfies SDP.  We force $M'$ to satisfy SDP as follows.
For all worlds $w$, we choose a representative world
$w_{\K_i([w])} \in \K'_i([w])$
such that if $(w,w') \in K_i$, then $w_{\K_i([w])} = w_{\K_i([w'])}$.
We then
    modify the construction so that,
for each world $v \in \K_i(w)$, we have
$\P'_i([v]) = \P'_i(w_{\K_i([w])})$.
It is easy to see that for all
    $\psi\Condi\chi \in \Sub(\phi)$, we have that $(M,w) \sat
    \psi\Condi\chi$ if
    and only if $(M,w_{\K_i([w])}) \sat \psi\Condi\chi$.
Thus, it is easy to show that after this modification we still
    have that $(M',[w]) \sat \psi$ if and only if $(M,w)
    \sat \psi$ for all $\psi \in \Sub(\phi)$.
\eprf

\rethm{thm:small-model-one-agent}
Let $\A$ be a subset of $\{ \CONS, \NORM, \REF, \SDP, \UNIF, \RANK \}$
    containing CONS and either SDP or UNIF. If $\phi$ talks about the
    knowledge and plausibility of only one agent, then $\phi$ is
    satisfiable in a Kripke structure satisfying $\A$ if and only if it
    is satisfiable in a preferential Kripke structure satisfying $\A$
    with at most
    $\Card{\Sub(\phi)}^3$ worlds.
\erethm

\prf
Assume $M = (W,\pi,\K_1,\Plass_1)$ is a structure satisfying $\phi$. Since
    CONS is in $\A$, we must have that $\Omega_{(w,1)} \subseteq
    \K_1(w))$. Without loss of
    generality, we can assume that $\K_1$ consists of one equivalence
    class, that is, that $\K_1 = W \cross W$. Since CONS and SDP imply
    UNIF, and since $\A$ contains CONS and either SDP or UNIF, we
    conclude that $M$ satisfies UNIF.
Using techniques from \cite{FrH3Full} we can assume, without loss of
    generality, that for each world $w$,
the plausibility space
$\Plass_1(w)$ is preferential
    (\ie induced by some preference ordering) and
    that $\Omega_{(w,1)}$ has at most $\Card{\Sub(\phi)}^2$ worlds.

Choose $w_0 \in W$ such that $(M,w_0) \sat \phi$. For each formula
    $\neg\Know_1\psi \in \Sub(\phi)$ such that $(M,w_0) \sat
    \neg\Know_1\psi$, we select a world $w_\psi$ such that $(M,w_\psi)
    \sat \neg\psi$. Let $T$ be $\{w_0\} \union \{ w_\psi :
    \neg\Know_1\psi \in \Sub(\phi) \}$. Note that the cardinality of
    $T$ is at most $\Card{\Sub(\phi)}$. Define $M' =
    (W',\pi',\K'_1,\P'_1)$ by
    taking $W'$ to be the union of $\Omega'_{(w,1)}$ for each $w \in T$,
    taking $\pi'$ to be $\pi$ restricted to $W'$, and taking
    $\P'_1(w) = \Plass_1(w)$.
Clearly $\Card{W'}$ is at most $\Card{\Sub(\phi)}^3$.  A straightforward
argument for all subformulas $\psi$ of $\phi$ and all worlds $w' \in
W'$, we have  $(M,w') \sat \psi$ if and only if $(M',w') \sat\psi$.  It follows
that $(M',w_0) \sat \phi$, so $\phi$ is satisfiable in a small
preferential structure.
\eprf

\rethm{thm:complex}
Let $\A$ be a subset of $\{ \CONS, \NORM, \REF, \SDP, \UNIF, \RANK \}$. If
    $\CONS \in \A$, but it is not the case that UNIF or SDP is in $\A$,
    then the validity problem with respect to structures satisfying $\A$
    is complete for exponential time. Otherwise, the validity problem
    is complete for polynomial space.
\erethm

\prf
The proof combines ideas from \cite{FH3,FrH3Full,HM2}.
We briefly
    sketch the main ideas here, referring the reader to the other papers
for details.

The polynomial space lower bound follows from the polynomial space
    lower bound for logics of knowledge alone \cite{HM2}. For the
    exponential lower bound we use exactly the lower bound described
    Fagin and Halpern \cite{FH3} for the combination of knowledge and
    probability (which is in turn based on the lower bound for PDL
    \cite{FL}). This lower bound construction uses only formulas
    involving $\Know_i$ and probabilistic statements of the form
    $w_i(\phi) = 1$ (\ie the probability of $\phi$ is 1).
    Since $\PBox_i\phi$ has exactly the same properties as $w_i(\phi)
    = 1$, the same construction applies to our logic.

In the cases where we claim a polynomial space upper bound, this is
    shown by proving that if a formula $\phi$ is satisfiable at all,
    it is satisfiable in a structure that looks like a tree, with
    polynomial branching and depth no greater than the depth of
    nesting of $\K_i$ and $\Condi$ operators in $\phi$. The result now
    follows along similar lines to corresponding results for logics of
    knowledge.

Finally, the exponential time upper bound follows by showing that if a
    formulas is satisfiable at all, it is satisfiable in an
    exponential size structure that can be constructed in deterministic
    exponential time; the technique is similar to that used to show
    that logics of knowledge with common knowledge are decidable in
    deterministic exponential time \cite{HM2} or that PDL is decidable
    in deterministic exponential time \cite{Pratt79}.
\eprf

\rethm{thm:complex-one-agent}
Let $\A$ be a subset of $\{ \CONS, \NORM, \REF, \SDP, \UNIF, \RANK \}$
    containing CONS and either UNIF or SDP. For the case of one agent,
    the validity problem in structures satisfying $\A$ is co-NP-complete.
\erethm

\prf
We show that the satisfiability problem is NP-complete. It follows
    that the validity problem is co-NP-complete. The lower bound is
    immediate, since clearly the logic is at least as hard as
    propositional logic. For the upper bound, by
    \Tref{thm:small-model-one-agent}, $\phi$ is satisfiable in a
    structure satisfying $\A$ if and only if $\phi$ is satisfiable in
    a structure $M$ of size polynomial in $\Card{\phi}$. We simply
    guess a structure $M$ and check that $\phi$ is satisfiable. It
    is easy to show that model checking can be done in polynomial time
    (see \cite{HM2,FrH3Full}).
\eprf

\subsection{Proofs for Section~\protect\ref{sec:axiomtime}}
\label{prf:axiomtime}

\rethm{thm:ax-time}
The axiom system \AXT\ is a sound and complete axiomatization of
$\L^{KCT}$ with respect to $\Sysclass$.
\erethm

\prf
As usual, we focus on completeness. Again, we construct a
canonical
interpreted
system $\Sys$ such that if $\phi \in \L^{KCT}$ is
consistent, then $\phi$ is satisfied in $\Sys$. The outline of the
proof is similar to that of \Tref{thm:ax}.

We proceed as follows. Let $V$ be a maximal \AXT-consistent set of
formulas in $\L^{KCT}$. We define $V/\Next = \{ \phi : \Next \phi \in V
\}$. We claim that $V/\Next$ is also a maximal \AXT-consistent set.
To show that $V/\Next$ is maximal,
assume that $\phi \not\in V/\Next$. Then $\Next \phi \not \in
V$. From axiom T2, we have that $\Next\neg\phi \in V$, and thus, $\neg
\phi \in V/\Next$. This shows that $V/\Next$ is maximal. To show that
$V/\Next$ is \AXT-consistent, assume that there are formulas
$\phi_1,\ldots \phi_n \in V/\Next$ such that $\vdash_{\sAXT}
\neg(\phi_1\land\ldots\land\phi_n)$. From K1, T1 and RT1 we get that
$\false \in V/\Next$. Thus, $\Next \false \in V$. Using T2 we get
that $\neg \Next\True \in V$. Using RT1, however, we get that $\Next
\True \in V$, which contradicts the assumption that $V$ is
consistent. Thus, $V/\Next$ is \AXT-consistent.
Finally, we define $V/\Next^m$ to the
result of $m$ applications of $/\Next$. Repeated applications of the
above argument show that
$V/\Next^m$ is a maximal \AXT-consistent set for all $m \ge 0$.

\newcommand{\Intensiont}[1]{[#1]_{(r^V,m,i)}}

We construct a canonical
interpreted
system as follows. Let $\Sys = (\R, \pi,
\Plass_1,\ldots,\Plass_n)$, where
\begin{itemize}\denselist
 \item $\R = \{ r^V : V \subseteq \L^{KCT} \mbox{ is a maximal
 \AXT-consistent set}\}$ such that
\begin{itemize}\denselist
 \item $r^V_e(m) = V/\Next^m$, and
 \item $r^V_i(m) = (V/\Next^m)/\Know_i$,
\end{itemize}
 \item $\pi(r^V,m)(p) = $ {\bf true} if and only if $p \in r^V_e(m)$, and
 \item $\Plass_i(r^V,m) = (W_{(r^V,m,i)}, \Pl_{(r^V,m,i)})$, where
\begin{itemize}\denselist
 \item $W_{(r^V,m,i)} = \{ (r^U,n) : (V/\Next^m)/\PBox_i \subseteq
 U/\Next^n$ \}, and
 \item $\Pl_{(r^V,m,i)}$ is such that $\Pl_{(r^V,m,i)}(\Intensiont{\phi})
    \le \Pl_{(r^V,m,i)}(\Intensiont{\psi})$ if and only if
    $(\phi\lor\psi) \Condi \psi \in V/\Next^m$, where
$\Intensiont{\phi} = \{ (r^U,k) \in W_{(r^V,m,i)} : \phi \in U/\Next^k \}$.
\end{itemize}
\end{itemize}

Using the arguments in the completeness proof for conditional logic of
\cite{FrThesis,FrH5Full},
we can show
that $\Plass_i(r,m)$ is well-defined for all $i$.  Finally, we have to
show that $(\Sys,r^V,m) \sat \phi$ if and only if $\phi \in
r^V_e(m)$. As usual, this is done by induction on the structure of
$\phi$. This is identical to the proof in of
\Tref{thm:ax} except for the $\Next$ modality, which is handled by
standard arguments. We omit the details here.
\eprf

\rethm{thm:ax-time-extend}
Let $\A$ be a subset of $\{ \RANK, \NORM, \REF, \UNIF, \CONS, \SDP \}$
    and let $A$
    be the corresponding subset of $\{$C5, C6, C7, C8, C9, C10$\}$. Then
    $\AXT \cup A$ is a sound and
    complete axiomatization with respect to systems in $\Sysclass$
    satisfying $\A$.
\erethm

\prf
Again, we focus on completeness. We obtain completeness in each case
    by modifying the proof of \Tref{thm:ax-time}. We construct a
    canonical system as in that proof, checking consistency with the
    extended axiom system. The resulting system has the property that
    $(\Sys,r^V,m) \sat \phi$ if and only if $\phi \in
    V/\Next^m$. We just need to show that this system satisfies the
    corresponding semantic restrictions. The desired relationship
    between these semantic properties and axioms is proved in
\cite{FrThesis,FrH5Full} and the proof of
Theorem~\ref{thm:ax-extend}.
\eprf

\subsection{Proofs for
Section~\protect\ref{sec:cond-as-minimal-change}}
\label{prf:cond-as-minimal-change}

\repro{pro:bel-change}
Let $\Sys$ be a  synchronous system satisfying perfect recall and PRIOR.
    If $\phi$ characterizes agent $i$'s knowledge at $(r,m+1)$ with
    respect to his knowledge at $(r,m)$, then
$(\Sys,r,m+1) \sat \psi \Condi \xi$ if and only if
$(\Sys,r,m) \sat \Next(\phi\land\psi) \Condi \Next\xi$.
\erepro

\prf
Expanding the
    definition we get that
    $\R(\intension{\Next(\phi\land\psi)}_{(r,m)}) = \{  r' \in
    \Omega_{(r,i)} :
    (r',m) \sim_i (r,m), (r',m+1) \sat \phi\land\psi \}$.
    Similarly, we get that $\R(\intension{\psi}_{(r,m+1)}) = \{
    r' \in \Omega_{(r,i)}: (r',m+1) \sim_i (r,m+1), (r',m+1) \sat
    \psi \}$. However, since $\phi$ characterizes agent $i$'s
    knowledge at time $m+1$ with respect to his knowledge at time $m$,
    we get that $(r',m+1) \sim_i (r,m+1)$ if
    and only if $(r',m) \sim_i (r,m)$ and $(r,m+1) \sat \phi$. We
    conclude that $\R(\intension{\Next(\phi\land\psi)}_{(r,m)}) =
    \R(\intension{\psi}_{(r,m+1)})$. The lemma now
    follows directly from \Pref{pro:local-change}.
\eprf

\relem{lem:BatBon}
Let $\Sys$ be a synchronous static system satisfying PRIOR, RANK, SDP, and
perfect recall that has finite branching. Then $(\Sys,r,m) \sat
\Bel_i\phi \equiv \Bel_i\Next\Bel_i\phi$ for all propositional
formulas $\phi$.
\erelem

\prf
For all points $(r,m)$
in $\Sys$,
note that $\Omega_{(r,m,i)} = \cup \{ A_{\psi}
\}$, where $A_\psi$ is the set of
points $(r',m) \sim_i (r,m)$
such that
the agent's new knowledge at time
$m+1$ is $\psi$. If $\Sys$ has finite branching, this is a finite
partition of $\Omega_{(r,m,i)}$. Additionally, note that if
$\Pl_{(r,m,i)}$ is a ranking, and $C_1,\ldots, C_k$ is a finite
partition of $C$, then since $\Pl_{(r,m,i)}(C) = \max_{1\le j\le k}
\Pl_{(r,m,i)}(C_j)$,
there must be some
$j$ such that
$\Pl_{(r,m,i)}(C_j) = \Pl_{(r,m,i)}(C)$. In particular,
for all $C \subseteq W_{(r,m,i)}$,
either $\Pl_{(r,m,i)}(C) = \top$ or $\Pl_{(r,m,i)}(W_{(r,m,i)} - C) =
\top$.

For the ``$\rimp$'' part, suppose that $(\Sys,r,m) \sat
\Bel_i\phi$.
If $\Pl_{(r,m,i)}(W_{(r,m,i)}) =
\bot$, then $(\Sys,r,m) \sat \Bel_i\Next\Bel_i\phi$ vacuously. If
$\Pl_{(r,m,i)}(W_{(r,m,i)}) \neq \bot$, then
$\Pl_{(r,m,i)}(\intension{\phi}_{(r,m,i)}) >
\Pl_{(r,m,i)}(\intension{\neg\phi}_{(r,m,i)})$.
Assume that $\psi$ is such that $\Pl_{(r,m,i)}(A_\psi) = \top$. It is
easy to verify that since $\Pl_{(r,m,i)}$ is a ranking, we get
that  $\Pl_{(r,m,i)}(A_\psi \inter
\intension{\phi}_{(r,m,i)}) > \Pl_{(r,m,i)}(A_\psi \inter
\intension{\neg\phi}_{(r,m,i)})$.
Let $r'$ be a run such that $(r',m) \in A_\psi$.
By SDP,
we get that $\Pl_{(r,m,i)} = \Pl_{(r',m,i)}$, and thus
$\Pl_{(r',m,i)}(A_\psi \inter
\intension{\phi}_{(r',m,i)}) > \Pl_{(r',m,i)}(A_\psi \inter
\intension{\neg\phi}_{(r',m,i)})$.
By definition of $A_\psi$, we have that $(r'',m+1) \sim_i (r',m+1)$ if
and only if $(r'',m) \in A_\psi$. Since $\Sys$ satisfies PRIOR,
$\Pl_{(r',m+1,i)}$ is the
result of conditioning $\Pl_{(r,m,i)}$ on $A_\psi$.
Moreover, since propositions are static, we get that
$\Pl_{(r',m+1,i)}(\intension{\phi}_{(r',m+1,i)}) >
\Pl_{(r',m+1,i)}(\intension{\neg\phi}_{(r',m+1,i)})$. Thus,
$(\Sys,r',m) \sat \Next\Bel_i\phi$. We conclude that $A_\psi \subseteq
\intension{\Next\Bel_i\phi}_{(r,m,i)}$, and thus
$\Pl_{(r,m,i)}(\intension{\Next\Bel_i\phi}_{(r,m,i)}) = \top$.
Moreover, since $A_\psi \subseteq
\intension{\Next\Bel_i\phi}_{(r,m,i)}$ for all $A_\psi$ such that
$\Pl_{(r,m,i)}(A_\psi) = \top$, we get that
$\Pl_{(r,m,i)}(\intension{\neg\Next\Bel_i\phi}_{(r,m,i)}) \le
\max \{ \Pl_{(r,m,i)}(A_\psi) : \Pl_{(r,m,i)} ( A_\psi ) < \top \} <
\top$. We conclude that
$\Pl_{(r,m,i)}(\intension{\Next\Bel_i\phi}_{(r,m,i)}) >
\Pl_{(r,m,i)}(\intension{\neg\Next\Bel_i\phi}_{(r,m,i)})$, and thus,
$(\Sys,r,m) \sat \Bel_i\Next\Bel_i\phi$.

{\sloppy
For the ``$\limp$'' part, suppose that $(\Sys,r,m) \sat
\Bel_i\Next\Bel_i\phi$.
If
$\Pl_{(r,m,i)}(W_{(r,m,i)}) = \bot$, then $(\Sys,r,m) \sat
\Bel_i\phi$ vacuously. If
$\Pl_{(r,m,i)}(W_{(r,m,i)}) \neq \bot$, then
$\Pl_{(r,m,i)}(\intension{\Next\Bel_i\phi}_{(r,m,i)})$ $>$\linebreak
$\Pl_{(r,m,i)}(\intension{\neg\Next\Bel_i\phi}_{(r,m,i)})$. Thus, since
$\Pl_{(r,m,i)}$ is a ranking,
$\Pl_{(r,m,i)}(\intension{\Next\Bel_i\phi}_{(r,m,i)}) = \top$.
Let $(r',m)$ be some point in $A_\psi$ for some $\psi$.
By SDP, we have that $(\Sys,r',m) \sat \Next\Bel\phi$ if
and only if $(\Sys,r'',m) \sat \Next\Bel\phi$ for all points $(r'',m)
\in A_\psi$. Thus,
$\intension{\Next\Bel_i\phi}_{(r,m,i)} = A_{\psi_1} \union \ldots
\union A_{\psi_k}$ for some $\psi_1, \ldots, \psi_k$.
Since $\Pl_{(r,m,i)}(\intension{\Next\Bel_i\phi}_{(r,m,i)}) >
\Pl_{(r,m,i)}(\intension{\neg\Next\Bel_i\phi}_{(r,m,i)})$,
we get that $\Pl_{(r,m,i)}(A_\psi) = \top$ only if $\psi = \psi_j$ for
some $1 \le j \le k$. Moreover, since $A_{\psi_1},\ldots, A_{\psi_k}$
is a finite partition of $\intension{\Next\Bel_i\phi}_{(r,m,i)}$,
there must be at least one $1 \le j \le k$ such that
$\Pl_{(r,m,i)}(A_{\psi_j}) = \top$.
Let $\psi_j$ be such that $\Pl_{(r,m,i)}(A_{\psi_j}) = \top$.
Suppose that $(r',m) \in A_{\psi_j}$. Then we have that
$\Pl_{(r',m+1, i)}(\intension{\phi}_{(r',m+1,i)}) >
\Pl_{(r',m+1,i)}(\intension{\neg\phi}_{(r',m+1,i)})$.  Since
$\Sys$ is synchronous, static, and satisfies perfect recall, PRIOR,
and SDP, we get that $\Pl_{(r,m,i)}(A_{\psi_j} \inter
\intension{\phi}_{(r,m,i)}) >
\Pl_{(r,m,i)}(A_{\psi_j} \inter
\intension{\neg\phi}_{(r,m,i)})$.
Since $\Pl_{(r,m,i)}$ is a ranking, we get that
$\Pl_{(r,m,i)}(A_{\psi_j} \inter
\intension{\phi}_{(r,m,i)}) = \top$, and
thus, $\Pl_{(r,m,i)}(\intension{\phi}_{(r,m,i)}) = \top$.
Finally, if $\Pl_{(r,m,i)}(A_\psi) < \top$, then
$\Pl_{(r,m,i)}(A_{\psi} \inter \intension{\neg\phi}_{(r,m,i)}) <
\top$. Thus,  since $\Pl_{(r,m,i)}(\intension{\neg\phi}_{(r,m,i)}) =
\max_{\psi}  \Pl_{(r,m,i)}(A_{\psi} \inter
\intension{\neg\phi}_{(r,m,i)})$, we get that
$\Pl_{(r,m,i)}(\intension{\neg\phi}_{(r,m,i)}) < \top$.
We conclude that $(\Sys,r,m) \sat \Bel_i\phi$.
}
\eprf

\subsection{Proofs for Section~\protect\ref{sec:prior-prop}}
\label{prf:prior-prop}

\rethm{thm:coherence}
Let $\A$ be a subset of $\{ \QUAL, \NORM, \REF, \RANK \}$ and
let $\Sys$ be a coherent synchronous system satisfying perfect recall, CONS,
    and $\A$. Then there is a synchronous system $\Sys'$
    satisfying perfect recall, PRIOR, and $\A$, and a mapping $f:\R
\mapsto
    \R'$ such that for all temporally linear formulas $\phi\in\L^{KCT}$,
    we have $(\Sys,r,m)\sat \phi$ if and only if $(\Sys',f(r),m) \sat \phi$.
\erethm

\newcommand{\PLex}{\oplus}
\prf
\commentout{
We start with some technical results that we will need later.
Let $S_1 = (W_1,\Pl_1)$ and $S_2 = (W_2,\Pl_2)$ be two
    plausibility spaces.
Recall that $S_1$ and $S_2$ are (order) isomorphic if there
    is a bijection $h$ from $W_1$ to $W_2$ such that,
for $A, B \subseteq W_1$, we have $\Pl_1(A) \le \Pl_1(B)$ if and only if
$\Pl_2(h(A)) \le \Pl_2(h(B))$.
We actually need a slightly more general notion here.
    Let $P = (W,\Pr)$ be a probability space.  A set $A$ is called a
    {\em support\/} of $P$, if $\Pr(\overline{A}) = 0$. We can
    define a similar notion for plausibility spaces. Let $S = (W,\Pl)$ be a
    plausibility space. We say that $A \subseteq W$ is a {\em
    support\/} of $S$, if for all $B \subseteq W$, $\Pl(B) = \Pl(B
    \inter A)$. Thus, only $B \inter A$ is relevant for determining
    the plausibility of $B$. This certainly implies that
    $\Pl(\overline{A}) = \bot$, since we must have $\Pl(\overline{A})
    = \Pl(A \inter \overline{A}) = \Pl(\emptyset)$, but the converse
    does not hold in general. In probability spaces,
    $\Pr(\overline{A}) = 0$ implies that $\Pr(B) = \Pr(B \inter A)$
    for all $B$, but the analogous condition does not hold for
    arbitrary plausibility spaces.
Let $S = (W,\Pl)$ be a plausibility space. Define
    $S|_A = (W|_A, \Pl|_A)$ where $W|_A = W \inter A$ and $\Pl|_A$ is
    the restriction of $\Pl$ to $W|_A$.
We say that two plausibility spaces $S_1$ and $S_2$ are
{\em essentially (order) isomorphic\/} if there are supports $C_1$ and
    $C_2$ of $S_1$ and $S_2$, respectively,
such that $S_1|_{C_1}$ is isomorphic to $S_2|_{C_2}$.
It is easy to see that, as expected, essential isomorphism
    defines an equivalence relation among plausibility spaces.
    Moreover, essential isomorphism preserves all the properties of
    plausibility spaces
 that we defined above, such as QUAL and RANK.
Finally, it is easy to see that if $S = (W,\Pl)$, then $(W,\Pl(\cdot |
    A))$ is essentially
    isomorphic to $S|_A$ when we use any conditioning method
    that satisfies COND.
}
To construct $\Sys'$, we use a general technique for taking a
``sum'' of a sequence of plausibility spaces.
Let $\lambda$ be an ordinal and let $\{ S_i : 0 \le i < \lambda \}$ be a
    sequence of
    plausibility spaces, where $S_i = (\Omega_i,\Pl_i)$ and the
$\Omega_i$'s
    are pairwise disjoint.
Define $\PLex_i S_i$ as
    $(\union_i\Omega_i, \Pl_{\PLex S_i})$, where
    $\Pl_{\PLex S_i}(A) \ge \Pl_{\PLex S_i}(B)$ if either $\Pl_i(A
 \inter \Omega_i) = \Pl_i(B \inter \Omega_i) = \bot$ for all $i$, or
there exists some $i$ such that
    $\Pl_i(A \inter \Omega_i) \ge \Pl_i(B \inter \Omega_i)$,
$\Pl_i(A \inter \Omega_i) > \bottom$, and
 $\Pl_j(A \inter \Omega_j) = \Pl_j(B \inter \Omega_j) = \bot$ for
    all $j < i$.
We can think of $\PLex_i S_i$ as a lexicographic
   combination of the $S_i$'s.

\Lem\label{lem:Plex}
\begin{enumerate}\denselist
 \item[(a)] $\PLex_i S_i$ is a plausibility space,
 \item[(b)] if $S_i$ is qualitative for all $i$, then $\PLex_i S_i$ is
    qualitative,
 \item[(c)] if $S_i$ is ranked for all $i$, then $\PLex_i S_i$ is
    ranked,
 \item[(d)]
($\PLex_i S_i)|_C$
is isomorphic to
$\PLex_i (S_i|_C)$
under the identity mapping.
\item[(e)] $(\PLex_i S_i)|_{W_j}$ is
isomorphic to $S_j$
    under the identity mapping.
\item[(f)] If $W_1, \ldots, W_k = \emptyset$, then
$\PLex_i S_i$ is isomorphic to $\PLex_{i \ge k+1} S_i$.
\end{enumerate}
\eLem

\prf
We have to show that $\le$ is reflexive, transitive, and
    satisfies A1. It is easy to see that, by definition, $\le$ is
    reflexive. Next, we consider
    transitivity. Suppose that $\Pl_{\PLex S_i}(A) \ge \Pl_{\PLex S_i}(B)$
    and $\Pl_{\PLex S_i}(B) \ge \Pl_{\PLex S_i}(C)$. If
    $\Pl_i( B \inter \Omega_i ) =
    \bot_i$ for all $i$, then clearly
              $\Pl_(C \inter W_i) = \bot_i$ for all $i$ (since
$\Pl_{\PLex
    S_i}(B) \ge \Pl_{\PLex S_i}(C)$), so
$\Pl_{\PLex
    S_i}(A) \ge \Pl_{\PLex S_i}(C)$. So suppose that $\Pl(B \inter
    \Omega_i) > \bot_i$ for some $i$. Let $i$ and $j$ be the smallest indexes
    such that $\Pl_i( A \inter \Omega_i )  > \bot_i$ and $\Pl_j( B
    \inter \Omega_j ) > \bot_j$. It is easy to see that $i \le j$, and
    that $\Pl_k( C \inter \Omega_k ) = \bot_k$ for all $k \le j$. If
    $i < j$, we conclude that $\Pl_i(A \inter \Omega_i) \ge \Pl_i(C
    \inter \Omega_i) = \bot_i$, and thus $\Pl_{\PLex
    S_i}(A) \ge \Pl_{\PLex S_i}(C)$. On the other hand, if $i = j$,
    then by definition $\Pl_i(A \inter \Omega_i ) \ge \Pl_i(B \inter
    \Omega_i)$, and $\Pl_i(B \inter \Omega_i) \ge \Pl_i(C \inter
    \Omega_i)$. Since $\le$ is transitive in $S_i$, we get that
    $\Pl_i(A \inter \Omega_i ) \ge \Pl_i(C \inter \Omega_i)$. Thus, we
    conclude that $\Pl_{\PLex S_i}(A) \ge \Pl_{\PLex S_i}(C)$,
    as desired.
Finally, we consider A1. Suppose that $A \subseteq B$. Then $A \inter
    \Omega_i \subseteq
    B \inter \Omega_i$ for all $i$. Since each $S_i$ satisfies A1,
    we have that $\Pl_i(A \inter \Omega_i ) \le \Pl_i(C \inter
    \Omega_i)$ for all $i$. It easily follows that
    $\Pl_{\PLex S_i}(A) \le \Pl_{\PLex S_i}(B)$.

Suppose that $S_i$ is qualitative for all $i$. We have to
    show that $\PLex_i S_i$ is also qualitative. We start by considering A2.
    Suppose that $A, B$, and $C$ are pairwise disjoint sets such that
    $\Pl_{\PLex S_i}(A \union B) > \Pl_{\PLex S_i}(C)$ and
    $\Pl_{\PLex S_i}(A \union C) > \Pl_{\PLex S_i}(B)$. Let $i$ and
    $j$ be the minimal indexes such that $\Pl_i((A\union B)\inter
    \Omega_i) > \bot_i$
    and $\Pl_j((A\union C)\inter \Omega_j) > \bot_j$. We claim that $i
    = j$. Assume, by way of contradiction, that $i  < j$. Then,
    $\Pl_i((A \union C) \inter \Omega_i )
    = \bot_i$ and hence $\Pl_i(A \inter
    \Omega_i) = \bot_i$. Moreover, since $\Pl_{\PLex S_i}(A \union C)
    > \Pl_{\PLex S_i}(B)$, we get that $\Pl_i(B \inter
    \Omega_i) = \bot_i$. Using A3 in $S_i$,
    we conclude that $\Pl_i((A\union B)\inter \Omega_i) = \bot_i$,
    which contradicts our assumption that  $\Pl_i((A\union B)\inter
    \Omega_i) > \bot_i$.
Symmetric arguments show that we also cannot have $j < i$.
Thus, $i = j$.
By definition
    $\Pl_i((A \union B)\inter\Omega_i) >
    \Pl_{i}(C\inter\Omega_i)$ and  $\Pl_i((A \union C)\inter\Omega_i) >
    \Pl_{i}(B\inter\Omega_i)$. Using A2 we conclude that
    $\Pl_i(A\inter\Omega_i) > \Pl_{i}((B\union C)\inter\Omega_i)$. It
    is also easy to verify, using A3, that $\Pl_j((B
    \union C) \inter \Omega_j) = \bot_j$ for all $j < i$. Thus, we get that
    $\Pl_{\PLex S_i}(A) > \Pl_{\PLex S_i}(B \union C)$, as desired.
Next, consider A3. The construction of $\PLex S_i$ is such that
    $\Pl_{\PLex S_i}(A) = \bottom$ if and only if $\Pl_1(A \inter
    \Omega_i) = \bottom$ for all $i$. It is easy to see
    that A3 follows from A3 in each $S_i$.

Finally, part (c) follows immediately from the definition, part (d)
    follows immediately from COND, part (e) is a
    special case of part (d), and part (f) follows immediately from the
definition.
\eprf

Returning to the proof of \Tref{thm:coherence},
first suppose that REF is not in $\A$.
Let $\Sys = (\R,\pi,\Plass_1, \ldots, \Plass_n)$ be a coherent synchronous
    system satisfying perfect recall and CONS.
Roughly speaking, the proof goes as follows. We construct a system
   $\Sys'$ which consists of countably many copies of $\R$.
   The runs in $\R^m$, the $m$th copy of $\R$, are used to simulate the
    agent's plausibility assessment at time $m$. More precisely, for
    all times $m$, we
define
    a prior on $\R^m$ that corresponds to the agent's
     plausibility measure at time $m$ in $\Sys$. These priors are then
    combined using $\PLex$ to construct the agent's prior in $\Sys'$.
Since $\PLex$ orders the priors lexicographically, if $m < m'$,
the priors on $\R^m$ dominate those on $\R^{m'}$.
The construction
guarantees that at time $m$, the agent considers possible only runs in
$\R^m \union\R^{m+1} \union \ldots$.
Since the prior on $\R^m$ dominates the
    rest, the agent's
    plausibility measure at time $m$ is similar to that
    at time $m$ in $\Sys$. This similarity
is what guarantees that conditional formulas are evaluated in the same
way in $\Sys$ and
    $\Sys'$.
    This ``peeling away'' of copies of $\R$
ensures that all temporally linear formulas holding
    in runs in
    $\Sys$ are also satisfied in the corresponding runs in $\Sys'$.

The formal construction proceeds as follows.
    Let $R \subseteq \R$ and $l
    \in \IN^*$ (recall that $\IN^* = \IN \union \{ \infty\}$).
    Define $R^l = \{ r^l : r \in R \}$, where,
    for each $i \in \{e,1\ldots,n\}$, we have
$$r^l_i(m) = \left\{ \begin{array}{ll}
\<r_i(m),m\> & \mbox{\ if $l \ge m$} \\
\<r_i(l),m\> & \mbox{\ if $l < m$.}
\end{array}\right.
$$
Let $\Sys' = (\R',\pi',\P'_1,\ldots,\P'_n)$, where $\R' =
    \union_{l \in \mbox{\scriptsize$I\!\!N$}^*} \R^l$, $\pi'$ is
    defined so that
    if $m \le l$ then $\pi'(r^l,m) = \pi(r,m)$ and if $m > l$ then
    $\pi(r^l,m) = \pi(r,l)$, and $\Plass_i'$ is defined by the priors
    described below.

To define a prior on $\R'$, we first define a plausibility space
$\P^m_{(r,i)}$ on $\R^m$ for each $m \in \IN$, run $r \in \R$, and agent
$i$.  We want the time $m$ projection of $\P^m_{(r,i)}$ to be isomorphic
to $\Plass_i(r,m)$.  To achieve this, we define
    $\P^m_{(r,i)} = (\R^m_{(r,i)}, \Pl^m_{(r,i)})$, where
    $\R^m_{(r,i)} = \R(\Omega_{(r,m,i)})^m$ and
    $\Pl^m_{(r,i)}$ is defined so that for
   $A \subseteq \Omega_{(r,m,i)}$, we have
   $\Pl^m_{(r,i)}((\R(A))^m) = \Pl_{(r,m,i)}(A)$.
For $\l \in \IN^*$,
we define the prior of agent $i$ at run $r^l$ to be the combination of
these priors for all time
points:
    $\P'_{(r^l,i)} = \PLex_m \P^{m}_{(r,i)}$.

It is easy to see that $\Sys'$ is synchronous.
It is also
    easy to check that $\Sys'$ satisfies perfect recall:
{F}rom the definition, we have that
$$
\K'_i(r^l,m) = \left\{
\begin{array}{ll}
\{ (r'^{l'},m) : (r',m) \in \K_i(r,m), l' \ge m \} &
\mbox{\ if $l \ge m$ } \\
\{ (r'^{l},m) : (r',l) \in \K_i(r,l) \} &
\mbox{\ if $l <  m$. }
\end{array}\right.
$$
Moreover, since $\Sys$ satisfies perfect recall, we have that $\R(\K_i(r,m+1))
    \subseteq \R(\K_i(r,m))$.
We conclude that $\R'(\K'_i(r^l,m+1)) \subseteq \R'(\K'_i(r^l,m))$,
which is just what we need for perfect recall.

Let $\phi \in \L^{KC}$ (so that $\phi$ does not include any temporal
    modalities) and $l \ge m$. We show that $(\Sys',r^l,m) \sat\phi$
    if and only if $(\Sys,r,m) \sat \phi$. As usual we prove this by
    induction on the structure of $\phi$. The only interesting cases
    are these that directly involve modalities.

We start with the $\Know_i$ modality.  Suppose that
    $(\Sys,r,m) \sat \Know_i\phi$. Then for all points $(s,m) \in
    \K_i(r,m)$, we have $(\Sys,s,m) \sat \phi$. Let $(s^{k},m) \in
    \K'_i(r^l,m)$. From the definition of $\Sys'$ we get that $(s,m)
    \in K_i(r,m)$ and $k \ge m$. Using the induction
    hypothesis, we get that $(\Sys',s^{k},m) \sat \phi$. We
    conclude that $(\Sys',r^l,m) \sat \Know_i\phi$. Now suppose that
    $(\Sys,r,m) \not\sat\Know_i\phi$. Then there is a point $(s,m)
    \in \K_i(r,m)$ such that
    $(\Sys,s,m) \sat \neg\phi$. Using the induction hypothesis we
    conclude that $(\Sys,s^m,m) \sat \neg\phi$. Since $(s^m,m) \in
    \K'_i(r^l,m)$, we conclude that $(\Sys',r^l,m) \not\sat
    \Know_i\phi$.

We now turn to the $\Condi$ modality.
The definition of PRIOR implies that $\P'_i(r^l,m)$ is the projection of
    $\P'_{(r^l,i)}$ conditioned on $\R'(\K'_i(r^l,m))$.
Now $\P'_{(r^l,i)} = \PLex_m \Pl^{m}_{(r,i)}$.
    Parts (d) and (f) of \Lref{lem:Plex} imply that
$\P'_{(r^l,i)}|_{\R'(\K'_i(r^l,m))}$ is isomorphic
    to $\PLex_{k \ge m} (\P^{k}_{(r,i)})|_{\R'(\K'_i(r^l,m))})$.
Consider the first term in the ``sum'',
 $\P^{m}_{(r,i)}|_{\R'(\K'_i(r^l,m))}$.
    Since $\Sys$ satisfies CONS, we have that
    $\Omega_{(r,m,i)} \subseteq \K_i(r,m)$. Thus,
conditioning on $\R'(K'_i(r^l,m))$ does not remove any runs from
$\R^m_{(r,i)} = (\R(\Omega_{(r,m,i)})^m$.
It follows 
   that $\P^{m}_{(r,i)}|_{\R'(\K'_i(r^l,m))} = \P^{m}_{(r,i)}$
which is isomorphic to $\Plass_i(r,m)$ under the mapping
$r'^m \mapsto
    (r',m)$. Finally, since
    $\P^m_{(r,i)}$ is the first plausibility space
    in the ``sum'', it determines the
    ordering of all
    pairs of sets, unless both of them are assigned plausibility
    $\bot$ by $\Pl^m_{(r,i)}$.
Putting all
this together, we conclude
    that if $A',B' \subseteq \Omega'_{(r^l,m,i)}$ and $A, B \subseteq
    \Omega_{(r,m,i)}$ such that
$(\R(A))^m = \R'(A') \inter \R^m$ and
$(\R(B))^m = \R'(B') \inter \R^m$, and if $\Pl_{(r,m,i)}(A)
    > \bot$, then $\Pl'_{(r^l,m,i)}(A') \ge \Pl'_{(r^l,m,i)}(B')$ if and
    only if $\Pl_{(r,m,i)}(A) \ge \Pl_{(r,m,i)}(B)$.

Assume that $(\Sys,r,m) \sat \phi\Condi\psi$.
Thus, either $\Pl_{(r,m,i)}(\intension{\phi}_{(r,m,i)}) = \bot$ or
$\Pl_{(r,m,i)}(\intension{\phi \land \psi}_{(r,m,i)}) >
\Pl_{(r,m,i)}(\intension{\phi \land \neg \psi}_{(r,m,i)})$.
    If $\Pl_{(r,m,i)}(\intension{\phi}_{(r,m,i)}) =
    \bot$, then from
    the coherence of $\Sys$ it follows that
    if $A
    \subseteq \Omega_{(r,l',i)}$ and $\R(A) \subseteq
    \R(\intension{\phi}_{(r,m,i)})$, then
    $\Pl_{(r,l',i)}(A) = \bot$.
    This implies that
    $\Pl^{l'}_{(r,i)}( (\R(\intension{\phi}_{(r,m,i)})^{l'} \inter
    \R(\Omega_{(r,l',i)})^{l'}) = \bot$ for all
$l' \ge m$. Since
    $\K'_i(r^l,m)$ contains only
    points from $\R^{l'}$ for $l' \ge m$, we get that
    $\Pl'_{(r^l,m,i)}(\intension{\phi}_{(r^l,m,i)}) =
    \bot$. Thus, we conclude that $(\Sys',r^l,m) \sat
    \phi\Condi\psi$ in this case.
Now suppose that
$\Pl_{(r,m,i)}(\intension{\phi \land \psi}_{(r,m,i)}) >
\Pl_{(r,m,i)}(\intension{\phi \land \neg \psi}_{(r,m,i)})$.
If we could show that
 $(\R(\intension{\phi}_{(r,m,i)}))^m =
    \R'(\intension{\phi}_{(r^l,m,i)}) \inter \R^m$, and similarly for
    $\psi$, then we could apply the argument of the previous paragraph
to show that
$\Pl'_{(r^l,m,i)}(\intension{\phi \land \psi}_{(r^l,m,i)}) >
\Pl'_{(r^l,m,i)}(\intension{\phi \land \neg \psi}_{(r^l,m,i)})$.  This,
in turn, would allow us to conclude that $(\Sys',r^l,m) \sat
\phi \Condi\psi$.  The fact that
 $(\R(\intension{\phi}_{(r,m,i)}))^m =
    \R'(\intension{\phi}_{(r^l,m,i)}) \inter \R^m$ follows from the
following chain of equivalences:

\begin{tabular}{ll}
    &$s^m \in
 (\R(\intension{\phi}_{(r,m,i)}))^m$\\
 iff &$(s,m) \in
    \intension{\phi}_{(r,m,i)}$\\
    iff &$(s,m) \in
    \Omega_{(r,m,i)}$ and $(\Sys,s,m) \sat \phi$\\
 iff &$s^m \in (\R(\Omega_{(r,m,i)}))^m = \R^m_{(r,i)}$ and
    (by the induction hypothesis) $(\Sys',s^m,m) \sat \phi$\\
    iff &$(s^m,m) \in \Omega_{(r^l,m,i)}$ and $(\Sys',s^m,m) \sat \phi$\\
    iff &$(s^m,m) \in \intension{\phi}_{(r^l,m,i)}$\\
    iff  &$s^m \in \R(\intension{\phi}_{(r^l,m,i)}) \inter \R^m$.
\end{tabular}

\noindent
Thus, in either case, we conclude that $(\Sys',r^l,m) \sat
    \phi\Condi\psi$, as desired.

For the converse, suppose
that $(\Sys,r,m) \not\sat \phi\Condi\psi$. Then
    $\Pl_{(r,m,i)}(\intension{\phi}_{(r,m,i)}) > \bot$ and
    $\Pl_{(r,m,i)}(\intension{\phi\land\psi}_{(r,m,i)}) \not>
    \Pl_{(r,m,i)}(\intension{\phi\land\neg\psi}_{(r,m,i)})$.
By the same arguments as above, we get that
    $\Pl'_{(r^l,m,i)}(\intension{\phi\land\psi}_{(r^l,m,i)}) >
    \bot$ and
    $\Pl'_{(r^l,m,i)}(\intension{\phi\land\psi}_{(r^l,m,i)}) \not>
    \Pl'_{(r^l,m,i)}(\intension{\phi\land\neg\psi}_{(r^l,m,i)})$. Thus,
    $(\Sys',r^l,m) \not\sat \phi\Condi\psi$, as desired.

Finally, for $r \in \R$,
define $f(r) = r^\infty$. We have proved that if
    $\phi \in \L^{KC}$, then $(\Sys,r,m) \sat \phi$ if and only if
    $(\Sys',f(r),m) \sat \phi$. Since this holds for all $m$,
a straightforward argument by induction on structure shows that this
holds, not just for formulas in $\L^{KC}$, but for all temporally linear
formulas.
We now have to ensure that $\Sys'$ satisfies $\A$.
Suppose that $\Sys$ satisfies QUAL. Thus, $\Plass_i(r,m)$
   is qualitative for all agents $i$, runs $r \in R$, and times $m$. Using
    part (b)  of \Lref{lem:Plex}, we conclude that the prior
    $\P'_{(r,i)}$ is qualitative for all agents $i$ and runs $r \in R$.
This implies, using \Pref{pro:prior-prop}, that $\Sys'$ satisfies QUAL.
Similarly, if $\Sys$ satisfies RANK, using part (c) of \Lref{lem:Plex}
and \Pref{pro:prior-prop},
    we get that $\Sys'$ satisfies RANK.

Suppose that $\Sys$ satisfies NORM. Then
    $\Pl_{(r,m,i)}(\intension{\True} > \bot$ for
    all agents $i$, runs $r \in \R$, and times $m$. This implies that
    $\neg(\True\Condi\False)$ is valid in $\Sys$.
Suppose that
    $l \ge m$. Then
    since $\neg(\True\Condi\False) \in \L^{KC}$, we conclude from the
    proof above that $(\Sys',r^l,m) \sat \neg(\True\Condi\False)$.
    Thus, $\Pl'_{(r^l,m,i)}(\intension{\True}_{(r^l,m,i)}) > \bot$.
Suppose that
    $l < m$. By definition, we have that $\R'(\K'_i(r^l,m)) =
    (\R(\K_i(r,l)))^l$. Using part (e) of \Lref{lem:Plex}, we get that
$\P'_{(r,i)}|_{\R'(\K'_i(r^l,m))}$ is isomorphic to
    $\P^l_{(r,i)}$. However, the latter plausibility space is
    isomorphic to $\Plass_i(r,l)$. Thus, it satisfies $\top > \bot$.
    We conclude that $\Sys'$ satisfies NORM, as desired.

Up to now we have assumed that REF is not in $\A$.  If REF
is in $\A$, then REF does not hold for $\A$, although it does hold
at many points.  To understand the issue,
suppose that REF holds in $\Sys$.  Since $\Sys'$ satisfies PRIOR,
to show that REF holds in $\Sys'$,
according to \Pref{pro:prior-prop}
it suffices to show that
all priors satisfy REF.  This is indeed
the case if $l \ne \infty$.  For suppose that $r^l \in A \subseteq \R'$.
We want to show that $\Pl_{(r^l,i)}(A) > \bot$.
Recall that $\P'_{(r^l,i)} = \PLex_m \P^m_{(r,i)}$.
{F}rom the definition of $\PLex$, it easily
follows that if $\Pl^l_{(r,i)}(A \inter \R^l) > \bot$, then
$\Pl'_{(r^l,i)}(A) > \bot$.  By definition, we have that
$\Pl'_{(r^l,i)}(A \inter \R^l) = \Pl_{(r,l,i)}(A')$, where
$A' = \{(s,l): s^l \in A\}$.  Clearly $(r,l) \in A'$, since
$(r^l,m) \in A$.  Since $\Sys$ satisfies REF, we must have that
$\Pl_{(r,l,i)}(A') > \bot$.  It follows that $\Pl'_{(r^l,i)}$
satisfies REF if $l \ne \infty$.
This argument breaks down if $l = \infty$.  Indeed, it is clear that
$\P'_{(r^{\infty},i)}$ does {\em not\/} satisfy REF.
Since $\R^\infty$ is disjoint from $\R^m$ for $m < \infty$, and we
    only ``sum'' $\P^m_{(r,i)}$ for $m < \infty$ to obtain
    $\P'_{(r^\infty,i)}$, it
    follows that $\R^\infty$ is disjoint from
    $\Omega'_{(r^\infty,i)}$, so REF does not hold.

Fortunately, a slight modification of the construction of $\Sys'$ can
be used to deal with the case $\REF \in \A$.
Define $\P^\infty_{(r,i)} = ( R^\infty_{(r,i)}, \Pl^\infty_{(r,i)})$,
    where $R^\infty_{(r,i)} = \{r^\infty\}$ and
    $\Pl^\infty_{(r,i)}( \{ r^\infty \} ) > \bot$.
Modify the construction of $\Sys'$ so that the prior of agent
    $i$ in run $r^l$ is $\P''_{(r^l,i)} =
    \P'_{(r^l,i)} \PLex \P^\infty_{(r^l,i)}$.
(Thus, $\P''_{(r^l,i)} = \PLex_{m \le \infty} \P^m_{(r^l,i)}$.)
It is easy to check that $\Sys'$ now does satisfy REF.  The argument
in the case that $l \ne \infty$ remains unchanged.  On the other
hand, if $r^\infty \in A \subseteq \R'$, it is immediate that
$\P^\infty(A \inter \R^\infty) > \bot$, so we can now deal
with this case as well.  If QUAL, RANK, or NORM is in $\A$, it
is easy to see (using the same argument as above) that $\I'$
also satisfies QUAL, RANK, or NORM.

It remains to show that this modification of the prior does not
affect the evaluation of formulas.  That is, we must show that
$(\I,r,m) \sat \phi$ if and only if $(\I',r^l,m) \sat \phi$ for all $l \ge m$.
Again, we proceed by induction on the structure of formulas.
The argument for formulas of the form $K_i \phi$ goes through
unchanged, since the changes to $\Pl'$ did not affect the $\K_i$
relations.  The argument for formulas of the form $\phi \Condi\psi$
goes through with almost no change.  The only case that requires
attention is if $(\Sys,r,m) \sat \phi \Condi \psi$ and $\intension{\phi}_
{(r,m,i)} = \bot$.  Our earlier arguments showed that
    $\Pl^{l'}_{(r,i)}( (\R(\intension{\phi}_{(r,m,i)})^{l'} \inter
    \R(\Omega_{(r,l',i)}))^{l'}) = \bot$ for all
$l' \ge m$, $l' \ne \infty$.  These arguments go through without change.
We must now show that this also holds if $l' = \infty$.
But, from the definition of $\Pl^\infty$, we get that
    $\Pl^\infty_{(r,i)}( (\R(\intension{\phi}_{(r,m,i)})^\infty \inter
\R^\infty) = \bot$ unless $r^\infty \in
    \R(\intension{\phi}_{(r,m,i)})^\infty$.
This implies that $(r,m) \in \intension{\phi}_{(r,m,i)}$.
But this cannot happen, since $\Pl_{(r,m,i)}(\intension{\phi}_{(r,m,i)})
= \bot$ and $\Sys$ satisfies REF.
\eprf

\rethm{thm:coherence-UNIF}
Let $\A$ be a subset of $\{ \QUAL, \NORM, \REF, \SDP, \UNIF, \RANK \}$
and let $\Sys$ be a coherent synchronous system satisfying perfect
recall, CONS, PERSIST,
    and $\A$. Then there is a
    synchronous system $\Sys'$ satisfying perfect recall, PRIOR, and
    $\A$, and a mapping $f:\R \mapsto \R'$ such that for all
    temporally linear formulas $\phi\in\L^{KCT}$, $(\Sys,r,m)\sat
    \phi$ if and only if $(\Sys',f(r),m) \sat \phi$.
\erethm

\prf
Suppose that $\Sys =
    (\R,\pi,\Plass_1, \ldots, \Plass_n)$ is a coherent synchronous system
    satisfying perfect recall, CONS, PERSIST,
    and $\A$. If neither $\CONS$ nor $\UNIF$ are in $\A$,
    then \Tref{thm:coherence} guarantees that there is a system
    $\Sys'$ that satisfies the stated properties.

Suppose that $\UNIF \in \A$, but $\SDP, \REF \notin \A$.
(We sketch the modifications required to deal with SDP and REF below.)
It does not follow that the system $\Sys'$ constructed in the proof
satisfies UNIF.  To see why,
suppose $r,r'$ and $m > k$ are such that $(r',k)
    \in \Omega_{(r,k,i)}$ but $(r,m) \not\sim_i (r',m)$.
UNIF implies that $\Plass_i(r,k) = \Plass_i(r',k)$ and (since $\Sys$ also
satisfies CONS) that
$\Omega_{(r,m,i)} \inter \Omega_{(r',m,i)} = \emptyset$.
Hence, our construction guarantees that $\P'_{(r^k,i)} \ne \P'_{(r'^k,i)}$,
although $r'^k \in \Omega'_{(r^k,i)}$.  Thus, the prior
in $\Sys'$ does not satisfy UNIF.  It follows that $\Sys'$ does not
satisfy UNIF either, for $\P'_i(r^k,k) \ne \P'_i(r'^k,k)$, although
    $(r'^k,k) \in
    \Omega'_{(r^k,k,i)}$.

The solution to this problem is relatively straightforward.  We modify
our construction so that the prior does indeed satisfy UNIF.
In particular, we modify the prior $\P'$ to ensure that if $\Plass_i(r,k) =
\Plass_i(r',k)$, then $\P'_{(r^k,i)} = \P'_{(r'^k,k)}$.  Of course, we have
to do so carefully, so as to make sure that nothing goes wrong with the
rest of the argument in      \Tref{thm:coherence}.

\newcommand{\PTimes}{\otimes}

We start with a modification of the construction of $\PLex$ that takes
sets (rather than sequences) of plausibility spaces and returns a new
plausibility space.

\Lem\label{lem:Ptimes}
Let $\S$ be a set of
    plausibility spaces such that the sets $\{ \Omega :
    (\Omega,\Pl) \in \S \}$ are pairwise
    disjoint. Then there is a plausibility space $\PTimes \S$
    such that
\begin{itemize}\denselist
\item[(a)] if $S = (\Omega,\Pl) \in \S$, then $\PTimes \S |_{W}$ is
isomorphic to $S$ under the identity mapping,
 \item[(b)] if $S$ is qualitative for all $S \in \S$, then $\PTimes \S$ is
    qualitative,
\item[(c)] if $S$ is ranked for all $S \in \S$, then $\PTimes \S$ is
    ranked.
\end{itemize}
\eLem

\prf
Without loss of generality there is an ordinal $\lambda$ and a sequence
    $\{ S_i: 0 \le i < \lambda \}$ such that $S_i \in \S$ for all $i$,
    and for all $S \in \S$, exists an $i$ such that $S = S_i$.%
\footnote{
If $\S$ is uncountable, this construction may require the axiom of
choice.  There is a variant of the construction that does not require
the axiom of choice, but the additional complexities involved do not
seem worth the trouble.}
Define
$\PTimes \S = \PLex_i S_i$.
Part (a) of \Lref{lem:Plex} guarantees that $\PTimes \S$ is a
plausibility space.  Parts (a), (b), and (c) follow
    immediately from parts (e), (b), and (c) of \Lref{lem:Plex},
respectively. \eprf

\newcommand{\PPART}[1]{[#1]}

Recall that to satisfy UNIF and PRIOR, it suffices to find a partition of
    $R$ such that all the runs in each cell have the same prior.
We now examine a possible way of partitioning the runs in the system.
Let $r \in R$. Define $\PPART{r,m}_i = \{ (r',m) : (r',m) \sim_i (r,m),
    \Plass_i(r',m) = \Plass_i(r,m) \}$.
    Thus, $\PPART{r,m}_i$ is the set of points in which agent $i$ has
    the same knowledge state and plausibility assessment as at $(r,m)$.
    (Note that if $\Omega_{(r,m,i)} \neq \emptyset$, then since $\Sys$
    satisfies CONS, $\Plass_i(r',m) = \Plass_i(r,m)$ implies that $(r',m)
    \sim_i (r,m)$.)

\Lem\label{lem:PPART}

\begin{itemize}\denselist
 \item[(a)] For all times $m$, the collection $\{
    \R(\PPART{r,m}_i) : r \in \R\}$ is a partition of $\R$.
 \item[(b)] For all times $m$ and runs $r$, $\Omega_{(r,m,i)}
    \subseteq \PPART{r,m}_i$.
 \item[(c)] For all times $m$ and runs $r$, $\R(\PPART{r,m+1}_i)
    \subseteq \R(\PPART{r,m}_i)$.
 \item[(d)] For all times $m$ and runs $r, r'$ such that $(r',0) \in
    \PPART{r,0}_i$, if $(r',m) \sim_i (r,m)$, then $(r',m) \in \PPART{r,m}_i$.
\end{itemize}
\eLem

\prf
By definition, if $(r',m) \in \PPART{r,m}_i$, then $\PPART{r',m} =
    \PPART{r,m}_i$. Thus, if $\PPART{r,m}_i \neq
    \PPART{r',m}_i$, then $\PPART{r,m}_i \inter \PPART{r',m}_i =
    \emptyset$. Part (a) follows immediately.
For part (b),
suppose that $(r',m) \in \Omega_{(r,m,i)}$. Since $\Sys$ satisfies
    CONS, we have that $(r',m) \sim_i (r,m)$. Moreover, since $\Sys$
    satisfies UNIF, we have that $\Plass_i(r',m) = \Plass_i(r,m)$. Thus,
    $(r',m) \in \PPART{r,m}_i$. We conclude that $\Omega_{(r,m,i)}
    \subseteq \PPART{r,m}_i$, as desired.
For part (c),
suppose that $(r',m+1) \in \PPART{r,m+1}_i$. This implies that
    $(r',m+1) \sim_i (r,m+1)$
and $\Plass_i(r',m+1) = \Plass_i(r,m+1)$.
Since $\Sys$ satisfies perfect recall,
    we get that $(r',m) \sim_i (r,m)$. Moreover, since $\Sys$
    satisfies PERSIST, we get that $\Plass_i(r',m) = \Plass_i(r,m)$. We
    conclude that $(r',m) \in \PPART{r,m}_i$. Thus,
    $\R(\PPART{r,m+1}_i) \subseteq \R(\PPART{r,m}_i)$, as desired.
Finally, we prove part (d) by induction on $m$. When $m = 0$, part (d)
    obviously holds. Suppose that $m > 0$, $(r',0) \in
    \PPART{r,0}_i$, and $(r',m) \sim_i (r,m)$. Since $\Sys$ satisfies
    perfect recall, we have that $(r',m-1) \sim_i (r,m-1)$. Using the
    induction hypothesis, we get that $(r',m-1) \in \PPART{r,m-1}$.
    This implies that $\Plass_i(r',m-1) = \Plass_i(r,m-1)$. Using PERSIST, we
    conclude that $\Plass_i(r',m) = \Plass_i(r,m)$.
    Thus, $(r',m) \in \PPART{r,m}_i$, as desired.
\eprf

Using both $\PLex$ and $\PTimes$,
we now construct a prior over $\R'$ that satisfies UNIF.
For $r \in R$, let $\PPART{r}_i$ abbreviate $\R(\PPART{r,0}_i)$.
Define $\P^{m}_{\PPART{r}_i} = \PTimes \{
    \P^m_{(r',i)} : r' \in \PPART{r}_i \}$,
where $\P^m_{(r,i)} = (\R^m_{(r,i)}, \Pl^m_{(r,i)})$ is the prior
defined in the proof of \Tref{thm:coherence}
    that is isomorphic to $\Plass_i(r,m)$ under the mapping $r'^m
    \mapsto (r',m)$.
We must show that
$\P^m_{(r,i)}$
is well defined; that is,
we must show that if $\P^m_{(r',i)} \ne \P^m_{(r'',i)}$, then
$\R^m_{(r',i)}$ is disjoint from $\R^m_{(r'',i)}$.
Note that if $(r',m) \in
\PPART{r'',m}_i$, then
    $\P^m_{(r',i)}$ and $\P^m_{(r'',i)}$  are identical.
Using part (b) of \Lref{lem:PPART} we get that if $(r',m) \not
    \in \PPART{r'',m}_i$, then $\R^m_{(r',i)}
    \inter \R^m_{(r'',i)} = \emptyset$, as desired.
Thus, $\P^m_{\PPART{r}_i}$ is indeed well defined.
We now define $\P'_{(r^l,i)} =
\PLex_{m} \P^{m}_{\PPART{r}_i}$
as the prior of agent $i$ in run $r^l$.

We claim that this family of priors satisfies UNIF.  Notice that
$\Omega'_{(r^l,i)} = \union_{m,r' \in \PPART{r}_i} \R^m_{(r',i)}$.
If $r'^m \in \Omega'_{(r^l,i)}$ then, by definition, $r' \in
    \Omega_{(r,m,i)}$. Using parts (a) and (b) of \Lref{lem:PPART}, we
    get that $r' \in \PPART{r}_i$.  It easily
follows that $\PPART{r'}_i = \PPART{r}_i$, so indeed the construction
    guarantees
that $\P'_{(r^l,i)} = \P'_{(r'^m,i)}$, as desired.  Since the family of
priors satisfies UNIF, so does $\Sys'$.

Let $\phi \in \L^{KC}$ and $l \ge m$.
As in the proof of
    \Tref{thm:coherence},
    we now proceed by induction on the structure of formulas
    to show that $(\Sys',r^l,m) \sat  \phi$ if
    and only if $(\Sys,r,m) \sat \phi$.
    The only difference
    arises in dealing with the $\Condi$ modality.

As before, parts (d) and (f) of \Lref{lem:Plex} imply that
$\P'_{(r^l,i)}|_{\R'(\K'_i(r^l,m))}$ is isomorphic
    to $\PLex_{k \ge m} (\P^{k}_{\PPART{r}_i}|_{\R'(\K'_i(r^l,m))})$.
    Again, we consider the first term in the ``sum'',
    $\P^{m}_{\PPART{r}_i}|_{\R'(\K'_i(r^l,m))}$.
We want to show that $\P^{m}_{\PPART{r}_i}|_{\R'(\K'_i(r^l,m))} =
\P^{m}_{(r,i)}|_{\R'(\K'_i(r^l,m))}$.
Recall that $\P^{m}_{(r,i)}|_{\R'(\K'_i(r^l,m))}$ is the first term in
    the analogous ``sum'' in the proof of \Tref{thm:coherence}.
Thus, even though we are
using a different prior from that of the proof of
    \Tref{thm:coherence}, after conditioning, they are essentially the
same.  By \Lref{lem:Ptimes}, we have that
$\P^{m}_{\PPART{r}_i}|_{\R^m_{(r,i)}}
= \P^{m}_{(r,i)}$.  Thus, it suffices to
show that  $\union_{r' \in \PPART{r}_i}
\R^m_{(r',i)} \inter \R'(\K'_i(r^l,m)) = \R^m_{(r,i)} \inter
\R'(\K'_i(r^l,m)$.  The inclusion from right to left is immediate.
For the opposite inclusion, suppose that $s^m \in
\union_{r' \in \PPART{r}_i}
\R^m_{(r',i)} \inter \R'(\K'_i(r^l,m))$.  Since $s^m \in
\R'(\K'_i(r^l,m))$, we must have $(r,m) \sim_i (s,m)$.  Since $s^m \in
\union_{r' \in \PPART{r}_i} \R^m_{(r',i)}$, there must also be some run
$r' \in \PPART{r}_i$ such that $s \in \R^m_{(r',i)}$.
Since $s \in \R^m{(r',i)}$, we have that $(s,m)
\in \Omega_{(r',m,i)}$.  By part~(b) of \Lref{lem:PPART}, $(s,m)
\in \PPART{r',m}_i$.  By part~(c) of \Lref{lem:PPART}, we get that
$(s,0) \in \PPART{r',0}_i$.  Since $(r',0) \in \PPART{r,0}_i$, it
immediately follows that $\PPART{r',0}_i = \PPART{r,0}_i$.
Hence, $(s,0) \in \PPART{r,0}_i$.  Now by part~(d) of \Lref{lem:PPART},
we get that $(s,m) \in \PPART{r,m}_i$.  Thus, $\Plass_i(s,m) = \Plass_i(r,m)$.
Since $\Sys$ satisfies UNIF and $(s,m) \in \Omega_{(r',m,i)}$, it
follows that $\Plass_i(s,m) = \Plass_i(r',m)$.  Hence, $(s,m) \in
\Omega_{(r,m,i)}$.  Finally, we can
conclude that $s \in \R^m_{(r,i)}$, as
desired.  Given this equivalence, we can deal with the $\Condi$ case
just as we did in the proof of \Tref{thm:coherence}.

Finally, we need to ensure that $\Sys'$ satisfies $\A$. The proof of
    \Tref{thm:coherence} shows that if $\Sys$ satisfies
    NORM, then so does $\Sys'$. Using parts (b) and (c) of
    \Lref{lem:Ptimes}, it easily follows
    that if $\Sys$ satisfies QUAL or RANK, then
    so does $\Sys'$.
If REF and UNIF are both in $\A$ (but SDP is not),
then we need a further modification of the prior, in the same spirit of
that in the proof of \Tref{thm:coherence}.   Define
    $\P^\infty_{\PPART{r}_i} = ( \PPART{r}_i^\infty,
    \Pl^\infty_{\PPART{r}_i})$, where
   $\Pl^\infty_{\PPART{r}_i}(\emptyset) = \bot$ and
    $\Pl^\infty_{\PPART{r}_i}(A) = \top$ for all $A \neq
   \emptyset$.
We now take the prior of the agent to be
$\P''_{(r^l,i)} =  \P'_{(r^l,i)} \PLex \P^\infty_{\PPART{r}_i} $.
It is straightforward to show that
    the resulting system satisfies REF and the requirements of the
    theorem, using essentially the same arguments for dealing with REF
as in the proof of \Tref{thm:coherence}.

Finally, suppose $\SDP \in \A$ but REF is not.
Note that, since CONS and SDP
    imply UNIF, $\Sys$ satisfies UNIF, so we can assume without loss of
generality that UNIF is also in $\A$.
To get $\Sys'$ to satisfy SDP, we further modify $\P'$ so that it
depends only on the agent, and not the run.
Thus,
we define $\P^m_i = \PTimes \{ \P^m_{\PPART{r}_i} : r \in \R
\}$, and define $\P'_{(r^l,i)} = \PLex_m \P^m_i$.
Clearly, with this prior, $\Sys'$ satisfies SDP.  Again, we need to
check that this change in prior does not affect the rest of our
argument.  Once more, the only difficulty comes in dealing with the
$\Cond_i$ case.  Just as in the case of UNIF, we proceed by showing that
$\P^{m}_i|_{\R'(\K'_i(r^l,m))} =
\P^{m}_{(r,i)}|_{\R'(\K'_i(r^l,m))}$.
The argument is actually even easier than that for UNIF:
We show that $\union_{r'} \R^m_{(r',i)} \inter
\R'(\K'_i(r^l,m)) = \R^m_{(r,i)} \inter \R'(\K'_i(r^l,m)$.
Again, the inclusion from right to left is immediate.  For the
opposite inclusion, suppose that $s^m \in
\union_{r'} \R^m_{(r',i)} \inter \R'(\K'_i(r^l,m))$.
Since $s^m \in
\R'(\K'_i(r^l,m))$, we must have $(r,m) \sim_i (s,m)$.  Since $s^m \in
\union_{r'} R^m_{(r',i)}$, there must also be some run
$r'$ such that $s \in \R^m_{(r',i)}$.  Thus, $(s,m) \in \Omega_{(r',m,i)}$.
Since $\Sys$ satisfies CONS, we have $(s,m) \sim_i (r',m)$.
It follows that $(r',m) \sim_i (r,m)$.  Since $\Sys$ satisfies SDP,
we must have that
$\Omega_{(r,m,i)} = \Omega_{(r',m,i)}$, so
$(s,m) \in \Omega_{(r,m,i)}$.  Therefore,
$s \in \R^m_{(r,i)}$, as desired.

The modifications to deal with the case where both SDP and REF are in
$\A$ are identical to the case with UNIF, and are omitted here.
\eprf

\bibliographystyle{alpha}
\bibliography{z,refs,bghk,conf-long}
\end{document}